\documentclass[10pt]{article} 
\usepackage[accepted]{tmlr}

\usepackage{amsmath,amsfonts,bm}

\def\eqref#1{equation~\ref{#1}}

\def\1{\bm{1}}

\DeclareMathAlphabet{\mathsfit}{\encodingdefault}{\sfdefault}{m}{sl}
\SetMathAlphabet{\mathsfit}{bold}{\encodingdefault}{\sfdefault}{bx}{n}

\usepackage{hyperref}
\usepackage{url}

\usepackage{amsmath}
\usepackage{amssymb}
\usepackage{mathtools}
\usepackage{amsthm}
\usepackage{graphicx}
\usepackage{enumitem}
\usepackage{stfloats}
\usepackage{ulem}
\usepackage{adjustbox}
\usepackage{multirow}
\usepackage{array}
\usepackage{float}
\usepackage{multirow}

\usepackage{wrapfig}
\usepackage{xspace}
\usepackage{array}
\usepackage{layouts}
\usepackage{algorithm}
\usepackage{listings}
\usepackage{pifont}
\usepackage{wasysym}
\usepackage{footnote}
\usepackage{subcaption}
\usepackage{booktabs}

\usepackage{xcolor}
\hypersetup{
    colorlinks,
    linkcolor={red},
    citecolor={blue},
}

\definecolor{demphcolor}{RGB}{144,144,144}
\newcommand{\demph}[1]{\textcolor{demphcolor}{#1}}

\usepackage{xcolor,colortbl}
\definecolor{FCEDE6}{RGB}{252,237,230}
\colorlet{Light}{FCEDE6}
\newcommand{\CC}[1]{\cellcolor{Light}}

\newcommand{\model}{VoLTA}

\usepackage[capitalize,noabbrev]{cleveref}

\title{VoLTA: Vision-Language Transformer with \\ Weakly-Supervised Local-Feature Alignment}

\author{Shraman Pramanick$^{*1,2 \dagger}$ \ Li Jing$^{*2}$ \ Sayan Nag$^{*3}$ \ Jiachen Zhu$^{4}$ \ Hardik Shah$^{2}$ \\ Yann LeCun$^{2,4}$ \ Rama Chellappa$^{1}$ \\
\
$^{1}${Johns Hopkins University} \; $^{2}$Meta \; $^{3}$University of Toronto $^{4}$New York University \\
}

\def\month{09}  
\def\year{2023} 
\def\openreview{\url{https://openreview.net/forum?id=Kt2VJrCKo4}} 
\usepackage{arydshln}

\begin{document}

\maketitle

\begin{abstract}
Vision-language pre-training (VLP) has recently proven highly effective for various uni- and multi-modal downstream applications. However, most existing end-to-end VLP methods use high-resolution image-text-box data to perform well on fine-grained region-level tasks, such as object detection, segmentation, and referring expression comprehension. Unfortunately, such high-resolution images with accurate bounding box annotations are expensive to collect and use for supervision at scale. In this work, we propose \model\ (\textbf{V}isi\textbf{o}n-\textbf{L}anguage \textbf{T}ransformer with weakly-supervised local-feature \textbf{A}lignment), a new VLP paradigm that only utilizes image-caption data but achieves fine-grained region-level image understanding, eliminating the need for expensive box annotations. \model\ adopts graph optimal transport-based weakly-supervised alignment on local image patches and text tokens to germinate an \textit{explicit}, \textit{self-normalized}, and \textit{interpretable} low-level matching criterion. In addition, \model\ pushes multi-modal fusion deep into the uni-modal backbones during pre-training and removes fusion-specific transformer layers, further reducing memory requirements. Extensive experiments on a wide range of vision- and vision-language downstream tasks demonstrate the effectiveness of \model\ on fine-grained applications without compromising the coarse-grained downstream performance, often outperforming methods using significantly more caption and box annotations. Code and pre-trained model are available at \href{https://github.com/ShramanPramanick/VoLTA}{https://github.com/ShramanPramanick/VoLTA}.
\let\thefootnote\relax\footnotetext{$^*$Equal technical contribution.}
\let\thefootnote\relax\footnotetext{$^\dagger$Part of this work was done during an internship at Meta.}

\end{abstract}

\section{Introduction}

Inspired by the escalating unification of transformer-based modeling in vision \citep{dosovitskiy2021an, liu2021swin, chen2021crossvit} and language \citep{Devlin2019BERTPO, liu2019roberta} domains, coupled with readily available large-scale \textit{image-caption} pair data, vision-language pre-training (VLP) \citep{lu2019vilbert, li2020visualbert, kim2021vilt, kamath2021mdetr, zhang2021vinvl} has recently been receiving increasing attention. VLP has not only been proven the \textit{de-facto} for several vision-language tasks, but it has also been beneficial for traditional vision-only tasks, such as image classification and object detection. Such wide-range applications of VLP can broadly be categorized into two groups: $(i)$ tasks requiring image-level understanding, e.g., image classification, image \& text retrieval \citep{plummer2015flickr30k}, image captioning \citep{zhou2020unified}, visual question answering \citep{antol2015vqa}, and $(ii)$ tasks requiring region-level understanding, e.g., object detection, instance segmentation, and referring expression comprehension \citep{kazemzadeh2014referitgame, yu2016modeling}. Most existing VLP methods address only one group of application, leaving the question of a generalizable and unified VL framework under-explored.

\begin{figure*}[!t]
\centering
\includegraphics[scale=0.4]{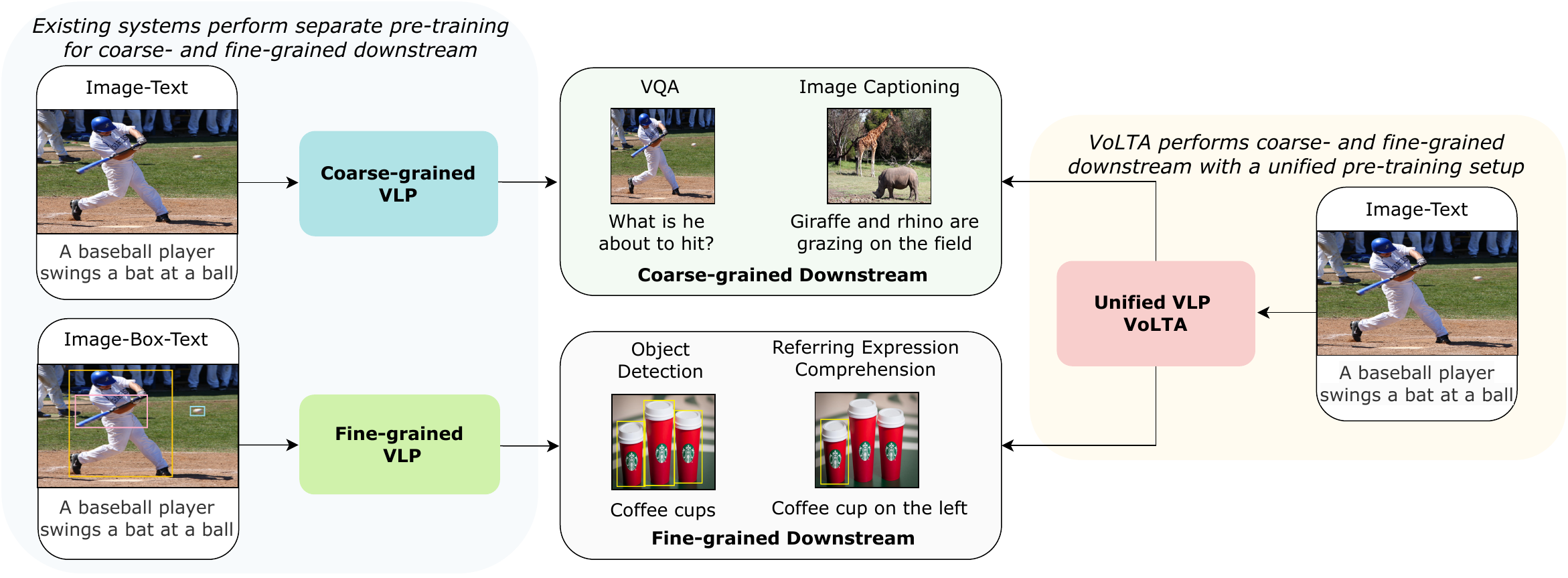}
\vspace{-2mm}
\caption{\textbf{Different Categories of VLP frameworks.} Existing VLP systems perform separate pre-training for image-level \citep{li2021align, wang2021simvlm, dou2022empirical} and region-level  \citep{kamath2021mdetr, li2022grounded, zhang2022glipv2} understanding and do not generalize well to coarse- and fine-grained downstream tasks. In contrast,  our proposed method, \model, unifies both downstream with a single pre-training setup and attains fine-grained region-level understanding without using expensive bounding box annotations during pre-training.}
\label{fig:existing_methods_volta_overview}
\vspace{-3mm}
\end{figure*}

Traditional VLP methods with image-level understanding \citep{li2021align, wang2021simvlm, dou2022empirical} utilize large-scale \textit{image-caption} pair datasets and are commonly trained with image-text contrastive objectives computed on global features. Hence, it is not trivial to extend such methods to region-level applications. On the other hand, VLP methods with region-level understanding \citep{kamath2021mdetr, li2022grounded, zhang2022glipv2} use \textit{image-text-box} grounding data and are designed to predict bounding boxes during pre-training. Consequently, they do not support image-level tasks. Furthermore, accurate bounding box annotations require high-resolution input images, which are often expensive to collect, annotate and use for pre-training at scale. Recently, FIBER \citep{dou2022coarse} addressed the problem of such unified VLP and proposed a two-stage pre-training algorithm requiring fewer box annotations than previous region-level pre-training methods. Moving a step forward, as shown in Figure \ref{fig:existing_methods_volta_overview}, we aim to eliminate the use of costly box annotations and ask the challenging but natural question: \textit{Can we attain region-level understanding from global image-caption annotations and unify image- and region-level tasks in a single VL framework?}

Subsequently, we focus on achieving region-level fine-grained understanding by weakly-supervised alignment of image patches and text tokens. Previous VLP methods \citep{chen2020uniter, kim2021vilt} in this direction use Wasserstein distance (WD) \citep{peyre2019computational}, a.k.a Earth Mover’s distance (EMD)-based optimal transport (OT) algorithms for such alignment problems. However, we argue that WD is not optimum for images with multiple similar entities. Thus, we propose to jointly utilize Gromov-Wasserstein distance (GWD) \citep{peyre2016gromov} and Wasserstein distance (WD) in a setup known as graph optimal transport \citep{chen2020graph}. Moreover, instead of using a commonly deployed contrastive objective, we propose to use redundancy reduction from Barlow Twins \citep{zbontar2021barlow}, which is less data-intensive and does not require hard-negative mining. We also follow \citet{dou2022coarse} and incorporate deep multi-modal fusion into the uni-modal backbones, removing the need for costly fusion-specific transformer layers. These steps when integrated yield \model, \textbf{V}isi\textbf{o}n-\textbf{L}anguage \textbf{T}ransformer with weakly-supervised local-feature \textbf{A}lignment, a unified VLP paradigm only utilizes \textit{image-caption} annotations but achieves fine-grained region-level image understanding, eliminating the need for expensive box annotations. Figure \ref{fig:main_visualization} visualizes the feature-level image-text alignment generated by \model, which can attend text tokens to the corresponding visual patches without relying on low-level supervision.

In summary, our contributions are three-fold. $(i)$ We propose to use graph optimal transport for weakly-supervised feature-level patch-token alignment in VLP. $(ii)$ We introduce \model, a unified VLP paradigm for image-level and region-level applications, but pre-trained only using \textit{image-caption} pairs. \model\ is memory, compute, and time-efficient and can easily be scaled up with readily available large-scale \textit{image-caption} data harvested from the web. $(iii)$ We present the results of a wide range of vision- and vision-language coarse- and fine-grained downstream experiments to demonstrate the effectiveness of \model\ compared to strong baselines pre-trained with significantly more caption and box annotations.

\section{Related Works}

\noindent \textbf{Uni-modal Self-supervised Pre-training:} In recent years, the machine learning community has observed a boom in self-supervised pre-training. In the language domain, representations learned by BERT \citep{Devlin2019BERTPO}, RoBERTa \citep{liu2019roberta} have become the default setting for many downstream tasks. Generative models such as GPT \citep{Radford2019LanguageMA, Brown2020LanguageMA} have also achieved impressive few-shot/zero-shot performances on novel applications. SimCSE \citep{Gao2021SimCSESC} uses contrastive learning to help learn useful sentence representations.

In the vision domain, several contrastive/joint-embedding methods \citep{he2020momentum, chen2020mocov2, Chen2021AnES, chen2020simple, Grill2020BootstrapYO, Chen2021ExploringSS,Caron2021EmergingPI,zbontar2021barlow,Bardes2022VICRegVR, shah2022max, Assran2022MaskedSN} have outperformed supervised counterparts. Recently, generative models such as BEiT \citep{Bao2022BEiTBP} and MAE \citep{He2021MaskedAA} have also achieved impressive performances with much more scalable potential. 

\vspace{1mm}
\noindent \textbf{Vision-Language Pre-training (VLP):} Vision-language pre-training mainly relies on \textit{image-text} pair datasets to learn joint visual-language representations. One line of work is to train separate vision and language encoders and only fuse in the representation space. CLIP \citep{radford2021learning}, UniCL \citep{yang2022unified}, and ALIGN \citep{jia2021scaling} use the image-text contrastive loss to learn aligned representations. SLIP \citep{Mu2021SLIPSM} combines self-supervised visual representation learning and contrastive multi-modal learning. M3AE \citep{Geng2022MultimodalMA}, FLAVA \citep{singh2022flava} combines masked image modeling and masked language modeling. Another line of work uses cross attention to fuse vision and language information in the early stage \citep{kamath2021mdetr, dou2022empirical, lu2019vilbert, li2020oscar, Kiela2019SupervisedMB, kim2021vilt, zhang2021vinvl, li2022blip, Wang2022ImageAA, pramanick2023egovlpv2, softmask2023, blip22023, oneR2023, git2023}. These works focus on learning semantic-level aligned vision-language representations. In addition, UniTAB \citep{yang2022unitab}, OFA \citep{wang2022ofa}, GLIP \citep{li2022grounded}, and FIBER \citep{dou2022coarse} use expensive grounding \textit{image-text-box} annotations to learn the fine-grained aligned representations. Our work uses representation space alignment and cross-attention fusion, but we do not use any box annotation to learn robust feature-level alignments.

\vspace{1mm}
\noindent \textbf{Unsupervised Representation Alignment:} Unsupervised multi-modal alignment typically relies on specific metrics. 
Wasserstein distance \citep{peyre2019computational}, a.k.a EMD-based optimal transport (OT) algorithms have been widely adopted to various domain alignment tasks, including sequence-to-sequence learning \citep{chen2018improving}, few-shot learning \citep{zhang2020deepemd}, knowledge distillation \citep{balaji2019normalized}, unsupervised domain adaptation \citep{balaji2019normalized}, generative networks \citep{han2015learning, genevay2018learning, mroueh2018sobolev, mroueh2019sobolev}, and multi-modal learning \citep{yuan2020advancing, chen2020uniter, kim2021vilt, li2022clip, pramanick2022multimodal}. Previous VLP methods \citep{chen2020uniter, kim2021vilt}, which use OT-based patch-word alignment, only utilize the Wasserstein distance. However, we argue that jointly modeling GWD \citep{peyre2016gromov} and WD results in a superior multi-modal alignment for intricate images. To the best of our knowledge, this is the first work to apply WD and GWD-based optimal transport for feature-level alignment in VLP.

\section{Proposed System - \model}

\begin{figure*}[!t]
\centering
\includegraphics[width=1\textwidth]{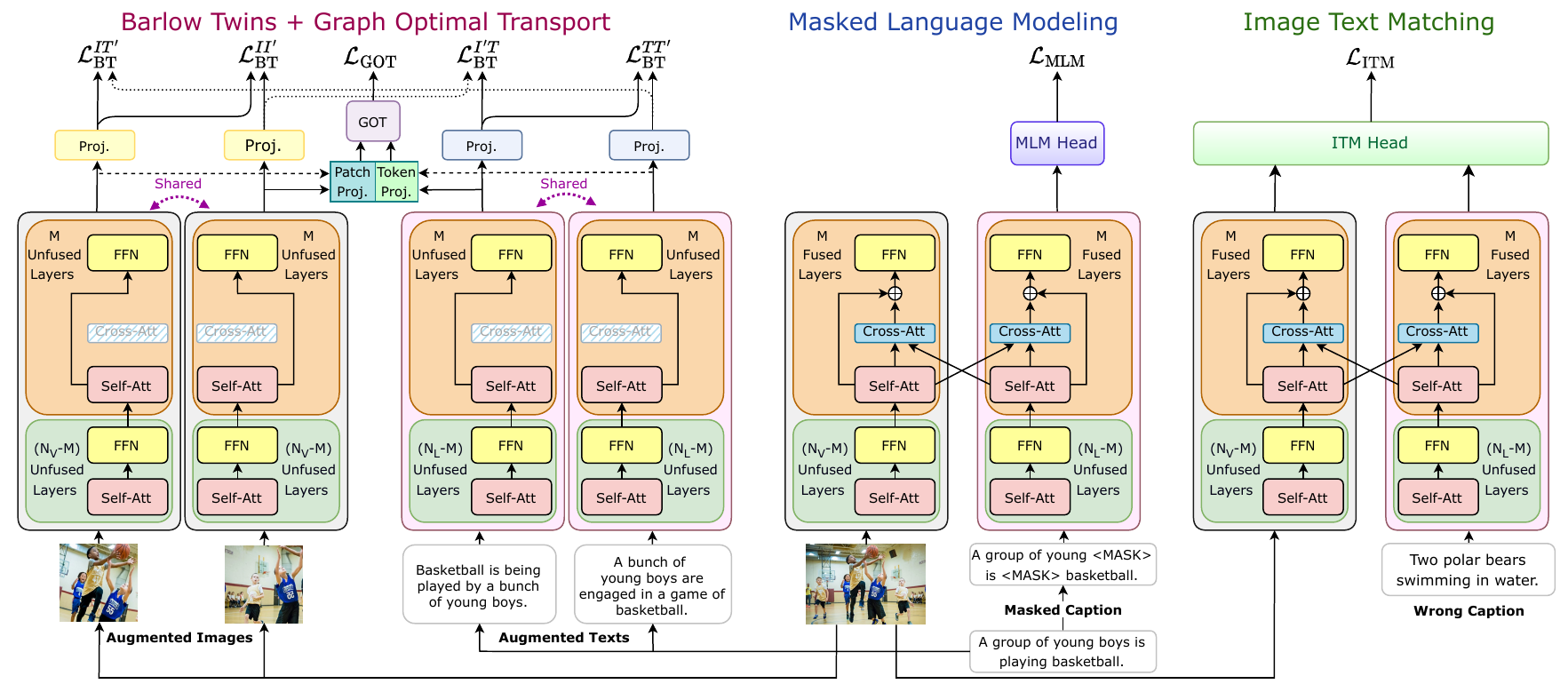}
\vspace{-4mm}
\caption{\textbf{Computation of four different objectives, $\mathcal{L}_\mathrm{BT}$, $\mathcal{L}_\mathrm{GOT}$, $\mathcal{L}_\mathrm{MLM}$, and $\mathcal{L}_\mathrm{ITM}$ by the proposed \model\ framework.} Inspired by \citet{dou2022coarse}, \model\ inserts cross-modal attention fusion (CMAF) inside uni-modal backbones with a gating mechanism. During \model\ pre-training, every forward iteration consists of three steps - $(i)$ CMAF is switched off, \model\ acts as dual encoder, $\mathcal{L}_\mathrm{BT}$ and $\mathcal{L}_\mathrm{GOT}$ are computed. $(ii)$ CMAF is switched on, \model\ acts as fusion encoder, and image-masked caption pair is fed into the model to compute $\mathcal{L}_\mathrm{MLM}$. $(iii)$ CMAF is kept on, randomly sampled image-caption pair is fed into the model to compute $\mathcal{L}_\mathrm{ITM}$. Such a fusion strategy results in a lightweight and flexible model compared to using fusion-specific transformer layers.}
\label{fig:system_overview}
\vspace{-2mm}
\end{figure*}

In this section, we present our proposed approach, \model, which contains three broad modules - $(i)$ intra- and inter-modality redundancy reduction, $(ii)$ weekly-supervised cross-modal alignment of local features, and $(iii)$ cross-modal attention fusion (CMAF). Next, we introduce the fine-tuning strategies for various uni- and multi-modal downstream tasks as supported by \model. An overview of the different modules of \model\ is presented in Figure \ref{fig:system_overview}.

\subsection{Intra- \& Inter-modality Redundancy Reduction}

We use Barlow Twins (BT) \citep{zbontar2021barlow}, a non-contrastive covariance regularization as the foundational objective of \model. The recent success of contrastive vision-language pre-training \citep{radford2021learning, li2021supervision, jia2021scaling, kim2021vilt, yang2022unified, dou2022coarse, dou2022empirical} has already shown that, compared to a single modality, \textit{image-caption} pairs offer a significantly higher-level of abstractive and semantic concepts about the training samples. However, common contrastive VLP objectives, like InfoNCE \citep{oord2018representation}, are data-hungry, as they require large batch sizes and well-mined hard negatives. On the other hand, the BT objective operates on the dimensions of the embeddings across the two views of training samples. Hence, it is more robust to batch size and can be trained using lower memory resources. In this work, we extend the BT objective for a multi-modal setup.

The original BT algorithm, which operates on joint embeddings of distorted samples, was proposed only for image modality. Specifically, for each image of a batch $\mathcal{X}$, two distorted views are obtained using a distribution of data augmentation $\mathcal{T}$ with disparate probabilities. These distorted images are then fed into a shared image encoder containing a feature extraction network (e.g., ResNet \citep{he2016deep}) cascaded with trainable linear projection layers, producing a batch of parallel embeddings $z^A$ and $z^B$. The BT loss computed using the encoded embeddings can be denoted as:
\vspace{-2mm}
\begin{gather}\label{eq:loss_BT_L}
\vspace{-3mm}
    \mathcal{L}_\mathrm{BT} \triangleq \sum_{i}\big(1 - C_{ii}\big)^2 + \lambda \sum_{i}\sum_{j \neq i}\big(C_{ij}\big)^2,
    \text{where,} \; C_{ij} = \frac{\sum_{b}z_{b,i}^A z_{b,j}^B}{\sqrt{\sum_{b}\big(z_{b,i}^A\big)^2}\sqrt{\sum_{b}\big(z_{b,j}^B\big)^2}}
\vspace{-5mm}
\end{gather}
\vspace{-4mm}

$\lambda$ is a positive weighting factor; $C$ is the cross-correlation matrix computed between $z^A$ and $z^B$ along the batch dimension; $b$ stands for sample indices in a batch; $i,j$ refers to the dimension indices of $z^A$ and $z^B$. The first term in Equation \ref{eq:loss_BT_L} is the \textit{invariance} term which attempts to equate the diagonal elements of the cross-correlation $C$ matrix to $1$, whereas the second term is the \textit{redundancy reduction term} which pushes the off-diagonal elements of $C$ matrix to $0$.

In this work, we use BT for \textit{image-caption} pairs. Specifically, we use stochastic data augmentations for both images and text\footnote{Augmentation details are provided in Appendix \ref{sec:augmentation}.}, and directly apply the BT objective for all the $2\times2$ pairs, resulting in additional supervision. Note this simple, straightforward, and instinctive extension enables us to apply redundancy reduction in between and across modalities, which intuitively results in superior visual representation. Moreover, in this bi-modal setting, we can pre-train a text encoder in parallel with the image encoder and, thus, can generalize our system to a broader range of uni- and multi-modal downstream applications. 

\noindent \textbf{Intra-modal Objective:} Intra-modal objective refers to applying the BT loss in-between pairs of image and text embeddings. Given an \textit{image-caption} pair, we first have two augmented views $(I,I')$ for each image, and two augmented views $(T,T')$ for each text. Then, we resort to Equation \ref{eq:loss_BT_L} individually for the image and text pairs. 
\begin{gather}\label{eq:loss_intra}
\vspace{-3mm}
    \mathcal{L}_\mathrm{BT}^{k} \triangleq \sum_{i}\big(1 - C^{k}_{ii}\big)^2 + \lambda \sum_{i}\sum_{j \neq i}\big(C^{k}_{ij}\big)^2,
    \forall \; k \in\{II', TT'\}
\vspace{-5mm}
\end{gather}

\vspace{-2mm}
\noindent \textbf{Inter-modal Objective:} Inter-modal objective refers to applying the BT loss across image and text embeddings. Since the image and text encoders can output features with different shapes, we design the projector layers with same output dimension. Hence, in addition to the original BT loss between $(I,I')$ in \citet{zbontar2021barlow}, we get three more loss terms - $(T,T')$, $(I,T')$, $(I',T)$, leading to $3\times$ diverse and high-quality additional supervision. The inter-modal BT losses can be directly computed following Equation \ref{eq:loss_BT_L}.

\vspace{-5mm}

\begin{gather}\label{eq:loss_inter}
\vspace{-5mm}
    \mathcal{L}_\mathrm{BT}^{k} \triangleq \sum_{i}\big(1 - C^{k}_{ii}\big)^2 + \lambda \sum_{i}\sum_{j \neq i}\big(C^{k}_{ij}\big)^2,
    \forall \; k \in \{IT', I'T\}
\end{gather}

\vspace{-2mm}

The resulting bi-modal BT loss is $\mathcal{L}_\mathrm{BT} = \sum_{k} \mathcal{L}^{k}_\mathrm{BT}$, $\forall \; k \in \{II', TT', IT', I'T\}$.

\begin{wrapfigure}{R}{0.50\textwidth}
\vspace{-11mm}
\centering
\hspace{0.75mm} \includegraphics[scale = 0.58]{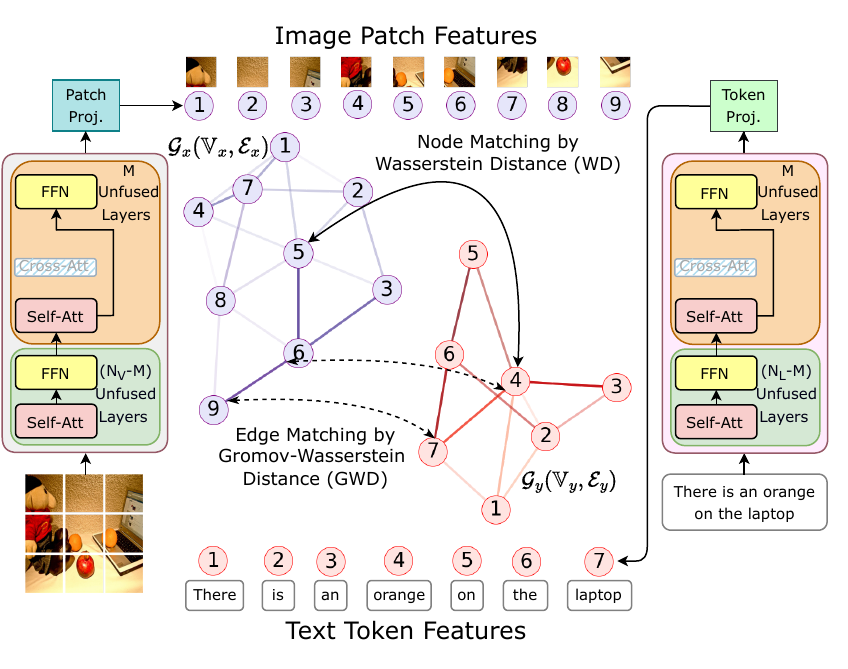}
\vspace{-7mm}
\caption{\textbf{Illustration of the graph optimal transport (GOT) algorithm used for patch-word alignment in the proposed \model\ framework.} We obtain patch- and token-level features from the last layers of corresponding visual and textual transformer encoders and use these encoded local-feature vectors to construct modality-specific dynamic graphs - $\mathcal{G}_{x}(\mathbb{V}_{x}, \mathcal{E}_{x})$ for image patches and $\mathcal{G}_{y}(\mathbb{V}_{y}, \mathcal{E}_{y})$ for text tokens. Next, we perform patch-word alignment by utilizing the Gromov-Wasserstein distance and Wasserstein distance for edge and node matching, which preserves the topological graph structure. See Section \ref{sec:alignment_local_features} for details. Darker edges denote larger weights.}
\label{fig:got_illustration}
\vspace{-3mm}
\end{wrapfigure}

\subsection{Alignment of Local Features} \label{sec:alignment_local_features}

Though the inter-modal redundancy reduction provides high-quality semantic supervision, it is computed on the global image- and text features and, thus, only simulates implicit and non-interpretable multi-modal alignment. However, fine-grained region-level downstream applications like detection, segmentation, and reference expression comprehension require local visual feature descriptors with specific spatial information. To achieve this, most existing top-performing VLP methods, including UniTAB \citep{yang2022unitab}, OFA \citep{wang2022ofa}, GLIP \citep{li2022grounded}, and FIBER \citep{dou2022coarse}, use high-resolution image-text-box data for fine-grained pre-training. However, bounding box annotations are expensive to collect and use for supervision. Hence, we seek an alternate weekly-supervised solution for local feature-level alignment using global image-caption annotations.

Recently, WD \citep{peyre2019computational}, a.k.a EMD-based OT algorithms have been used for weakly-supervised patch-word alignment in VLP \citep{chen2020uniter, kim2021vilt}. Such OT-based learning methods are optimized for distribution matching by minimizing the cost of a transport plan. We pose the patch-word alignment as a more structured graph-matching problem and use the graph optimal transport (GOT) algorithm, which utilizes GWD \citep{peyre2016gromov} in conjunction with WD to ensure the preservation of topological information during cross-modal alignment. More specifically, we obtain the patch- and token-level features from the last layers of corresponding visual and textual transformer encoders, and use these encoded local-feature vectors to construct modality-specific dynamic graphs - $\mathcal{G}_{x}(\mathbb{V}_{x}, \mathcal{E}_{x})$ for image patches and $\mathcal{G}_{y}(\mathbb{V}_{y}, \mathcal{E}_{y})$ for text tokens. Each node in these graphs $i \in \{\mathbb{V}_{x}, \mathbb{V}_{y}\}$ is represented by corresponding feature vectors, and intermediate edges $e \in \{\mathbb{E}_{x}, \mathbb{E}_{y}\}$ by thresholded cosine similarity.

\noindent \textbf{Importance of GOT in Patch-Word Alignment:} As mentioned previously, GOT adopts two types of OT distances - WD for node matching and GWD for edge matching. In contrast, previous vision-language pre-training algorithms using OT for patch-word alignment only considered WD  \citep{chen2020uniter, kim2021vilt}. However, we argue that intricate images with multiple objects with similar shapes and colors require both WD and GWD for accurate, fine-grained matching. For example, in Figure \ref{fig:got_illustration}, there are multiple \textit{``orange"} present in the image. WD can only match nodes in the graph, and will treat all \textit{``orange"} entities as identical and will ignore neighboring relations like \textit{``on the laptop"}. However, by using proper edge matching with GWD, we can preserve the graph's topological structure. We can correctly identify which \textit{``orange"} in the image the sentence is referring to. Hence, we couple WD and GWD mutually beneficially and use a joint transport plan for accurate patch-word matching. 

Once $\mathcal{G}_{x}$ and $\mathcal{G}_{y}$ are computed, we follow \citet{chen2020graph} to compute WD and GWD.

\noindent \textbf{Wasserstein Distance} calculates the pairwise distances between two sets of cross-domain node embeddings. Consider two discrete distributions, $\phi \in \mathbf{P}(\mathbb{X})$ and $\psi \in \mathbf{P}(\mathbb{Y})$, where $\phi = \sum_{i=1}^n u_i \delta_{x_i}$ and $\psi = \sum_{j=1}^m v_j \delta_{v_j}$; and $\delta_{x}$ being the Delta-Dirac function centered on $x$. Since $\phi$ and $\psi$ are both probability distributions, sum of weight vectors is $1$, $\sum_{i}u_{i} = 1 = \sum_{j}v_j$. The WD distance between $\phi$ and $\psi$ is defined as: 
\begin{equation}\label{eq:wd}
\vspace{-1mm}
	\mathcal{D}_\mathrm{w}(\phi,\psi) = \min_{\mathbf{T}\in \Pi(u,v)}\sum_{i} \sum_{j} \mathbf{T}_{ij} \cdot c(x_i,y_j)
\end{equation}

where $\Pi(u,v) = \{ \mathbf{T} \in \mathbb{R}_+^{n\times m} | \mathbf{T}\mathbf{1}_m=u, \mathbf{T}^\top\mathbf{1}_n=v \}$, $c(x_i,y_j)$ is cosine distance metric, and $\mathbf{T}$ is the transport plan, interpreting the amount of mass shifted from $\phi_i$ to $\psi_j$.

\noindent \textbf{Gromov-Wasserstein Distance} assists in edge matching and preserves graph topology by calculating distances between pairs of nodes in each domain and measuring how these distances compare to the counter domain. In the same discrete graph matching setting, GWD between $\phi$ and $\psi$ can be mathematically represented as:
\begin{equation}\label{eq:gwd}
\vspace{-1mm}
	\mathcal{D}_\mathrm{gw}(\phi,\psi) = \min_{\hat{\mathbf{T}}\in \Pi(u,v)}\sum_{i,i',j,j'} \hat{\mathbf{T}}_{ij} \hat{\mathbf{T}}_{i'j'} L(x_i,y_j,x_i',y_j')
\end{equation}
where intra-graph structural similarity between two node pairs $(x_i,x_i')$ and $(y_j,y_j')$ is represented as $L(x_i,y_j,x_i',y_j')=\|c_1(x_i,x_i') - c_2(y_i,y_i')\|$, $c_i$  being cosine similarity between a node pair in any graph $\mathcal{G}_{i}$.  Transport plan $\hat{\mathbf{T}}$ is periodically updated to align the edges in different graphs belonging to disparate modalities.



We further follow \citet{chen2020graph} to combine WD and GWD transport plans, leading to a unified GOT objective given as:
\begin{equation} \label{eq:got}
\vspace{-2mm}
    \mathcal{L}_\mathrm{GOT}(\phi,\psi) = \gamma\mathcal{D}_\mathrm{w}(\phi,\psi) + (1 - \gamma) \mathcal{D}_\mathrm{gw}(\phi,\psi)
\end{equation}

where $\gamma$ regulates the importance of two loss terms.

\subsection{Cross-Modal Attention Fusion (CMAF)}

BT and GOT losses are computed in a dual encoder setting, which does not contain cross-modal interactions and is not suitable for complex multi-modal feature representation. Most existing methods, including UNITER \citep{chen2020uniter}, ViLT \citep{kim2021vilt}, METER \citep{dou2022empirical}, and GLIP \citep{li2022grounded} design cross-modal fusion by stacking additional transformer layers on top of uni-modal encoders, introducing a large number of added parameters during pre-training. We follow a more efficient solution proposed by FIBER \citep{dou2022coarse}, which inserts cross-modal fusion into the uni-modal backbones with a gating mechanism. Specifically, at the top $M$ transformer layers in the vision and language backbone, cross-attention signals, weighted by a gating scalar $\alpha$, are added to self-attention:
\begin{align}\label{eq:cross_attention}
\vspace{-3mm}
    \hat{x} &= \text{Self-Att}(x) \nonumber \\
    x &= x + \hat{x} + \alpha * \text{Cross-Att}(\hat{x}, y)\\
    x &= x + \text{FFN}(x) \nonumber
\end{align}
where $\alpha$ is a trainable parameter initialized to $0$. Following existing literature \citep{li2021align, wang2021vlmo, dou2022empirical, dou2022coarse}, we use masked language modeling (MLM) and image-text matching (ITM) to pre-train the cross-attention parameters. For MLM, we randomly mask $15\%$ text tokens, and the loss aims to reconstruct the masked tokens. We feed the network with randomly sampled image-caption pairs for ITM, and the loss predicts whether they are matched. The gating mechanism is a good choice for CMAF because $(i)$ cross-attention parameters can easily be switched off by setting the gating scalar $\alpha$ to $0$ when computing the BT and GOT losses. Thus, we can learn the cross-attention parameters without affecting the original computational flow of uni-modal backbones. $(ii)$ gating mechanism is more lightweight and memory-efficient than adding fusion-specific layers (GLIP and METER use 4$\times$ more fusion parameters than \model).

Overall, \model\ training pipeline can be summarized in the following three steps:

\begin{itemize}[leftmargin=*]
\vspace{-2mm}
    \item \textbf{BT \& GOT:} CMAF is switched off $(\alpha = 0)$, \model\ acts as dual encoder, $\mathcal{L}_\mathrm{BT}$ and $\mathcal{L}_\mathrm{GOT}$ are computed. 
    \item \textbf{MLM \& ITM}: CMAF is switched on $(\alpha \neq 0)$, \model\ now acts as fusion encoder, $\mathcal{L}_\mathrm{MLM}$ and $\mathcal{L}_\mathrm{ITM}$ losses are computed.
    \item \textbf{Back-propagation:} the four losses are added, giving $\mathcal{L}_\mathrm{total} = \mathcal{L}_\mathrm{BT} + w_\mathrm{GOT}*\mathcal{L}_\mathrm{GOT} + \mathcal{L}_\mathrm{MLM} + \mathcal{L}_\mathrm{ITM}$, and back-propagated into the model end-to-end. An ablation on different pre-training objectives of \model\ and values of $w_\mathrm{GOT}$ is given in Section \ref{sec:albation_study}. 
    
\end{itemize}

The overall \model\ pipeline for computation of different training objectives is shown in Figure \ref{fig:system_overview}. The pseudo-code for \model\ is presented in Appendix \ref{sec:pseudo}.

\subsection{Finetuning For Downstream Tasks}

We adopt \model\ to various vision- and vision-language downstream tasks. We switch off the inserted cross-attention modules for the vision-only tasks and use the image encoder. We utilize the learned cross-attention parameters as required for the vision-language tasks, following \citet{dou2022coarse}. For example, VQA and visual reasoning employ all cross-attention modules, whereas captioning requires only image-to-text cross-attention. Again, during IRTR, we switch off all cross-attentions and use \model\ in a dual encoder setting. We keep all cross-attention parameters during multi-modal object detection and referring expression comprehension and train an object detection head from scratch using the language-aware image features.

\section{Experiments, Results, and Analysis}

\subsection{Pre-training \& Downstream datasets} 

Following \citet{chen2020uniter} and \citet{huang2021seeing}, we perform pre-training by appending the VG dataset \citep{krishna2017visualgenome} with COCO$2017$ \citep{lin2014microsoft}, together consisting of $231$k images. We divide our downstream tasks into three categories - $(i)$ \textbf{Uni-modal tasks} such as image classification on ImageNet \citep{deng2009imagenet}, VOC$07$ \citep{everingham2010pascal}, COCO; object detection on VOC$07+12$, COCO, and instance segmentation on COCO. $(ii)$ \textbf{Multi-modal fine-grained tasks} such as region-level VL tasks - referring expression comprehension (REC) on RefCOCO, RefCOCO$+$, RefCOCOg \citep{kazemzadeh2014referitgame, yu2016modeling}, and language-conditioned object detection on COCO and LVIS \citep{gupta2019lvis}. $(iii)$ \textbf{Multi-modal coarse-grained tasks} such as image-level VL tasks - visual question answering on VQAv$2$ \citep{antol2015vqa}, visual reasoning on NLVR$^{2}$ \citep{suhr2019corpus}, image- and text retrieval on Flicker$30$k \citep{plummer2015flickr30k} and captioning on COCO. We exclude any overlap between our pre-training and downstream validation/test splits. Detailed statistics of all downstream datasets are given in Appendix \ref{sec:downstream_datasets}. 

\subsection{Network Architectures} 

Following FIBER \citep{dou2022coarse}, we adopt Swin-Base \citep{liu2021swin} and RoBERTa-Base \citep{liu2019roberta} as our vision and text encoders, which are initialized with weights from uni-modal pre-training. We collect patch- and token features from the last transformer layers, feed them into the local projector network, and compute GOT loss. Furthermore, we apply \texttt{AvgPool} on patch and token features, feed them into the global projector network, and compute BT loss. Both local and global projector networks have three linear layers with dimensions $2048$-$2048$-$1024$, with batch normalization and ReLU after the first two layers. Section \ref{sec:albation_study} gives an ablation on projector dimension. We use the image and text features after the \texttt{AvgPool} layer during downstream tasks. For CMAF, we insert the cross-attention into the top 6 blocks of the vision and text encoders. Moreover, for direct comparison with existing uni-modal baselines, we re-train \model\ with ResNet$50$ \citep{he2016deep} and Swin-Tiny image encoders.

\begin{table*}[!t]
\centering
\small
\vspace{2mm}
\setlength{\tabcolsep}{4pt}
\resizebox{1\textwidth}{!}{\begin{tabular}{l | c c c | c | l | c c | c c | c}
\hline
\multicolumn{5}{c |}{\bf Linear probing on ImageNet Validation Set} & \multicolumn{6}{c}{\bf Linear probing on VOC$07$ and COCO}\\

\hline
\multirow{2}{*}{\bf Method} & \multirow{2}{*}{\bf Pre-train} & \multirow{2}{*}{\bf Arch.} & \multirow{2}{2.45 cm}{\bf \centering Pre-training Supervision }& \multirow{2}{*}{\bf Top-$1$} & \multirow{2}{*}{\bf Method} & \multirow{2}{*}{\bf \centering Pre-train} & \multirow{2}{*}{\bf Arch.} & \multicolumn{2}{c |}{\bf VOC$07$} & \bf COCO \\

\cline{9-11}
& & & & & & & & \bf SVM & \bf MLP & \bf MLP (PC/O) \\
\hline
\demph{Sup.} & \demph{IN-$1$K} & \demph{RN$50$} & \demph{Label} & \demph{76.5} & Sup. & IN-$1$K & RN$50$ & 87.5 & 90.8 & 55.2/60.8 \\
Sup. & IN-100 & RN$50$ & Label & 53.3$^\dagger$ & BYOL & IN-$1$K & RN$50$ & 86.6 & $-$ & $-$ \\
\cline{1-5}

MoCo & COCO & RN$50$ & NA & 44.5$^\dagger$ & BT & IN-$1$K & RN$50$ & 86.2 & 91.9$^\ddagger$ & 56.1/63.0$^\ddagger$ \\
MoCo-v2 & COCO & RN$50$ & NA & 49.3$^\dagger$ & VICReg & IN-$1$K & RN$50$ & 86.6 & 91.1$^\ddagger$ & 51.0/57.9$^\ddagger$ \\
\cline{6-11}

CAST & COCO & RN50 & Caption & 48.7 & CAST & COCO &	RN50 & 74.0 & $-$ & 51.0/57.9 \\
VirTex & COCO & RN$50$ & Caption & 52.8 & VirTex & COCO	& RN50 & 88.7 & $-$ & $-$ \\
ICMLM & COCO & RN$50$ & Caption & 51.9 & ICMLM & COCO & RN50 & 87.5 & $-$ & $-$ \\
MCT & COCO & RN$50$ & Caption & 54.9 & LocTex & COCO & RN50 & 88.4 & $-$ & $-$ \\
\demph{MCT} & \demph{COCO} & \demph{RN$50$} & \demph{Caption$+$Tag} & \demph{55.3} & \demph{LocTex}	& \demph{COCO$+$OpenIm} & \demph{RN50} & \demph{92.6} & \demph{$-$} & \demph{$-$} \\

\rowcolor{Light}
\model (w/o MLM, ITM) & COCO & RN$50$ & Caption & 55.3 & \model (w/o MLM, ITM) & COCO & RN$50$ & 89.6 & 94.3 & 71.4/74.3\\
\rowcolor{Light}
\model (w/o MLM, ITM) & COCO & Swin-T & Caption & 56.3 & \model (w/o MLM, ITM) & COCO & Swin-T & 88.2 & 93.5 & 73.4/75.7\\
\rowcolor{Light}
\model (w/o MLM, ITM) & COCO & Swin-B & Caption & \bf 62.5 & \model (w/o MLM, ITM) & COCO & Swin-B & 88.5 & 93.9 & 74.1/76.1\\
\rowcolor{Light}
\model & COCO & Swin-B & Caption & \bf 62.5 & \model & COCO & Swin-B & \bf 89.7 & \bf 95.0 & \bf 74.5/76.4\\
\hline
\end{tabular}}
\caption{\textbf{Uni-modal downstream: linear image classification.} We benchmark learned representations on image classification tasks by training linear classifiers on fixed features. We report top-$1$ accuracy on ImageNet-$1$k validation set, classification mAP on VOC$07$, and per-class (PC) and overall (O) F$1$ scores on COCO. Numbers with $\dagger$ are re-implemented by \cite{yuan2021multimodal}, and the numbers with $\ddagger$ are re-implemented by us. methods trained with significantly larger datasets are colored gray. The best results are in \textbf{bold}.}
\label{tab:imagenet}
\vspace{-3mm}
\end{table*}

\begin{table}[!t]
\centering
\small
\vspace{2mm}
\setlength{\tabcolsep}{4pt}
\resizebox{0.8\textwidth}{!}{\begin{tabular}{@{}l | c c c | c c c | c c c | c c c@{}}
\hline

\multirow{2}{*}{\bf Method} & \multirow{2}{*}{\bf \centering Pre-train} & \multirow{2}{*}{\bf Arch.} & \multirow{2}{2.3 cm}{\bf \centering Pre-training Supervision} & \multicolumn{3}{c |}{\bf VOC$07$+$12$ det} & \multicolumn{3}{c |}{\bf COCO det} & \multicolumn{3}{c}{\bf COCO instance seg} \\ 

\cline{5-13}

& & & & \bf AP$_{all}$ & \bf AP$_{50}$ & \bf AP$_{75}$ & \bf AP$^{bb}$ & \bf AP$^{bb}_{50}$ & \bf AP$^{bb}_{75}$ & \bf AP$^{mk}$ & \bf AP$^{mk}_{50}$ & \bf AP$^{mk}_{75}$  \\

\hline
Sup. & IN-$1$K & RN$50$ & Label & 53.5 & 81.3 & 58.8 & 38.2 & 58.2 & 41.2 & 33.3 & 54.7 & 35.2 \\
\demph{MoCo-v$2$}$^{\dagger}$ & \demph{IN-$1$K} & \demph{RN$50$} & \demph{NA} & \demph{57.4} & \demph{82.5} & \demph{64.0} & \demph{39.3} & \demph{58.9} & \demph{42.5} & \demph{34.4} & \demph{55.8} & \demph{36.5} \\
\hline 
MoCo & COCO & RN$50$ & NA & 47.5 & 75.4 & 51.1 & 38.5 & 58.5 & 42.0 & 35.0 & 55.6 & 37.5 \\
MoCo-v$2$ & COCO & RN$50$ & NA & 48.4 & 75.5 & 52.1 & 39.8 & 59.6 & 43.1 & 35.8 & 56.9 & 38.8\\
CAST & COCO & RN50 & Caption & 54.2 & 80.1 & 59.9 & 39.4 & 60.0 & 42.8 & 35.8 & 57.1 & 38.6 \\
VirTex & COCO & RN50 & Caption & 55.6 & 81.4 & 61.5 & 40.9 & 61.7 & 44.8 & 36.9 & 58.4 & 39.7 \\
LocTex & COCO & RN50 & Caption & 53.9 & 80.9 & 59.8 & 40.6 & 60.6 & 44.1 & 35.2 & 57.0 & 37.4 \\
MCT & COCO & RN50 & Caption & 56.1 & 82.1 & 62.4 & 41.1 & \bf 61.8 & \bf 44.9 & \bf 36.9 & 58.2 & 40.0 \\
\rowcolor{Light}
\model & COCO & RN50 & Caption & \bf 56.6 & \bf 84.4 & \bf 62.7 & \bf 41.9 & \bf 61.8 & 44.8 & 36.5 & \bf 58.5 & \bf 40.8 \\
\hline 
\demph{Sup.}$^{\ddagger}$ & \demph{IN-1K} & \demph{Swin-T} & \demph{Label} & \demph{$-$} & \demph{$-$} & \demph{$-$} & \demph{50.5} & \demph{69.3} & \demph{54.9} & \demph{43.7} & \demph{66.6} & \demph{47.1} \\
\demph{MoBY}$^{\ddagger}$ & \demph{IN-1K} & \demph{Swin-T} & \demph{NA} & \demph{$-$} & \demph{$-$} & \demph{$-$} & \demph{50.2} & \demph{68.8} & \demph{54.7} & \demph{43.5} & \demph{66.1} & \demph{46.9} \\
\rowcolor{Light}
\model & COCO & Swin-T & Caption & $-$ & $-$ & $-$ & \bf 50.9 & \bf 69.6 & \bf 55.5 & \bf 43.8 & \bf 66.9 & \bf 47.5 \\
\hline 
\demph{Sup.}$^{\ddagger}$ & \demph{IN-1K} & \demph{Swin-B} & \demph{Label} & \demph{$-$} & \demph{$-$} & \demph{$-$} & \demph{51.9} & \demph{70.9} & \demph{56.5} & \demph{45.0} & \demph{68.4} & \demph{48.7} \\
ViTDet & IN-1K & ViT-B & NA	& $-$ & $-$ & $-$ & 51.6 & $-$ & $-$ & \bf 45.9 & $-$ & $-$ \\
CLIP & LAION-20M & ViT-B & Caption & $-$ & $-$ & $-$ & 45.2 & $-$ & $-$ & 40.4 & $-$ & $-$ \\
SLIP & LAION-20M & ViT-B & Caption & $-$ & $-$ & $-$ & 44.7 & $-$ & $-$	& 41.0 & $-$ & $-$ \\
MaskCLIP & LAION-20M & ViT-B & Caption & $-$ & $-$ & $-$ & 46.6 & $-$ & $-$ & 41.7 & $-$ & $-$ \\
\rowcolor{Light}
VoLTA & COCO & Swin-B & Caption	& $-$ & $-$ & $-$ & \bf 52.1 & \bf 71.3 & \bf 56.6 & 45.2 & \bf 68.5 & \bf 49.0 \\

\hline
\end{tabular}}
\caption{\textbf{Uni-modal downstream: object detection and instance segmentation \textcolor{black}{with fine-tuning}.} We benchmark learned representations on VOC$07+12$ object detection task using faster R-CNN \citep{ren2015faster}, and on COCO$2017$ object detection and instance segmentation using mask R-CNN \citep{he2017mask}, both with C$4$ backbone variant \citep{wu2019detectron2}. The best results are in \textbf{bold}. \textcolor{black}{Methods marked with $\dagger$ and $\ddagger$ are not direct comparison to other baselines as they use multiple MLP layers \& significant data augmentations and cascade mask R-CNN \citep{cai2018cascade} during fine-tuning, respectively.}}
\label{tab:voc_coco_detection_segmentation}
\vspace{-3mm}
\end{table}

\vspace{-2mm}
\subsection{Implementation Details} 

We perform pre-training for $20$ epochs with $256$ batch-size on $64$ V$100$ GPUs. Following \citet{zbontar2021barlow}, we use the LARS optimizer \citep{you2017large} with a learning rate of $0.2$ for the weights and $0.0048$ for the biases and batch normalization parameters. We use a learning rate warm-up period of $2$ epochs, after which we reduce the learning rate by a factor of $1000$ using a cosine decay schedule \citep{loshchilov2016sgdr}.  We use $1$e$-6$ weight decay, excluding the biases and batch normalization parameters. We conduct a grid search for the GOT loss hyperparameter ($w_\mathrm{GOT}$), and we empirically found the best value to be $100$. Appendix \ref{sec:hyperparameter_values} explains other necessary pre-training and downstream hyper-parameters details.

\subsection{Results on Vision-only tasks}

We first experiment on three uni-modal tasks - classification, object detection, and instance segmentation. For a direct comparison with existing ResNet$50$ and Swin-T baselines, we re-train identical encoders with \model\ pipeline. Furthermore, since the uni-modal tasks do not utilize cross-attention parameters, we perform an ablation by dropping the MLM and ITM objectives from \model.

\noindent \textbf{Image Classification:} Table \ref{tab:imagenet} presents the linear probing results of uni-label classification on ImageNet and multi-label classification on VOC$07$ and COCO. \textcolor{black}{For all uni-modal tasks, we report results with COCO pre-training for a fair comparison with existing baselines.} For ImageNet, we adopt all COCO baselines from \citet{yuan2021multimodal}. Even without the MLM and ITM objectives, \model\ achieves better performance than all baselines across three datasets with ResNet$50$ backbone. The Swin backbones and cross-attention module further improve the performance. For VOC$07$, we report the results for both SVM and MLP-based linear classifiers. \model\ with ResNet$50$ backbone achieves state-of-the-art results on VOC$07$ SVM evaluation, beating the nearest baseline, SwAV, by $0.7$ mAP score. These results indicate the ability of \model\ to learn effective image-level visual features.

\begin{table}[!t]
\centering
\small
\vspace{2mm}
\setlength{\tabcolsep}{4pt}
\resizebox{0.65\textwidth}{!}{\begin{tabular}{@{}l | c c | c c c | c c c | c c@{}}
\hline

\multirow{2}{*}{\bf Method} & \multicolumn{2}{c |}{\bf \centering \#Pre-train Data} & \multicolumn{3}{c |}{\bf RefCOCO} & \multicolumn{3}{c |}{\bf RefCOCO+} & \multicolumn{2}{c}{\bf RefCOCOg}  \\ 

\cline{2-3}\cline{4-6}\cline{7-9}\cline{10-11}

& \bf I-T & \bf I-T-B & \bf val & \bf testA & \bf testB & \bf val & \bf testA & \bf testB & \bf val & \bf test  \\

\hline

MAttNet & $-$ & $-$ & 76.4 & 80.4 & 69.3 & 64.9 & 70.3 & 56.0 & 66.7 & 67.0 \\
\textcolor{black}{VLBERT} & $3$M & $-$ & $-$ & $-$ & $-$ & 71.6 & 77.7 & 61.0 & $-$ & $-$ \\
ViLBERT & 3M & $-$ & $-$ & $-$ & $-$ & 72.3 & 78.5 & 62.6 & $-$ & $-$ \\
Ernie-VL-L & $4$M & {$-$} & {$-$} & {$-$} & {$-$} & {75.9} & {82.4} & {66.9} & {$-$} & {$-$} \\
{Rosita} & {$4$M} & {$-$} & {84.8} & {88.0} & {78.3} & {76.1} & {82.0} & {67.4} & {78.2} & {78.3} \\
{UNITER-L} & {$4$M} & {$-$} & {81.4} & {87.0} & {74.2} & {75.9} & {81.5} & {66.7} & {74.9} & {75.8}\\
{VILLA-L} & {$4$M} & {$-$} & {82.4} & {87.5} & {74.8} & {76.2} & {81.5} & {66.9} & {76.2} & {76.7}\\

\hline
\multicolumn{11}{l}{{\demph{\it{Models pre-trained on Im-Txt-Box data}}}}\\
\hline

\demph{MDETR-B} & $-$ & \demph{$1.3$M} & \demph{87.5} & \demph{90.4} & \demph{82.7} & \demph{81.1} & \demph{85.5} & \demph{73.0} & \demph{83.4} & \demph{83.3}\\
\demph{UniTAB} & $-$ & \demph{$1.3$M} & \demph{86.3} & \demph{88.8} & \demph{80.6} & \demph{78.7} & \demph{83.2} & \demph{69.5} & \demph{80.0} & \demph{80.0}\\
\demph{X-VLM} & \demph{$4$M} & \demph{$6.15$M} & \demph{-} & \demph{-} & \demph{-} & \demph{80.2} & \demph{86.4} & \demph{71.0} & \demph{-} & \demph{-}\\
\demph{OFA-L} & \demph{$16$M} & \demph{$3$M} & \demph{90.1} & \demph{92.9} & \demph{85.3} & \demph{84.5} & \demph{90.1} & \demph{77.8} & \demph{84.5} & \demph{85.2}\\
\demph{FIBER-B} & \demph{$4$M} & \demph{$0.8$M} & \demph{90.7} & \demph{92.6} & \demph{87.3} & \demph{85.7} & \demph{90.1} & \demph{79.4} & \demph{87.1} & \demph{87.3}\\

\hline
\rowcolor{Light}
\textcolor{black}{\model-B} & \textcolor{black}{$231$k} & \textcolor{black}{$-$} & \textcolor{black}{\bf 86.1} & \textcolor{black}{\bf 88.6} & \textcolor{black}{\bf 81.8} & \textcolor{black}{\bf 77.0} & \textcolor{black}{\bf 82.7} & \textcolor{black}{\bf 67.8} & \textcolor{black}{\bf 78.3} & \textcolor{black}{\bf 78.3} \\

\hline
\end{tabular}}
\caption{\textbf{Multi-modal fine-grained downstream: referring expression comprehension.} Methods pre-trained on image-text-box (I-T-B) data are colored gray. Best comparable results are in \textbf{bold}. \model-B denotes Swin-B backbone.} 
\label{tab:refcoco}
\vspace{-5mm}
\end{table}

\begin{table}[!t]
\centering
\vspace{2mm}
\small
\setlength{\tabcolsep}{4pt}
\resizebox{0.40\columnwidth}{!}{\begin{tabular}{@{}l | c | c c c c @{}}
\hline

\multirow{2}{*}{\bf Method} & \multicolumn{1}{c |}{\bf COCO Val 2017} & \multicolumn{4}{c}{\bf LVIS MiniVal}  \\ 

\cline{2-2}\cline{3-6}

& \bf AP & \bf APr & \bf APc & \bf APf & \bf AP  \\

\hline
\multicolumn{6}{l}{{\demph{\it{Models pre-trained on Im-Txt-Box and/or with larger size}}}}\\
\hline
\demph{Mask R-CNN} & \demph{$-$} & \demph{26.3} & \demph{34.0} & \demph{33.9} & \demph{33.3} \\
\demph{MDETR} & \demph{$-$} & \demph{20.9} & \demph{24.9} & \demph{24.3} & \demph{24.2} \\
\demph{GLIP-B} & \demph{57.0} & \demph{31.3} & \demph{48.3} & \demph{56.9} & \demph{51.0} \\
\demph{GLIP-L} & \demph{60.8} & \demph{$-$} & \demph{$-$} & \demph{$-$} & \demph{$-$} \\
\demph{FIBER-B} & \demph{58.4} & \demph{50.0} & \demph{56.9} & \demph{58.1} & \demph{56.9} \\
\hline
\rowcolor{Light}
\textcolor{black}{\model-B} & \textcolor{black}{\bf 51.6} & \textcolor{black}{\bf 34.4} & \textcolor{black}{\bf 43.1} & \textcolor{black}{\bf 43.8} & \textcolor{black}{\bf 42.7} \\

\hline
\end{tabular}}
\caption{\textbf{Multi-modal fine-grained downstream: language-conditioned object detection on COCO and LVIS.} All available baselines are pre-trained on Im-Txt-Box data and are colored gray. \model-B denotes Swin-B backbone.}
\label{tab:coco_lvis}
\vspace{-4mm}

\end{table}

\noindent \textbf{Object Detection \& Instance Segmentation:} Next, we perform two uni-modal region-level tasks - object detection on VOC$07+12$ and COCO$2017$, and instance segmentation on COCO$2017$. As shown in Table \ref{tab:voc_coco_detection_segmentation}, \model\ yields the state-of-the-art performance in both tasks across the majority of metrics. \textcolor{black}{The fine-grained region-level understanding helps \model\ to perform well on detection and segmentation tasks.}

\subsection{Results on Fine-grained Vision-Language tasks}

Next, we perform region-level multi-modal downstream tasks - referring expression comprehension (REC) and language-guided object detection.  

\noindent \textbf{REC:} This task aims to localize target objects in an image described by a referring expression phrased in natural language and, thus, perfectly evaluates the fine-grained feature representation capability of \model. As depicted in Table \ref{tab:refcoco}, \model\ beats larger-sized UNITER-L and VILLA-L models on the challenging testB split of both RefCOCO and RefCOCO$+$. Moreover, \model\ performs comparably with MDETR and UniTAB, even without being trained on grounding data. These results indicate our model's efficacy in learning fine-grained local visual features.  

\noindent \textbf{Object Detection:} We evaluate \model\ on two challenging language-conditioned object detection benchmarks - COCO and LVIS. Note that, all existing baselines for this tasks are pre-trained on fine-grained \textit{image-text-box} data, whereas \model\ only utilizes \textit{image-caption} pairs. Table \ref{tab:coco_lvis} shows that \model\ performs comparatively with these strong baselines. Note that \model\ beats Mask R-CNN, MDETR, and GLIP-B on LVIS APr, which denotes average precision on rare objects. Thus, we conclude that \model\ achieves impressive localization ability and robustness, even without utilizing any grounding annotations. 

\begin{table*}[!t]
\centering
\vspace{-2mm}
\vspace{2mm}
\small
\setlength{\tabcolsep}{4pt}
\resizebox{0.85\textwidth}{!}{\begin{tabular}{@{}l | c | c c | c c | c c | l | c | c c c c@{}}
\hline
\multirow{2}{*}{\bf Method} & \multirow{2}{2 cm}{\bf \centering \#Pre-train Data} & \multicolumn{2}{c |}{\bf VQAv$2$} & \multicolumn{2}{c |}{\bf NLVR$^{2}$} & \multicolumn{2}{c |}{\bf F$30$k IRTR} & \multirow{2}{*}{\bf Method} & \multirow{2}{2 cm}{\bf \centering \#Pre-train Data} & \multicolumn{4}{c}{\bf COCO Captioning}\\ 

\cline{3-4}\cline{5-6}\cline{7-8}\cline{11-14}

& & \bf dev & \bf std & \bf dev & \bf test-P & \bf IR@$1$ & \bf TR@$1$ & & & \bf B$@4$ & \bf M & \bf C & \bf S \\

\hline
\multicolumn{8}{l |}{{\it{Models pre-trained on COCO ($123$k) and/or VG ($108$k)}}} & \multicolumn{6}{l}{{\it{Models fine-tuned without CIDEr optimization}}}\\
\cline{1-14}
SCAN & $108$k & $-$ & $-$ & $-$ & $-$ & 48.6 & 67.4 & VirTex & $123$k & $-$ & $-$ & 95.5 & 18.1 \\
SCG & $108$k & $-$ & $-$ & $-$ & $-$ & 49.3 & 71.8  & VL-T5 & $180$k & 34.5 & 28.7 & 116.5 & 21.9\\
PFAN & $108$k & $-$ & $-$ & $-$ & $-$ & 50.4 & 70.0 & VL-BART & $180$k & 35.1 & 28.7 & 116.6 & 21.5\\

MaxEnt & $123$k & 54.1 & 54.8 & $-$ & $-$ & $-$ & $-$ & \textcolor{black}{\CC{}\model-B} & \textcolor{black}{\CC{}$231$k} & \textcolor{black}{\CC{}38.2} & \textcolor{black}{\CC{}30.7} & \textcolor{black}{\CC{}126.6} & \textcolor{black}{\CC{}22.5} \\
\cline{9-14}
VisualBERT & $123$k & 70.8 & 71.0 & 67.4 & 67.0 & $-$ & $-$ & \textcolor{black}{\CC{}\model-GOLD-B} & \textcolor{black}{\CC{}$231$k} & \textcolor{black}{\CC{}38.9} & \textcolor{black}{\CC{}30.5} & \textcolor{black}{\CC{}128.5} & \textcolor{black}{\CC{}23.4} \\
\cline{9-14}
LXMERT & $231$k & 72.4 & 72.5 & 74.9 & 74.5 & $-$ & $-$ & \multicolumn{6}{l}{{\it{Models fine-tuned with CIDEr optimization}}}\\
\textcolor{black}{SOHO} & \textcolor{black}{$231$k} & \textcolor{black}{73.2} & \textcolor{black}{73.4} & \textcolor{black}{76.3} & \textcolor{black}{77.3} & \textcolor{black}{72.5} & \textcolor{black}{\bf 86.5} & \textcolor{black}{\CC{}\model-B} & \textcolor{black}{\CC{}$231$k} & \textcolor{black}{\CC{}39.7} & \textcolor{black}{\CC{}30.5} & \textcolor{black}{\CC{}133.6} & \textcolor{black}{\CC{}23.7}\\
\hline
\rowcolor{Light}
\textcolor{black}{\model-B} & \textcolor{black}{$231$k} & \textcolor{black}{\bf 74.6} & \textcolor{black}{\bf 74.6} & \textcolor{black}{\bf 76.7} & \textcolor{black}{\bf 78.1} & \textcolor{black}{\bf 72.7} & \textcolor{black}{83.6} & \textcolor{black}{\model-GOLD-B} & \textcolor{black}{$231$k} & \textcolor{black}{\bf 40.2} & \textcolor{black}{\bf 30.9} & \textcolor{black}{\bf 137.5} & \textcolor{black}{\bf 23.7} \\
\hline
\end{tabular}}
\caption{\textbf{Multi-modal coarse-grained downstream: visual question answering, visual reasoning, retrieval, and captioning.} We only compare with methods pre-trained on a comparable amount of dataset. For captioning, $4$ metrics are reported - B$@4$: BLEU$@4$, M: METEOR, C: CIDEr, S: SPICE. The best results are in \textbf{bold}. \model-B denotes Swin-B backbone.}
\label{tab:vqa_nlvr_irtr_captioning}
\vspace{-3.5mm}
\end{table*}

\begin{table}[!t]
\centering
\vspace{2mm}
\small
\setlength{\tabcolsep}{4pt}
\resizebox{0.675\columnwidth}{!}{\begin{tabular}{@{}c c c c c | c c c | c c c |c c @{}}
\hline

\multicolumn{5}{c |}{\textcolor{black}{\bf \model}} & \multicolumn{3}{c |}{\textcolor{black}{\bf RefCOCO}} & \multicolumn{3}{c |}{\textcolor{black}{\bf RefCOCO$+$}} & \multicolumn{2}{c }{\textcolor{black}{\bf RefCOCOg}} \\ 

\hline

\bf \multirow{2}{*}{\textcolor{black}{$\mathcal{L}_\mathrm{BT}$}} & \multicolumn{2}{c}{\textcolor{black}{\bf $\mathcal{L}_\mathrm{GOT}$}} & \bf \multirow{2}{*}{\textcolor{black}{$\mathcal{L}_\mathrm{MLM}$}} & \bf \multirow{2}{*}{\textcolor{black}{$\mathcal{L}_\mathrm{ITM}$}} & \multirow{2}{*}{\textcolor{black}{\bf val}} & \multirow{2}{*}{\textcolor{black}{\bf testA}} & \multirow{2}{*}{\textcolor{black}{\bf testB}} & \multirow{2}{*}{\textcolor{black}{\bf val}} & \multirow{2}{*}{\textcolor{black}{\bf testA}} & \multirow{2}{*}{\textcolor{black}{\bf testB}} & \multirow{2}{*}{\textcolor{black}{\bf val}} & \multirow{2}{*}{\textcolor{black}{\bf test}} \\

& \textcolor{black}{$\mathcal{L}_\mathrm{w}$} & \textcolor{black}{$\mathcal{L}_\mathrm{gw}$} & & & & & & \\
\hline

\textcolor{black}{\ding{51}} & \textcolor{black}{$-$} & \textcolor{black}{$-$} & \textcolor{black}{$-$} & \textcolor{black}{$-$} & \textcolor{black}{81.7} & \textcolor{black}{84.1} & \textcolor{black}{77.8} & \textcolor{black}{71.2} & \textcolor{black}{76.6} & \textcolor{black}{62.2} & \textcolor{black}{71.7} & \textcolor{black}{71.7} \\
$-$ & $-$ &	$-$	& \ding{51} & \ding{51} & 82.0 & 84.5 & 77.8 & 71.5 & 77.1 & 62.2 & 71.4 & 71.8 \\
\textcolor{black}{\ding{51}} & \textcolor{black}{$-$} & \textcolor{black}{$-$} & \textcolor{black}{\ding{51}} & \textcolor{black}{\ding{51}} & \textcolor{black}{82.7} & \textcolor{black}{85.2} & \textcolor{black}{78.1} & \textcolor{black}{72.0} & \textcolor{black}{77.7} & \textcolor{black}{62.5} & \textcolor{black}{72.8} & \textcolor{black}{72.7} \\
\textcolor{black}{\ding{51}} & \textcolor{black}{\ding{51}} & \textcolor{black}{$-$} & \textcolor{black}{\ding{51}} & \textcolor{black}{\ding{51}} & \textcolor{black}{83.9} & \textcolor{black}{86.6} & \textcolor{black}{80.5} & \textcolor{black}{73.9} & \textcolor{black}{79.5} & \textcolor{black}{64.1} & \textcolor{black}{74.6} & \textcolor{black}{74.3} \\
\ding{51} &	$-$	& \ding{51}	& \ding{51} & \ding{51} & 82.8 & 85.5 & 78.5 & 72.2 & 77.8 & 62.8 & 72.9 & 72.8 \\
\rowcolor{Light}
\ding{51} & \ding{51} & \ding{51} & \ding{51} & \ding{51} & \bf 86.1 & \bf 88.6 & \bf 81.8 & \bf 77.0 & \bf 82.7 & \bf 67.8 & \bf 78.3 & \bf 78.3 \\
\hline
\end{tabular}}
\caption{\textcolor{black}{\textbf{Ablation study on different losses of the training objective of \model\ for referring expression comprehension tasks.} Each model is pre-trained on $231$k samples from COCO$2017$ and VG.}}\label{tab:loss_fine_ablation}
\vspace{-2mm}
\end{table}

\subsection{Results on Coarse-grained Vision-Language tasks}

Next, we perform image-level multi-modal downstream tasks - visual question answering (VQA), visual reasoning, retrieval, and captioning. 

\noindent \textbf{VQA \& Visual Reasoning:} As reported in Table \ref{tab:vqa_nlvr_irtr_captioning}, \model\ achieves the best performance on VQA and visual reasoning across the baselines pre-trained with a comparable amount of data. Moreover, on VQA, \model\ beats LXMERT, which is trained with $2\times$ more data. These results demonstrate the efficacy of our method even when utilizing a mid-scale pre-training corpus.  

\noindent \textbf{Retrieval:} Most existing VLP methods use a fusion encoder for image and text retrieval and feed every image-text pair into the model. Though such fine-tuning often results in higher performance, it introduces quadratic time cost and is not scalable. Following \cite{dou2022coarse}, we adopt a more efficient strategy. We drop the cross-attention parameters for this task and compute the dot product of image and text features extracted separately in the dual-encoder setting. As shown in Table \ref{tab:vqa_nlvr_irtr_captioning}, even with such an approach, \model\ produces superior performance among the baselines trained with a similar amount of data, beating all three baselines by a significant margin.


\begin{table}
\centering
\begin{subtable}[c]{0.34\textwidth}
\vspace{-2.5em}
\centering
\small
\setlength{\tabcolsep}{4pt}
\resizebox{1\textwidth}{!}{\begin{tabular}{@{}c c c c | c | c @{}}
\hline

\multirow{2}{*}{\textcolor{black}{$\mathcal{L}_{\mathrm{BT}}^{II'}$}} & \multirow{2}{*}{\textcolor{black}{$\mathcal{L}_{\mathrm{BT}}^{TT'}$}} & \multirow{2}{*}{\textcolor{black}{$\mathcal{L}_{\mathrm{BT}}^{IT'}$}} & \multirow{2}{*}{\textcolor{black}{$\mathcal{L}_{\mathrm{BT}}^{I'T}$}} & \multicolumn{1}{c |}{\textcolor{black}{\bf VOC$07$}} & \multicolumn{1}{c}{\textcolor{black}{\bf COCO}} \\ 

\cline{5-6}

& & & & \textcolor{black}{MLP} & \textcolor{black}{MLP (PC/O)}\\

\hline

\textcolor{black}{\ding{51}} & \textcolor{black}{$-$} & \textcolor{black}{$-$} & \textcolor{black}{$-$} & \textcolor{black}{90.8} & \textcolor{black}{72.2/71.7}\\
\textcolor{black}{\ding{51}} & \textcolor{black}{\ding{51}} & \textcolor{black}{$-$} & \textcolor{black}{$-$} & \textcolor{black}{91.8} & \textcolor{black}{74.0/73.7}\\
$-$ & $-$ & \ding{51} & \ding{51} & \textcolor{black}{86.5} & \textcolor{black}{69.0/69.8} \\
\rowcolor{Light}
\ding{51} & \ding{51} & \ding{51} & \ding{51} & \bf 94.0 & \bf 74.5/76.4\\

\hline

\end{tabular}}

\subcaption{\textbf{Ablation study on Intra- and Inter-modal Barlow Twins objective} for multi-label image classification on VOC$07$ and COCO.}
\label{tab:BT_loss_ablation}
\end{subtable}
\hspace{0.75em}
\begin{subtable}[c]{0.25\textwidth}
\centering
\small
\setlength{\tabcolsep}{4pt}
\resizebox{0.8\textwidth}{!}{\begin{tabular}{@{}c | c | c @{}}
\hline

\multirow{2}{*}{$w_\mathrm{GOT}$} & \multicolumn{1}{c |}{\bf VOC$07$} & \multicolumn{1}{c}{\bf COCO} \\ 

\cline{2-3}

& MLP & MLP (PC/O)\\

\hline

50 & 93.4 & 74.1/76.0\\
\rowcolor{Light}
100 & \bf 94.0 & \bf 74.5/76.4\\
200 & 93.2 & 73.1/75.5\\
500 & 93.1 & 72.8/75.3\\

\hline

\end{tabular}}
\subcaption{\textbf{Ablation study on the value of $w_\mathrm{GOT}$}, the weight of GOT loss in $\mathcal{L}_\mathrm{total}$ in the objective of \model\ for multi-label image classification on VOC$07$ and COCO.}
\label{tab:wgot_classification_ablation}
\end{subtable}
\hspace{0.75em}
\begin{subtable}[c]{0.3\textwidth}
\vspace{-1.4em}
\centering
\small
\setlength{\tabcolsep}{4pt}
\resizebox{0.9475\textwidth}{!}{\begin{tabular}{@{}c | c | c @{}}
\hline

\multirow{2}{*}{Projector Config.} & \multicolumn{1}{c |}{\bf VOC$07$} & \multicolumn{1}{c}{\bf COCO} \\ 

\cline{2-3}

& MLP & MLP (PC/O)\\

\hline

8192-8192-128 & 91.3 & 71.3/73.0\\
8192-8192-256 & 91.9 & 72.2/73.3\\
2048-2048-512 & 93.4 & 73.9/76.1\\
\rowcolor{Light}
2048-2048-1024 & \bf 94.0 & \bf 74.5/76.4\\

\hline

\end{tabular}}
\subcaption{\textbf{Ablation study on the dimension of local and global projector networks} of \model\ for multi-label image classification on VOC$07$ and COCO.}
\label{tab:projector_classification_ablation}
\end{subtable}
\caption{\textbf{Ablation on Intra- and Inter-modal Barlow Twins objective (a), the value of $w_\mathrm{GOT}$ (b), and the dimension of projector networks (c).}  We report classification mAP on VOC$07$, and per-class (PC) and overall (O) F$1$ scores on COCO. Each model is pre-trained on $123$k train-val samples from COCO$2017$.}
\label{tab:wgot_projector_classification_ablation}
\vspace{-3mm}
\end{table}

\noindent \textbf{Captioning:} We perform captioning on the COCO dataset to evaluate if \model\ can adopt a generation task. We integrate GOLD \citep{pang2021text} into \model\ during fine-tuning as it produces significant improvements. As shown in Table \ref{tab:vqa_nlvr_irtr_captioning}, our approach maintains superior captioning performance across all baselines pre-trained with comparable data. Using CIDEr optimization further improves performance.

It is worth mentioning that besides achieving a superior result than all baselines using a comparable amount of data on multi-modal coarse-grained tasks, \model\ also outperforms multiple methods pre-trained using magnitude more data. These results, shown in Table \ref{tab:vqa_nlvr_irtr_captioning_appendix}, indicate the effectiveness and generalizability of \model\ across these tasks.

\subsection{Ablation Study} \label{sec:albation_study}
We perform ablation studies on the pre-training objectives, GOT loss weight, and the dimension of projectors.

\noindent \textbf{Pre-training Objectives:} We ablate the effectiveness of different pre-training objectives and evaluate the pre-trained models on fine-grained downstream tasks. First, we pre-train \model\ only with the multi-modal BT loss. In this setup, \model\ only acts as a dual encoder; thus, the cross-attention parameters are not pre-trained. Next, we add MLM and ITM loss which helps the model to learn cross-modal information via attention fusion. Next, we add the GOT pre-training objective. Note that GOT adopts two types of OT distances $-$ WD for node matching and GWD for edge matching. As shown in table \ref{tab:loss_fine_ablation}, applying WD and GWD together improves the performance of reference expression comprehension across RefCOCO, RefCOCO$+$, and RefCOCOg datasets. Specifically on RefCOCOg, adding $\mathcal{L}_\mathrm{gw}$ to $\mathcal{L}_\mathrm{w}$  yields a significant $4.0\%$ boost in the challenging test set. Since this dataset contains intricate images with multiple similar objects with different shapes and colors, GWD is crucial in distinguishing between them. However, we see that adding GWD without WD is not helpful. This is because though GWD can capture the edge similarity between graphs, it cannot directly address graph alignment since it does not consider node information. For example, the word pair (boy, girl) has a similar cosine similarity as the pair (football, basketball). Still, the semantic meanings of the two pairs are different and should not be matched. But GWD will treat these two pairs as the same since it only considers the cosine similarity between nodes. Hence, when applied together, WD and GWD objectives only result in an effective region-word alignment.

We also verify the effectiveness of the multi-modal BT objective by ablating the intra- and inter-modal terms. The first row of Table \ref{tab:BT_loss_ablation} is identical to the original image-only BT objective. Next, we introduce the text branch and add the same BT objective between the two views of the caption. Afterward, we add the inter-modal BT objectives. As shown in Table \ref{tab:BT_loss_ablation}, each loss term improved the image classification performance, demonstrating the importance of intra- and inter-modal objectives. Overall, this set of experiments demonstrates that all objectives are necessary for our model to perform well on different fine-grained multi-modal tasks.

\noindent \textbf{GOT Loss Weight:} In our loss formulation, we introduce a GOT loss weight $w_\mathrm{GOT}$ which regulates the alignment of local features through GOT loss. By conducting a grid search on uni-modal downstream classification tasks, we assessed the impact of $w_\mathrm{GOT}$ as shown in Table \ref{tab:wgot_classification_ablation} and experimentally found its best value to be 100 in our case. It is to be noted that a very high value of $w_\mathrm{GOT}$ considerably degrades the performance of downstream tasks.

\noindent \textbf{Ablation on Projector Dimension:} The design of the projector head plays a pivotal role in the downstream performance of the model \citep{garrido2022duality}. To investigate the impact of hidden and feature (projector output) dimensions, we have tested $4$ different configurations on uni-modal downstream classification tasks. It can be observed (see Table \ref{tab:projector_classification_ablation}) that an increase in the number of parameters in the projector head does not necessarily lead to an increase in performance. For example, a projector configuration of $8192$-$8192$-$256$ has roughly eight times more parameters than $2048$-$2048$-$1024$. However, the latter performs better in downstream tasks (Table \ref{tab:projector_classification_ablation}), indicating that the output dimension of the projector plays a crucial role in the final performance of the model.

\begin{figure*}[!t]
\centering
\includegraphics[scale=0.35]{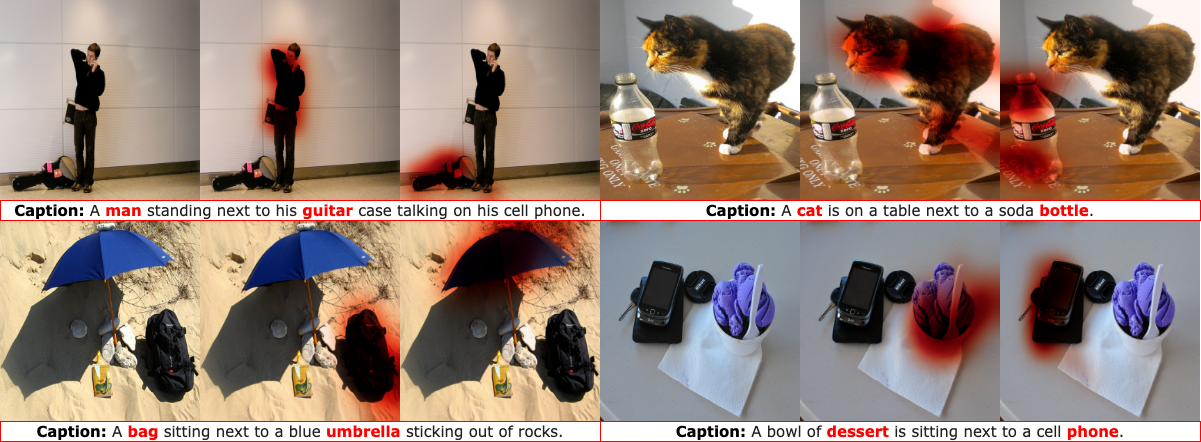}
\vspace{-1mm}
\caption{\textbf{Visualization of fine-grained patch-word alignment, produced by \model.} We look at the optimal transport plan from GOT to represent the alignment between the words in \textcolor{red}{red} with the corresponding input image. All image-caption pairs are taken from the COCO$2017$ train split. The visualizations  are generated with $224$p images, resulting in sequences of $196$ tokens for $16 \times 16$ patches.}
\label{fig:main_visualization}
\vspace{-2mm}
\end{figure*}

\subsection{Qualitative Results \& Error Analysis}

Figure \ref{fig:main_visualization} shows the fine-grained alignment of image regions and caption words achieved by the pre-trained \model\ system. The transport plan from GOT module outputs the similarity across every image patch and caption token. To obtain the visualizations in Figure \ref{fig:main_visualization}, we choose the similarity scores between the \textcolor{red}{red} words in the caption with every image patch. Next, we apply bilinear interpolation to these similarity scores to convert them to the same dimension as the input image. Finally, we superimpose these interpolated similarity maps on the input images to obtain Figure \ref{fig:main_visualization} as the outcome. In most cases, the pre-trained model accurately learns to localize various objects using only global image-caption data. However, objects in extremely cluttered scenarios are occasionally not focused. We show such error cases in Section \ref{sec:error_analysis}.

\section{Conclusion}

We present \model, a unified VLP paradigm that utilizes image-caption data but achieves fine-grained region-level image understanding, eliminating the use of expensive box annotations. \model\ adopts graph optimal transport-based weakly supervised patch-token alignment and produces an explicit, self-normalized, and interpretable low-level matching criterion. Extensive experiments demonstrate the effectiveness of \model\ on a wide range of coarse- and fine-grained tasks.

\section*{Acknowledgement}
The codebase for this work is built on the Barlow Twins \citep{zbontar2021barlow}, GOT \citep{chen2020graph}, and FIBER \citep{dou2022coarse} repository. We would like to thank the respective authors for their contribution, and the Meta AI team for discussions and feedback. Shraman Pramanick and Rama Chellappa were partially supported by an ONR MURI Grant N00014-20-1-2787.

\bibliography{main}

\begin{thebibliography}{112}
\providecommand{\natexlab}[1]{#1}
\providecommand{\url}[1]{\texttt{#1}}
\expandafter\ifx\csname urlstyle\endcsname\relax
  \providecommand{\doi}[1]{doi: #1}\else
  \providecommand{\doi}{doi: \begingroup \urlstyle{rm}\Url}\fi

\bibitem[Anderson et~al.(2016)Anderson, Fernando, Johnson, and
  Gould]{anderson2016spice}
Peter Anderson, Basura Fernando, Mark Johnson, and Stephen Gould.
\newblock Spice: Semantic propositional image caption evaluation.
\newblock In \emph{European Conference on Computer Vision}, pp.\  382--398.
  Springer, 2016.

\bibitem[Antol et~al.(2015)Antol, Agrawal, Lu, Mitchell, Batra, Zitnick, and
  Parikh]{antol2015vqa}
Stanislaw Antol, Aishwarya Agrawal, Jiasen Lu, Margaret Mitchell, Dhruv Batra,
  C~Lawrence Zitnick, and Devi Parikh.
\newblock Vqa: Visual question answering.
\newblock In \emph{International Conference on Computer Vision}, pp.\
  2425--2433, 2015.

\bibitem[Assran et~al.(2022)Assran, Caron, Misra, Bojanowski, Bordes, Vincent,
  Joulin, Rabbat, and Ballas]{Assran2022MaskedSN}
Mahmoud Assran, Mathilde Caron, Ishan Misra, Piotr Bojanowski, Florian Bordes,
  Pascal Vincent, Armand Joulin, Michael~G. Rabbat, and Nicolas Ballas.
\newblock Masked siamese networks for label-efficient learning.
\newblock \emph{arXiv}, abs/2204.07141, 2022.

\bibitem[Balaji et~al.(2019)Balaji, Chellappa, and Feizi]{balaji2019normalized}
Yogesh Balaji, Rama Chellappa, and Soheil Feizi.
\newblock Normalized wasserstein for mixture distributions with applications in
  adversarial learning and domain adaptation.
\newblock In \emph{International Conference on Computer Vision}, pp.\
  6500--6508, 2019.

\bibitem[Banerjee \& Lavie(2005)Banerjee and Lavie]{banerjee2005meteor}
Satanjeev Banerjee and Alon Lavie.
\newblock Meteor: An automatic metric for mt evaluation with improved
  correlation with human judgments.
\newblock In \emph{Proceedings of the ACL Workshop on Intrinsic and Extrinsic
  Evaluation Measures for Machine Translation and/or Summarization}, pp.\
  65--72, 2005.

\bibitem[Bao et~al.(2021)Bao, Dong, Piao, and Wei]{Bao2022BEiTBP}
Hangbo Bao, Li~Dong, Songhao Piao, and Furu Wei.
\newblock Beit: Bert pre-training of image transformers.
\newblock In \emph{International Conference on Learning Representations}, 2021.

\bibitem[Bardes et~al.(2022)Bardes, Ponce, and LeCun]{Bardes2022VICRegVR}
Adrien Bardes, Jean Ponce, and Yann LeCun.
\newblock Vicreg: Variance-invariance-covariance regularization for
  self-supervised learning.
\newblock In \emph{International Conference on Learning Representations}, 2022.

\bibitem[Brown et~al.(2020)Brown, Mann, Ryder, Subbiah, Kaplan, Dhariwal,
  Neelakantan, Shyam, Sastry, Askell, et~al.]{Brown2020LanguageMA}
Tom Brown, Benjamin Mann, Nick Ryder, Melanie Subbiah, Jared~D Kaplan, Prafulla
  Dhariwal, Arvind Neelakantan, Pranav Shyam, Girish Sastry, Amanda Askell,
  et~al.
\newblock Language models are few-shot learners.
\newblock \emph{Advances in Neural Information Processing Systems},
  33:\penalty0 1877--1901, 2020.

\bibitem[Cai \& Vasconcelos(2018)Cai and Vasconcelos]{cai2018cascade}
Zhaowei Cai and Nuno Vasconcelos.
\newblock Cascade r-cnn: Delving into high quality object detection.
\newblock In \emph{Proceedings of the IEEE/CVF Conference on Computer Vision
  and Pattern Recognition}, pp.\  6154--6162, 2018.

\bibitem[Caron et~al.(2021)Caron, Touvron, Misra, J'egou, Mairal, Bojanowski,
  and Joulin]{Caron2021EmergingPI}
Mathilde Caron, Hugo Touvron, Ishan Misra, Herv'e J'egou, Julien Mairal, Piotr
  Bojanowski, and Armand Joulin.
\newblock Emerging properties in self-supervised vision transformers.
\newblock \emph{International Conference on Computer Vision}, pp.\  9630--9640,
  2021.

\bibitem[Chen et~al.(2021{\natexlab{a}})Chen, Fan, and Panda]{chen2021crossvit}
Chun-Fu~(Richard) Chen, Quanfu Fan, and Rameswar Panda.
\newblock {CrossViT: Cross-Attention Multi-Scale Vision Transformer for Image
  Classification}.
\newblock In \emph{International Conference on Computer Vision},
  2021{\natexlab{a}}.

\bibitem[Chen et~al.(2019)Chen, Zhang, Zhang, Tao, Gan, Zhang, Li, Shen, Chen,
  and Carin]{chen2018improving}
Liqun Chen, Yizhe Zhang, Ruiyi Zhang, Chenyang Tao, Zhe Gan, Haichao Zhang, Bai
  Li, Dinghan Shen, Changyou Chen, and Lawrence Carin.
\newblock Improving sequence-to-sequence learning via optimal transport.
\newblock In \emph{International Conference on Learning Representations}, 2019.
\newblock URL \url{https://openreview.net/forum?id=S1xtAjR5tX}.

\bibitem[Chen et~al.(2020{\natexlab{a}})Chen, Gan, Cheng, Li, Carin, and
  Liu]{chen2020graph}
Liqun Chen, Zhe Gan, Yu~Cheng, Linjie Li, Lawrence Carin, and Jingjing Liu.
\newblock Graph optimal transport for cross-domain alignment.
\newblock In \emph{International Conference on Machine Learning}, pp.\
  1542--1553. PMLR, 2020{\natexlab{a}}.

\bibitem[Chen et~al.(2020{\natexlab{b}})Chen, Kornblith, Norouzi, and
  Hinton]{chen2020simple}
Ting Chen, Simon Kornblith, Mohammad Norouzi, and Geoffrey Hinton.
\newblock A simple framework for contrastive learning of visual
  representations.
\newblock In \emph{International Conference on Machine Learning}, pp.\
  1597--1607. PMLR, 2020{\natexlab{b}}.

\bibitem[Chen \& He(2021)Chen and He]{Chen2021ExploringSS}
Xinlei Chen and Kaiming He.
\newblock Exploring simple siamese representation learning.
\newblock \emph{Proceedings of the IEEE/CVF Conference on Computer Vision and
  Pattern Recognition}, pp.\  15745--15753, 2021.

\bibitem[Chen et~al.(2020{\natexlab{c}})Chen, Fan, Girshick, and
  He]{chen2020mocov2}
Xinlei Chen, Haoqi Fan, Ross Girshick, and Kaiming He.
\newblock Improved baselines with momentum contrastive learning.
\newblock \emph{arXiv preprint arXiv:2003.04297}, 2020{\natexlab{c}}.

\bibitem[Chen et~al.(2021{\natexlab{b}})Chen, Xie, and He]{Chen2021AnES}
Xinlei Chen, Saining Xie, and Kaiming He.
\newblock An empirical study of training self-supervised vision transformers.
\newblock \emph{International Conference on Computer Vision}, pp.\  9620--9629,
  2021{\natexlab{b}}.

\bibitem[Chen et~al.(2020{\natexlab{d}})Chen, Li, Yu, El~Kholy, Ahmed, Gan,
  Cheng, and Liu]{chen2020uniter}
Yen-Chun Chen, Linjie Li, Licheng Yu, Ahmed El~Kholy, Faisal Ahmed, Zhe Gan,
  Yu~Cheng, and Jingjing Liu.
\newblock Uniter: Universal image-text representation learning.
\newblock In \emph{European Conference on Computer Vision}, pp.\  104--120.
  Springer, 2020{\natexlab{d}}.

\bibitem[Cho et~al.(2021)Cho, Lei, Tan, and Bansal]{cho2021unifying}
Jaemin Cho, Jie Lei, Hao Tan, and Mohit Bansal.
\newblock Unifying vision-and-language tasks via text generation.
\newblock In \emph{International Conference on Machine Learning}, pp.\
  1931--1942. PMLR, 2021.

\bibitem[Dai et~al.(2021)Dai, Chen, Xiao, Chen, Liu, Yuan, and
  Zhang]{dai2021dynamic}
Xiyang Dai, Yinpeng Chen, Bin Xiao, Dongdong Chen, Mengchen Liu, Lu~Yuan, and
  Lei Zhang.
\newblock Dynamic head: Unifying object detection heads with attentions.
\newblock In \emph{Proceedings of the IEEE/CVF Conference on Computer Vision
  and Pattern Recognition}, pp.\  7373--7382, 2021.

\bibitem[Deng et~al.(2009)Deng, Dong, Socher, Li, Li, and
  Fei-Fei]{deng2009imagenet}
Jia Deng, Wei Dong, Richard Socher, Li-Jia Li, Kai Li, and Li~Fei-Fei.
\newblock Imagenet: A large-scale hierarchical image database.
\newblock In \emph{Proceedings of the IEEE Conference on Computer Vision and
  Pattern Recognition}, pp.\  248--255. Ieee, 2009.

\bibitem[Devlin et~al.(2019)Devlin, Chang, Lee, and
  Toutanova]{Devlin2019BERTPO}
Jacob Devlin, Ming-Wei Chang, Kenton Lee, and Kristina Toutanova.
\newblock Bert: Pre-training of deep bidirectional transformers for language
  understanding.
\newblock In \emph{Proceedings of the 2019 Conference of the North American
  Chapter of the Association for Computational Linguistics: Human Language
  Technologies, Volume 1 (Long and Short Papers)}, pp.\  4171--4186, 2019.

\bibitem[Dosovitskiy et~al.(2021)Dosovitskiy, Beyer, Kolesnikov, Weissenborn,
  Zhai, Unterthiner, Dehghani, Minderer, Heigold, Gelly, Uszkoreit, and
  Houlsby]{dosovitskiy2021an}
Alexey Dosovitskiy, Lucas Beyer, Alexander Kolesnikov, Dirk Weissenborn,
  Xiaohua Zhai, Thomas Unterthiner, Mostafa Dehghani, Matthias Minderer, Georg
  Heigold, Sylvain Gelly, Jakob Uszkoreit, and Neil Houlsby.
\newblock An image is worth 16x16 words: Transformers for image recognition at
  scale.
\newblock In \emph{International Conference on Learning Representations}, 2021.
\newblock URL \url{https://openreview.net/forum?id=YicbFdNTTy}.

\bibitem[Dou et~al.(2022{\natexlab{a}})Dou, Kamath, Gan, Zhang, Wang, Li, Liu,
  Liu, LeCun, Peng, et~al.]{dou2022coarse}
Zi-Yi Dou, Aishwarya Kamath, Zhe Gan, Pengchuan Zhang, Jianfeng Wang, Linjie
  Li, Zicheng Liu, Ce~Liu, Yann LeCun, Nanyun Peng, et~al.
\newblock Coarse-to-fine vision-language pre-training with fusion in the
  backbone.
\newblock \emph{Advances in Neural Information Processing Systems},
  2022{\natexlab{a}}.

\bibitem[Dou et~al.(2022{\natexlab{b}})Dou, Xu, Gan, Wang, Wang, Wang, Zhu,
  Zhang, Yuan, Peng, et~al.]{dou2022empirical}
Zi-Yi Dou, Yichong Xu, Zhe Gan, Jianfeng Wang, Shuohang Wang, Lijuan Wang,
  Chenguang Zhu, Pengchuan Zhang, Lu~Yuan, Nanyun Peng, et~al.
\newblock An empirical study of training end-to-end vision-and-language
  transformers.
\newblock In \emph{Proceedings of the IEEE/CVF Conference on Computer Vision
  and Pattern Recognition}, pp.\  18166--18176, 2022{\natexlab{b}}.

\bibitem[Everingham et~al.(2010)Everingham, Van~Gool, Williams, Winn, and
  Zisserman]{everingham2010pascal}
Mark Everingham, Luc Van~Gool, Christopher~KI Williams, John Winn, and Andrew
  Zisserman.
\newblock The pascal visual object classes (voc) challenge.
\newblock \emph{International Journal of Computer Vision}, 88\penalty0
  (2):\penalty0 303--338, 2010.

\bibitem[Gao et~al.(2021)Gao, Yao, and Chen]{Gao2021SimCSESC}
Tianyu Gao, Xingcheng Yao, and Danqi Chen.
\newblock Simcse: Simple contrastive learning of sentence embeddings.
\newblock In \emph{Proceedings of the 2021 Conference on Empirical Methods in
  Natural Language Processing}, pp.\  6894--6910, 2021.

\bibitem[Garrido et~al.(2022)Garrido, Chen, Bardes, Najman, and
  Lecun]{garrido2022duality}
Quentin Garrido, Yubei Chen, Adrien Bardes, Laurent Najman, and Yann Lecun.
\newblock On the duality between contrastive and non-contrastive
  self-supervised learning.
\newblock \emph{arXiv preprint arXiv:2206.02574}, 2022.

\bibitem[Genevay et~al.(2018)Genevay, Peyr{\'e}, and
  Cuturi]{genevay2018learning}
Aude Genevay, Gabriel Peyr{\'e}, and Marco Cuturi.
\newblock Learning generative models with sinkhorn divergences.
\newblock In \emph{International Conference on Artificial Intelligence and
  Statistics}, pp.\  1608--1617. PMLR, 2018.

\bibitem[Geng et~al.(2022)Geng, Liu, Lee, Schuurams, Levine, and
  Abbeel]{Geng2022MultimodalMA}
Xinyang Geng, Hao Liu, Lisa Lee, Dale Schuurams, Sergey Levine, and P.~Abbeel.
\newblock Multimodal masked autoencoders learn transferable representations.
\newblock \emph{arXiv}, abs/2205.14204, 2022.

\bibitem[Goyal et~al.(2017)Goyal, Doll{\'a}r, Girshick, Noordhuis, Wesolowski,
  Kyrola, Tulloch, Jia, and He]{goyal2017accurate}
Priya Goyal, Piotr Doll{\'a}r, Ross Girshick, Pieter Noordhuis, Lukasz
  Wesolowski, Aapo Kyrola, Andrew Tulloch, Yangqing Jia, and Kaiming He.
\newblock Accurate, large minibatch sgd: Training imagenet in 1 hour.
\newblock \emph{arXiv preprint arXiv:1706.02677}, 2017.

\bibitem[Grill et~al.(2020)Grill, Strub, Altch{\'e}, Tallec, Richemond,
  Buchatskaya, Doersch, Avila~Pires, Guo, Gheshlaghi~Azar,
  et~al.]{Grill2020BootstrapYO}
Jean-Bastien Grill, Florian Strub, Florent Altch{\'e}, Corentin Tallec, Pierre
  Richemond, Elena Buchatskaya, Carl Doersch, Bernardo Avila~Pires, Zhaohan
  Guo, Mohammad Gheshlaghi~Azar, et~al.
\newblock Bootstrap your own latent-a new approach to self-supervised learning.
\newblock \emph{Advances in Neural Information Processing Systems},
  33:\penalty0 21271--21284, 2020.

\bibitem[Gupta et~al.(2019)Gupta, Dollar, and Girshick]{gupta2019lvis}
Agrim Gupta, Piotr Dollar, and Ross Girshick.
\newblock Lvis: A dataset for large vocabulary instance segmentation.
\newblock In \emph{Proceedings of the IEEE Conference on Computer Vision and
  Pattern Recognition}, pp.\  5356--5364, 2019.

\bibitem[Han et~al.(2015)Han, Pool, Tran, and Dally]{han2015learning}
Song Han, Jeff Pool, John Tran, and William Dally.
\newblock Learning both weights and connections for efficient neural network.
\newblock \emph{Advances in Neural Information Processing Systems}, 28, 2015.

\bibitem[Han et~al.(2023)Han, Zhu, Yu, Zhang, Song, and Xiang]{famevil2023}
Xiao Han, Xiatian Zhu, Licheng Yu, Li~Zhang, Yi-Zhe Song, and Tao Xiang.
\newblock Fame-vil: Multi-tasking vision-language model for heterogeneous
  fashion tasks.
\newblock In \emph{Proceedings of the IEEE/CVF Conference on Computer Vision
  and Pattern Recognition}, pp.\  2669--2680, 2023.

\bibitem[He et~al.(2016)He, Zhang, Ren, and Sun]{he2016deep}
Kaiming He, Xiangyu Zhang, Shaoqing Ren, and Jian Sun.
\newblock Deep residual learning for image recognition.
\newblock In \emph{Proceedings of the IEEE Conference on Computer Vision and
  Pattern Recognition}, pp.\  770--778, 2016.

\bibitem[He et~al.(2017)He, Gkioxari, Doll{\'a}r, and Girshick]{he2017mask}
Kaiming He, Georgia Gkioxari, Piotr Doll{\'a}r, and Ross Girshick.
\newblock Mask r-cnn.
\newblock In \emph{International Conference on Computer Vision}, pp.\
  2961--2969, 2017.

\bibitem[He et~al.(2020)He, Fan, Wu, Xie, and Girshick]{he2020momentum}
Kaiming He, Haoqi Fan, Yuxin Wu, Saining Xie, and Ross Girshick.
\newblock Momentum contrast for unsupervised visual representation learning.
\newblock In \emph{Proceedings of the IEEE/CVF Conference on Computer Vision
  and Pattern Recognition}, pp.\  9729--9738, 2020.

\bibitem[He et~al.(2022)He, Chen, Xie, Li, Doll{\'a}r, and
  Girshick]{He2021MaskedAA}
Kaiming He, Xinlei Chen, Saining Xie, Yanghao Li, Piotr Doll{\'a}r, and Ross
  Girshick.
\newblock Masked autoencoders are scalable vision learners.
\newblock In \emph{Proceedings of the IEEE/CVF Conference on Computer Vision
  and Pattern Recognition}, pp.\  16000--16009, 2022.

\bibitem[Huang et~al.(2020)Huang, Zeng, Liu, Fu, and Fu]{huang2020pixel}
Zhicheng Huang, Zhaoyang Zeng, Bei Liu, Dongmei Fu, and Jianlong Fu.
\newblock Pixel-bert: Aligning image pixels with text by deep multi-modal
  transformers.
\newblock \emph{arXiv preprint arXiv:2004.00849}, 2020.

\bibitem[Huang et~al.(2021)Huang, Zeng, Huang, Liu, Fu, and
  Fu]{huang2021seeing}
Zhicheng Huang, Zhaoyang Zeng, Yupan Huang, Bei Liu, Dongmei Fu, and Jianlong
  Fu.
\newblock Seeing out of the box: End-to-end pre-training for vision-language
  representation learning.
\newblock In \emph{Proceedings of the IEEE/CVF Conference on Computer Vision
  and Pattern Recognition}, pp.\  12976--12985, 2021.

\bibitem[Jang et~al.(2023)Jang, Kong, Jeon, Kim, and Kwak]{oneR2023}
Jiho Jang, Chaerin Kong, Donghyeon Jeon, Seonhoon Kim, and Nojun Kwak.
\newblock Unifying vision-language representation space with single-tower
  transformer.
\newblock In \emph{Proceedings of the AAAI Conference on Artificial
  Intelligence}, 2023.

\bibitem[Jia et~al.(2021)Jia, Yang, Xia, Chen, Parekh, Pham, Le, Sung, Li, and
  Duerig]{jia2021scaling}
Chao Jia, Yinfei Yang, Ye~Xia, Yi-Ting Chen, Zarana Parekh, Hieu Pham, Quoc Le,
  Yun-Hsuan Sung, Zhen Li, and Tom Duerig.
\newblock Scaling up visual and vision-language representation learning with
  noisy text supervision.
\newblock In \emph{International Conference on Machine Learning}, pp.\
  4904--4916. PMLR, 2021.

\bibitem[Kamath et~al.(2021)Kamath, Singh, LeCun, Synnaeve, Misra, and
  Carion]{kamath2021mdetr}
Aishwarya Kamath, Mannat Singh, Yann LeCun, Gabriel Synnaeve, Ishan Misra, and
  Nicolas Carion.
\newblock Mdetr-modulated detection for end-to-end multi-modal understanding.
\newblock In \emph{International Conference on Computer Vision}, pp.\
  1780--1790, 2021.

\bibitem[Kazemzadeh et~al.(2014)Kazemzadeh, Ordonez, Matten, and
  Berg]{kazemzadeh2014referitgame}
Sahar Kazemzadeh, Vicente Ordonez, Mark Matten, and Tamara Berg.
\newblock Referitgame: Referring to objects in photographs of natural scenes.
\newblock In \emph{Proceedings of the 2014 Conference on Empirical Methods in
  Natural Language Processing}, pp.\  787--798, 2014.

\bibitem[Kiela et~al.(2019)Kiela, Bhooshan, Firooz, and
  Testuggine]{Kiela2019SupervisedMB}
Douwe Kiela, Suvrat Bhooshan, Hamed Firooz, and Davide Testuggine.
\newblock Supervised multimodal bitransformers for classifying images and text.
\newblock \emph{arXiv}, abs/1909.02950, 2019.

\bibitem[Kim et~al.(2021)Kim, Son, and Kim]{kim2021vilt}
Wonjae Kim, Bokyung Son, and Ildoo Kim.
\newblock Vilt: Vision-and-language transformer without convolution or region
  supervision.
\newblock In \emph{International Conference on Machine Learning}, pp.\
  5583--5594. PMLR, 2021.

\bibitem[Krishna et~al.(2017)Krishna, Zhu, Groth, Johnson, Hata, Kravitz, Chen,
  Kalantidis, Li, Shamma, et~al.]{krishna2017visualgenome}
Ranjay Krishna, Yuke Zhu, Oliver Groth, Justin Johnson, Kenji Hata, Joshua
  Kravitz, Stephanie Chen, Yannis Kalantidis, Li-Jia Li, David~A Shamma, et~al.
\newblock Visual genome: Connecting language and vision using crowdsourced
  dense image annotations.
\newblock In \emph{International Journal of Computer Vision}, volume 123, pp.\
  32--73. Springer, 2017.

\bibitem[Li et~al.(2022{\natexlab{a}})Li, He, Wei, Qian, Zhu, Xie, Zhuang,
  Tian, and Tang]{li2022loupe}
Juncheng Li, Xin He, Longhui Wei, Long Qian, Linchao Zhu, Lingxi Xie, Yueting
  Zhuang, Qi~Tian, and Siliang Tang.
\newblock Fine-grained semantically aligned vision-language pre-training.
\newblock In \emph{Advances in Neural Information Processing Systems},
  2022{\natexlab{a}}.

\bibitem[Li et~al.(2021{\natexlab{a}})Li, Selvaraju, Gotmare, Joty, Xiong, and
  Hoi]{li2021align}
Junnan Li, Ramprasaath Selvaraju, Akhilesh Gotmare, Shafiq Joty, Caiming Xiong,
  and Steven Chu~Hong Hoi.
\newblock Align before fuse: Vision and language representation learning with
  momentum distillation.
\newblock \emph{Advances in Neural Information Processing Systems},
  34:\penalty0 9694--9705, 2021{\natexlab{a}}.

\bibitem[Li et~al.(2022{\natexlab{b}})Li, Li, Xiong, and Hoi]{li2022blip}
Junnan Li, Dongxu Li, Caiming Xiong, and Steven Hoi.
\newblock Blip: Bootstrapping language-image pre-training for unified
  vision-language understanding and generation.
\newblock In \emph{International Conference on Machine Learning},
  2022{\natexlab{b}}.

\bibitem[Li et~al.(2023{\natexlab{a}})Li, Li, Savarese, and Hoi]{blip22023}
Junnan Li, Dongxu Li, Silvio Savarese, and Steven Hoi.
\newblock Blip-2: Bootstrapping language-image pre-training with frozen image
  encoders and large language models.
\newblock In \emph{International Conference on Machine Learning},
  2023{\natexlab{a}}.

\bibitem[Li et~al.(2020{\natexlab{a}})Li, Yatskar, Yin, Hsieh, and
  Chang]{li2020visualbert}
Liunian~Harold Li, Mark Yatskar, Da~Yin, Cho-Jui Hsieh, and Kai-Wei Chang.
\newblock What does bert with vision look at?
\newblock In \emph{Proceedings of the 58th Annual Meeting of the Association
  for Computational Linguistics}, pp.\  5265--5275, 2020{\natexlab{a}}.

\bibitem[Li et~al.(2022{\natexlab{c}})Li, Zhang, Zhang, Yang, Li, Zhong, Wang,
  Yuan, Zhang, Hwang, et~al.]{li2022grounded}
Liunian~Harold Li, Pengchuan Zhang, Haotian Zhang, Jianwei Yang, Chunyuan Li,
  Yiwu Zhong, Lijuan Wang, Lu~Yuan, Lei Zhang, Jenq-Neng Hwang, et~al.
\newblock Grounded language-image pre-training.
\newblock In \emph{Proceedings of the IEEE/CVF Conference on Computer Vision
  and Pattern Recognition}, pp.\  10965--10975, 2022{\natexlab{c}}.

\bibitem[Li et~al.(2022{\natexlab{d}})Li, Xu, Wang, Zhou, Lin, Zhu, Zeng, Ji,
  and Chang]{li2022clip}
Manling Li, Ruochen Xu, Shuohang Wang, Luowei Zhou, Xudong Lin, Chenguang Zhu,
  Michael Zeng, Heng Ji, and Shih-Fu Chang.
\newblock Clip-event: Connecting text and images with event structures.
\newblock In \emph{Proceedings of the IEEE/CVF Conference on Computer Vision
  and Pattern Recognition}, pp.\  16420--16429, 2022{\natexlab{d}}.

\bibitem[Li et~al.(2020{\natexlab{b}})Li, Yin, Li, Zhang, Hu, Zhang, Wang, Hu,
  Dong, Wei, et~al.]{li2020oscar}
Xiujun Li, Xi~Yin, Chunyuan Li, Pengchuan Zhang, Xiaowei Hu, Lei Zhang, Lijuan
  Wang, Houdong Hu, Li~Dong, Furu Wei, et~al.
\newblock Oscar: Object-semantics aligned pre-training for vision-language
  tasks.
\newblock In \emph{European Conference on Computer Vision}, pp.\  121--137.
  Springer, 2020{\natexlab{b}}.

\bibitem[Li et~al.(2021{\natexlab{b}})Li, Liang, Zhao, Cui, Ouyang, Shao, Yu,
  and Yan]{li2021supervision}
Yangguang Li, Feng Liang, Lichen Zhao, Yufeng Cui, Wanli Ouyang, Jing Shao,
  Fengwei Yu, and Junjie Yan.
\newblock Supervision exists everywhere: A data efficient contrastive
  language-image pre-training paradigm.
\newblock In \emph{International Conference on Learning Representations},
  2021{\natexlab{b}}.

\bibitem[Li et~al.(2023{\natexlab{b}})Li, Fan, Hu, Feichtenhofer, and
  He]{flip2023}
Yanghao Li, Haoqi Fan, Ronghang Hu, Christoph Feichtenhofer, and Kaiming He.
\newblock Scaling language-image pre-training via masking.
\newblock In \emph{Proceedings of the IEEE/CVF Conference on Computer Vision
  and Pattern Recognition}, pp.\  23390--23400, 2023{\natexlab{b}}.

\bibitem[Lin et~al.(2014)Lin, Maire, Belongie, Hays, Perona, Ramanan,
  Doll{\'a}r, and Zitnick]{lin2014microsoft}
Tsung-Yi Lin, Michael Maire, Serge Belongie, James Hays, Pietro Perona, Deva
  Ramanan, Piotr Doll{\'a}r, and C~Lawrence Zitnick.
\newblock Microsoft coco: Common objects in context.
\newblock In \emph{European Conference on Computer Vision}, pp.\  740--755.
  Springer, 2014.

\bibitem[Liu et~al.(2019)Liu, Ott, Goyal, Du, Joshi, Chen, Levy, Lewis,
  Zettlemoyer, and Stoyanov]{liu2019roberta}
Yinhan Liu, Myle Ott, Naman Goyal, Jingfei Du, Mandar Joshi, Danqi Chen, Omer
  Levy, Mike Lewis, Luke Zettlemoyer, and Veselin Stoyanov.
\newblock Roberta: A robustly optimized bert pretraining approach.
\newblock \emph{arXiv preprint arXiv:1907.11692}, 2019.

\bibitem[Liu et~al.(2021)Liu, Lin, Cao, Hu, Wei, Zhang, Lin, and
  Guo]{liu2021swin}
Ze~Liu, Yutong Lin, Yue Cao, Han Hu, Yixuan Wei, Zheng Zhang, Stephen Lin, and
  Baining Guo.
\newblock Swin transformer: Hierarchical vision transformer using shifted
  windows.
\newblock In \emph{International Conference on Computer Vision}, pp.\
  10012--10022, 2021.

\bibitem[Loshchilov \& Hutter(2016)Loshchilov and Hutter]{loshchilov2016sgdr}
Ilya Loshchilov and Frank Hutter.
\newblock Sgdr: Stochastic gradient descent with warm restarts.
\newblock \emph{arXiv preprint arXiv:1608.03983}, 2016.

\bibitem[Lu et~al.(2019)Lu, Batra, Parikh, and Lee]{lu2019vilbert}
Jiasen Lu, Dhruv Batra, Devi Parikh, and Stefan Lee.
\newblock Vilbert: Pretraining task-agnostic visiolinguistic representations
  for vision-and-language tasks.
\newblock \emph{Advances in Neural Information Processing Systems}, 32, 2019.

\bibitem[Ma et~al.(2022)Ma, Li, Li, and Huang]{ma2022cmal}
Zhiyuan Ma, Jianjun Li, Guohui Li, and Kaiyan Huang.
\newblock Cmal: A novel cross-modal associative learning framework for
  vision-language pre-training.
\newblock In \emph{Proceedings of the 30th ACM International Conference on
  Multimedia}, pp.\  4515--4524, 2022.

\bibitem[Mroueh et~al.(2018)Mroueh, Li, Sercu, Raj, and
  Cheng]{mroueh2018sobolev}
Youssef Mroueh, Chun-Liang Li, Tom Sercu, Anant Raj, and Yu~Cheng.
\newblock Sobolev gan.
\newblock In \emph{International Conference on Learning Representations}, 2018.

\bibitem[Mroueh et~al.(2019)Mroueh, Sercu, and Raj]{mroueh2019sobolev}
Youssef Mroueh, Tom Sercu, and Anant Raj.
\newblock Sobolev descent.
\newblock In \emph{International Conference on Artificial Intelligence and
  Statistics}, pp.\  2976--2985. PMLR, 2019.

\bibitem[Mu et~al.(2021)Mu, Kirillov, Wagner, and Xie]{Mu2021SLIPSM}
Norman Mu, Alexander Kirillov, David~A. Wagner, and Saining Xie.
\newblock Slip: Self-supervision meets language-image pre-training.
\newblock \emph{arXiv}, abs/2112.12750, 2021.

\bibitem[Oord et~al.(2018)Oord, Li, and Vinyals]{oord2018representation}
Aaron van~den Oord, Yazhe Li, and Oriol Vinyals.
\newblock Representation learning with contrastive predictive coding.
\newblock \emph{arXiv preprint arXiv:1807.03748}, 2018.

\bibitem[Pang \& He(2021)Pang and He]{pang2021text}
Richard~Yuanzhe Pang and He~He.
\newblock Text generation by learning from demonstrations.
\newblock In \emph{International Conference on Learning Representations}, 2021.

\bibitem[Papineni et~al.(2002)Papineni, Roukos, Ward, and
  Zhu]{papineni2002bleu}
Kishore Papineni, Salim Roukos, Todd Ward, and Wei-Jing Zhu.
\newblock Bleu: a method for automatic evaluation of machine translation.
\newblock In \emph{Proceedings of the 40th Annual Meeting of the Association
  for Computational Linguistics}, pp.\  311--318, 2002.

\bibitem[Park \& Han(2023)Park and Han]{softmask2023}
Jaeyoo Park and Bohyung Han.
\newblock Multi-modal representation learning with text-driven soft masks.
\newblock In \emph{Proceedings of the IEEE/CVF Conference on Computer Vision
  and Pattern Recognition}, pp.\  2798--2807, 2023.

\bibitem[Peyr{\'e} et~al.(2016)Peyr{\'e}, Cuturi, and Solomon]{peyre2016gromov}
Gabriel Peyr{\'e}, Marco Cuturi, and Justin Solomon.
\newblock Gromov-wasserstein averaging of kernel and distance matrices.
\newblock In \emph{International Conference on Machine Learning}, pp.\
  2664--2672. PMLR, 2016.

\bibitem[Peyr{\'e} et~al.(2019)Peyr{\'e}, Cuturi,
  et~al.]{peyre2019computational}
Gabriel Peyr{\'e}, Marco Cuturi, et~al.
\newblock Computational optimal transport: With applications to data science.
\newblock \emph{Foundations and Trends{\textregistered} in Machine Learning},
  11\penalty0 (5-6):\penalty0 355--607, 2019.

\bibitem[Plummer et~al.(2015)Plummer, Wang, Cervantes, Caicedo, Hockenmaier,
  and Lazebnik]{plummer2015flickr30k}
Bryan~A Plummer, Liwei Wang, Chris~M Cervantes, Juan~C Caicedo, Julia
  Hockenmaier, and Svetlana Lazebnik.
\newblock Flickr30k entities: Collecting region-to-phrase correspondences for
  richer image-to-sentence models.
\newblock In \emph{International Conference on Computer Vision}, pp.\
  2641--2649, 2015.

\bibitem[Pramanick et~al.(2022)Pramanick, Roy, and
  Patel]{pramanick2022multimodal}
Shraman Pramanick, Aniket Roy, and Vishal~M Patel.
\newblock Multimodal learning using optimal transport for sarcasm and humor
  detection.
\newblock In \emph{Proceedings of the IEEE/CVF Winter Conference on
  Applications of Computer Vision}, pp.\  3930--3940, 2022.

\bibitem[Pramanick et~al.(2023)Pramanick, Song, Nag, Lin, Shah, Shou,
  Chellappa, and Zhang]{pramanick2023egovlpv2}
Shraman Pramanick, Yale Song, Sayan Nag, Kevin~Qinghong Lin, Hardik Shah,
  Mike~Zheng Shou, Rama Chellappa, and Pengchuan Zhang.
\newblock Egovlpv2: Egocentric video-language pre-training with fusion in the
  backbone.
\newblock \emph{arXiv preprint arXiv:2307.05463}, 2023.

\bibitem[Radford et~al.(2019)Radford, Wu, Child, Luan, Amodei, and
  Sutskever]{Radford2019LanguageMA}
Alec Radford, Jeff Wu, Rewon Child, David Luan, Dario Amodei, and Ilya
  Sutskever.
\newblock Language models are unsupervised multitask learners.
\newblock 2019.

\bibitem[Radford et~al.(2021)Radford, Kim, Hallacy, Ramesh, Goh, Agarwal,
  Sastry, Askell, Mishkin, Clark, et~al.]{radford2021learning}
Alec Radford, Jong~Wook Kim, Chris Hallacy, Aditya Ramesh, Gabriel Goh,
  Sandhini Agarwal, Girish Sastry, Amanda Askell, Pamela Mishkin, Jack Clark,
  et~al.
\newblock Learning transferable visual models from natural language
  supervision.
\newblock In \emph{International Conference on Machine Learning}, pp.\
  8748--8763. PMLR, 2021.

\bibitem[Ren et~al.(2015)Ren, He, Girshick, and Sun]{ren2015faster}
Shaoqing Ren, Kaiming He, Ross Girshick, and Jian Sun.
\newblock Faster r-cnn: Towards real-time object detection with region proposal
  networks.
\newblock \emph{Advances in Neural Information Processing Systems}, 28, 2015.

\bibitem[Shah et~al.(2022)Shah, Sra, Chellappa, and Cherian]{shah2022max}
Anshul Shah, Suvrit Sra, Rama Chellappa, and Anoop Cherian.
\newblock Max-margin contrastive learning.
\newblock In \emph{Proceedings of the AAAI Conference on Artificial
  Intelligence}, volume~36, pp.\  8220--8230, 2022.

\bibitem[Singh et~al.(2022)Singh, Hu, Goswami, Couairon, Galuba, Rohrbach, and
  Kiela]{singh2022flava}
Amanpreet Singh, Ronghang Hu, Vedanuj Goswami, Guillaume Couairon, Wojciech
  Galuba, Marcus Rohrbach, and Douwe Kiela.
\newblock Flava: A foundational language and vision alignment model.
\newblock In \emph{Proceedings of the IEEE/CVF Conference on Computer Vision
  and Pattern Recognition}, pp.\  15638--15650, 2022.

\bibitem[Su et~al.(2019)Su, Zhu, Cao, Li, Lu, Wei, and Dai]{su2019vl}
Weijie Su, Xizhou Zhu, Yue Cao, Bin Li, Lewei Lu, Furu Wei, and Jifeng Dai.
\newblock Vl-bert: Pre-training of generic visual-linguistic representations.
\newblock In \emph{International Conference on Learning Representations}, 2019.

\bibitem[Suhr et~al.(2019)Suhr, Zhou, Zhang, Zhang, Bai, and
  Artzi]{suhr2019corpus}
Alane Suhr, Stephanie Zhou, Ally Zhang, Iris Zhang, Huajun Bai, and Yoav Artzi.
\newblock A corpus for reasoning about natural language grounded in
  photographs.
\newblock In \emph{Proceedings of the Annual Meeting of the Association for
  Computational Linguistics}, 2019.

\bibitem[Tan \& Bansal(2019)Tan and Bansal]{tan2019lxmert}
Hao Tan and Mohit Bansal.
\newblock Lxmert: Learning cross-modality encoder representations from
  transformers.
\newblock In \emph{Proceedings of the 2019 Conference on Empirical Methods in
  Natural Language Processing and the 9th International Joint Conference on
  Natural Language Processing}, pp.\  5100--5111, 2019.

\bibitem[Vedantam et~al.(2015)Vedantam, Lawrence~Zitnick, and
  Parikh]{vedantam2015cider}
Ramakrishna Vedantam, C~Lawrence~Zitnick, and Devi Parikh.
\newblock Cider: Consensus-based image description evaluation.
\newblock In \emph{Proceedings of the IEEE Conference on Computer Vision and
  Pattern Recognition}, pp.\  4566--4575, 2015.

\bibitem[Wang et~al.(2023{\natexlab{a}})Wang, Yang, Hu, Li, Lin, Gan, Liu, Liu,
  and Wang]{git2023}
Jianfeng Wang, Zhengyuan Yang, Xiaowei Hu, Linjie Li, Kevin Lin, Zhe Gan,
  Zicheng Liu, Ce~Liu, and Lijuan Wang.
\newblock Git: A generative image-to-text transformer for vision and language.
\newblock \emph{Transactions of Machine Learning Research}, 2023{\natexlab{a}}.

\bibitem[Wang et~al.(2023{\natexlab{b}})Wang, Zhou, Shou, and Yan]{ptp2023}
Jinpeng Wang, Pan Zhou, Mike~Zheng Shou, and Shuicheng Yan.
\newblock Position-guided text prompt for vision-language pre-training.
\newblock In \emph{Proceedings of the IEEE/CVF Conference on Computer Vision
  and Pattern Recognition}, pp.\  23242--23251, 2023{\natexlab{b}}.

\bibitem[Wang et~al.(2022{\natexlab{a}})Wang, Chen, Wu, Luo, Zhou, Zhao, Xie,
  Liu, Jiang, and Yuan]{wang2022omnivl}
Junke Wang, Dongdong Chen, Zuxuan Wu, Chong Luo, Luowei Zhou, Yucheng Zhao,
  Yujia Xie, Ce~Liu, Yu-Gang Jiang, and Lu~Yuan.
\newblock Omnivl: One foundation model for image-language and video-language
  tasks.
\newblock In \emph{Advances in Neural Information Processing Systems},
  2022{\natexlab{a}}.

\bibitem[Wang et~al.(2022{\natexlab{b}})Wang, Yang, Men, Lin, Bai, Li, Ma,
  Zhou, Zhou, and Yang]{wang2022ofa}
Peng Wang, An~Yang, Rui Men, Junyang Lin, Shuai Bai, Zhikang Li, Jianxin Ma,
  Chang Zhou, Jingren Zhou, and Hongxia Yang.
\newblock Ofa: Unifying architectures, tasks, and modalities through a simple
  sequence-to-sequence learning framework.
\newblock In \emph{International Conference on Machine Learning}, pp.\
  23318--23340. PMLR, 2022{\natexlab{b}}.

\bibitem[Wang et~al.(2023{\natexlab{c}})Wang, Ge, Zheng, Cheng, Shan, Qie, and
  Luo]{flm2023}
Teng Wang, Yixiao Ge, Feng Zheng, Ran Cheng, Ying Shan, Xiaohu Qie, and Ping
  Luo.
\newblock Accelerating vision-language pretraining with free language modeling.
\newblock In \emph{Proceedings of the IEEE/CVF Conference on Computer Vision
  and Pattern Recognition}, pp.\  23161--23170, 2023{\natexlab{c}}.

\bibitem[Wang et~al.(2021{\natexlab{a}})Wang, Bao, Dong, and Wei]{wang2021vlmo}
Wenhui Wang, Hangbo Bao, Li~Dong, and Furu Wei.
\newblock Vlmo: Unified vision-language pre-training with
  mixture-of-modality-experts.
\newblock \emph{arXiv preprint arXiv:2111.02358}, 2021{\natexlab{a}}.

\bibitem[Wang et~al.(2022{\natexlab{c}})Wang, Bao, Dong, Bjorck, Peng, Liu,
  Aggarwal, Mohammed, Singhal, Som, and Wei]{Wang2022ImageAA}
Wenhui Wang, Hangbo Bao, Li~Dong, Johan Bjorck, Zhiliang Peng, Qiang Liu, Kriti
  Aggarwal, Owais Mohammed, Saksham Singhal, Subhojit Som, and Furu Wei.
\newblock Image as a foreign language: Beit pretraining for all vision and
  vision-language tasks.
\newblock \emph{arXiv}, abs/2208.10442, 2022{\natexlab{c}}.

\bibitem[Wang et~al.(2023{\natexlab{d}})Wang, Bao, Dong, Bjorck, Peng, Liu,
  Aggarwal, Mohammed, Singhal, Som, et~al.]{beit2023}
Wenhui Wang, Hangbo Bao, Li~Dong, Johan Bjorck, Zhiliang Peng, Qiang Liu, Kriti
  Aggarwal, Owais~Khan Mohammed, Saksham Singhal, Subhojit Som, et~al.
\newblock Image as a foreign language: Beit pretraining for vision and
  vision-language tasks.
\newblock In \emph{Proceedings of the IEEE/CVF Conference on Computer Vision
  and Pattern Recognition}, pp.\  19175--19186, 2023{\natexlab{d}}.

\bibitem[Wang et~al.(2021{\natexlab{b}})Wang, Yu, Yu, Dai, Tsvetkov, and
  Cao]{wang2021simvlm}
Zirui Wang, Jiahui Yu, Adams~Wei Yu, Zihang Dai, Yulia Tsvetkov, and Yuan Cao.
\newblock Simvlm: Simple visual language model pretraining with weak
  supervision.
\newblock In \emph{International Conference on Learning Representations},
  2021{\natexlab{b}}.

\bibitem[Wei \& Zou(2019)Wei and Zou]{wei2019eda}
Jason Wei and Kai Zou.
\newblock Eda: Easy data augmentation techniques for boosting performance on
  text classification tasks.
\newblock In \emph{Proceedings of the 2019 Conference on Empirical Methods in
  Natural Language Processing and the 9th International Joint Conference on
  Natural Language Processing}, pp.\  6382--6388, 2019.

\bibitem[Wu et~al.(2019)Wu, Kirillov, Massa, Lo, and
  Girshick]{wu2019detectron2}
Yuxin Wu, Alexander Kirillov, Francisco Massa, Wan-Yen Lo, and Ross Girshick.
\newblock Detectron2.
\newblock \url{https://github.com/facebookresearch/detectron2}, 2019.

\bibitem[Xue et~al.(2021)Xue, Huang, Liu, Peng, Fu, Li, and
  Luo]{xue2021probing}
Hongwei Xue, Yupan Huang, Bei Liu, Houwen Peng, Jianlong Fu, Houqiang Li, and
  Jiebo Luo.
\newblock Probing inter-modality: Visual parsing with self-attention for
  vision-and-language pre-training.
\newblock \emph{Advances in Neural Information Processing Systems},
  34:\penalty0 4514--4528, 2021.

\bibitem[Yang et~al.(2022{\natexlab{a}})Yang, Li, Zhang, Xiao, Liu, Yuan, and
  Gao]{yang2022unified}
Jianwei Yang, Chunyuan Li, Pengchuan Zhang, Bin Xiao, Ce~Liu, Lu~Yuan, and
  Jianfeng Gao.
\newblock Unified contrastive learning in image-text-label space.
\newblock In \emph{Proceedings of the IEEE/CVF Conference on Computer Vision
  and Pattern Recognition}, pp.\  19163--19173, 2022{\natexlab{a}}.

\bibitem[Yang et~al.(2022{\natexlab{b}})Yang, Duan, Tran, Xu, Chanda, Chen,
  Zeng, Chilimbi, and Huang]{yang2022vision}
Jinyu Yang, Jiali Duan, Son Tran, Yi~Xu, Sampath Chanda, Liqun Chen, Belinda
  Zeng, Trishul Chilimbi, and Junzhou Huang.
\newblock Vision-language pre-training with triple contrastive learning.
\newblock In \emph{Proceedings of the IEEE/CVF Conference on Computer Vision
  and Pattern Recognition}, pp.\  15671--15680, 2022{\natexlab{b}}.

\bibitem[Yang et~al.(2022{\natexlab{c}})Yang, Gan, Wang, Hu, Ahmed, Liu, Lu,
  and Wang]{yang2022unitab}
Zhengyuan Yang, Zhe Gan, Jianfeng Wang, Xiaowei Hu, Faisal Ahmed, Zicheng Liu,
  Yumao Lu, and Lijuan Wang.
\newblock Unitab: Unifying text and box outputs for grounded vision-language
  modeling.
\newblock In \emph{European Conference on Computer Vision,},
  2022{\natexlab{c}}.

\bibitem[Yao et~al.(2022)Yao, Huang, Hou, Lu, Niu, Xu, Liang, Li, Jiang, and
  Xu]{yao2022filip}
Lewei Yao, Runhui Huang, Lu~Hou, Guansong Lu, Minzhe Niu, Hang Xu, Xiaodan
  Liang, Zhenguo Li, Xin Jiang, and Chunjing Xu.
\newblock Filip: Fine-grained interactive language-image pre-training.
\newblock In \emph{International Conference on Learning Representations}, 2022.

\bibitem[You et~al.(2022)You, Zhou, Xiao, Codella, Cheng, Xu, Chang, and
  Yuan]{you2022msclip}
Haoxuan You, Luowei Zhou, Bin Xiao, Noel Codella, Yu~Cheng, Ruochen Xu, Shih-Fu
  Chang, and Lu~Yuan.
\newblock Learning visual representation from modality-shared contrastive
  language-image pre-training.
\newblock In \emph{European Conference on Computer Vision}, 2022.

\bibitem[You et~al.(2017)You, Gitman, and Ginsburg]{you2017large}
Yang You, Igor Gitman, and Boris Ginsburg.
\newblock Large batch training of convolutional networks.
\newblock \emph{arXiv preprint arXiv:1708.03888}, 2017.

\bibitem[Yu et~al.(2016)Yu, Poirson, Yang, Berg, and Berg]{yu2016modeling}
Licheng Yu, Patrick Poirson, Shan Yang, Alexander~C Berg, and Tamara~L Berg.
\newblock Modeling context in referring expressions.
\newblock In \emph{European Conference on Computer Vision}, pp.\  69--85.
  Springer, 2016.

\bibitem[Yuan et~al.(2020)Yuan, Bai, Chen, Zhang, Tao, Li, Wang, Henao, and
  Carin]{yuan2020advancing}
Siyang Yuan, Ke~Bai, Liqun Chen, Yizhe Zhang, Chenyang Tao, Chunyuan Li, Guoyin
  Wang, Ricardo Henao, and Lawrence Carin.
\newblock Advancing weakly supervised cross-domain alignment with optimal
  transport.
\newblock In \emph{British Machine Vision Conference}, 2020.

\bibitem[Yuan et~al.(2021)Yuan, Lin, Kuen, Zhang, Wang, Maire, Kale, and
  Faieta]{yuan2021multimodal}
Xin Yuan, Zhe Lin, Jason Kuen, Jianming Zhang, Yilin Wang, Michael Maire,
  Ajinkya Kale, and Baldo Faieta.
\newblock Multimodal contrastive training for visual representation learning.
\newblock In \emph{Proceedings of the IEEE/CVF Conference on Computer Vision
  and Pattern Recognition}, pp.\  6995--7004, 2021.

\bibitem[Zbontar et~al.(2021)Zbontar, Jing, Misra, LeCun, and
  Deny]{zbontar2021barlow}
Jure Zbontar, Li~Jing, Ishan Misra, Yann LeCun, and St{\'e}phane Deny.
\newblock Barlow twins: Self-supervised learning via redundancy reduction.
\newblock In \emph{International Conference on Machine Learning}, pp.\
  12310--12320. PMLR, 2021.

\bibitem[Zeng et~al.(2022)Zeng, Zhang, and Li]{zeng2022xvlm}
Yan Zeng, Xinsong Zhang, and Hang Li.
\newblock Multi-grained vision language pre-training: Aligning texts with
  visual concepts.
\newblock In \emph{International Conference on Machine Learning}, 2022.

\bibitem[Zhang et~al.(2020)Zhang, Cai, Lin, and Shen]{zhang2020deepemd}
Chi Zhang, Yujun Cai, Guosheng Lin, and Chunhua Shen.
\newblock Deepemd: Few-shot image classification with differentiable earth
  mover's distance and structured classifiers.
\newblock In \emph{Proceedings of the IEEE/CVF Conference on Computer Vision
  and Pattern Recognition}, pp.\  12203--12213, 2020.

\bibitem[Zhang et~al.(2022)Zhang, Zhang, Hu, Chen, Li, Dai, Wang, Yuan, Hwang,
  and Gao]{zhang2022glipv2}
Haotian Zhang, Pengchuan Zhang, Xiaowei Hu, Yen-Chun Chen, Liunian~Harold Li,
  Xiyang Dai, Lijuan Wang, Lu~Yuan, Jenq-Neng Hwang, and Jianfeng Gao.
\newblock Glipv2: Unifying localization and vision-language understanding.
\newblock \emph{Advances in Neural Information Processing Systems}, 2022.

\bibitem[Zhang et~al.(2021)Zhang, Li, Hu, Yang, Zhang, Wang, Choi, and
  Gao]{zhang2021vinvl}
Pengchuan Zhang, Xiujun Li, Xiaowei Hu, Jianwei Yang, Lei Zhang, Lijuan Wang,
  Yejin Choi, and Jianfeng Gao.
\newblock Vinvl: Revisiting visual representations in vision-language models.
\newblock In \emph{Proceedings of the IEEE/CVF Conference on Computer Vision
  and Pattern Recognition}, pp.\  5579--5588, 2021.

\bibitem[Zhou et~al.(2020)Zhou, Palangi, Zhang, Hu, Corso, and
  Gao]{zhou2020unified}
Luowei Zhou, Hamid Palangi, Lei Zhang, Houdong Hu, Jason Corso, and Jianfeng
  Gao.
\newblock Unified vision-language pre-training for image captioning and vqa.
\newblock In \emph{Proceedings of the AAAI Conference on Artificial
  Intelligence}, volume~34, pp.\  13041--13049, 2020.

\end{thebibliography}
\bibliographystyle{tmlr}

\newpage
\appendix
\onecolumn
\appendix
\counterwithin{figure}{section}
\numberwithin{table}{section}

\section{Pesudo Code of \model} \label{sec:pseudo}

The training pseudo code for \model\ is as follows:
\vspace{-1.5mm}
\begin{algorithm}[!h]
   \caption{PyTorch-style pseudocode for VoLTA.}
   \label{alg:VoLTA}
   
    \definecolor{codeblue}{rgb}{0.25,0.5,0.5}
    \definecolor{codekw}{rgb}{0.85, 0.18, 0.50}
    \newcommand{\algofontsize}{9.0pt}
    \lstset{
      backgroundcolor=\color{white},
      basicstyle=\fontsize{\algofontsize}{\algofontsize}\ttfamily\selectfont,
      columns=fullflexible,
      breaklines=true,
      captionpos=b,
      commentstyle=\fontsize{\algofontsize}{\algofontsize}\color{codeblue},
      keywordstyle=\fontsize{\algofontsize}{\algofontsize}\color{codekw},
    }
\begin{lstlisting}[language=python]
# f_I: Image Encoder, f_T: Text Encoder
# task_names: string containing task names
# I: Image input, T: Text input, N: Batch size, D: Projector dim
# BT: Barlow Twins loss function 
# WD, GWD: Wasserstein and Gromov-Wasserstein loss functions
# MLM, ITM: MLM and ITM loss functions, respectively
# gamma: coefficient of GWD loss in GOT
# w_GOT: weight of GOT loss
    
def GOT(x_1, x_2, f_1, f_2):
    # compute embeddings
    z_A, z_B = f_1(x_1), f_2(x_2) # N x D
    
    # normalize representation along batch dimension
    z_A_norm = (z_A - z_A.mean(dim=0)) / z_A.std(dim=0)
    z_B_norm = (z_B - z_B.mean(dim=0)) / z_B.std(dim=0)
    
    # cosine distance matrix
    c = cosine_dist_matrix(z_A, z_B)}
    # Wasserstein distance
    loss_w = W_D(c, z_A.size(0), z_A.size(1), z_B.size(1))
    # Gromov-Wasserstein distance
    loss_gw = GW_D(z_A.transpose(2,1), z_B.transpose(2,1))
    
    return gamma * torch.mean(loss_gw) + (1 - gamma) * torch.mean(loss_w)

def VoLTA (I, T):
    total_loss = torch.tensor(0.)
    for x in loader: # load a batch with N samples
        # two augmented versions of I, T
        I1, I2 = augment_image(I); T1, T2 = augment_text(T) 
        
        if "BTGOT" in task_names:
            # BT loss
            intra_loss = BT(I1, I2, f_I) + BT(T1, T2, f_T)}
            inter_loss = BT(I1, T1, f_I, f_T) + BT(I2, T2, f_I, f_T)}
            BT_loss = intra_loss + inter_loss
            total_loss += BT_loss
            
            # GOT loss
            GOT_loss = GOT(I1, T1, f_I, f_T) + GOT(I2, T2, f_I, f_T)}
            total_loss += w_GOT * GOT_loss
            
        # cross-attention is enabled
        if "MLM" in task_names:
        
            # MLM loss
            MLM_loss = MLM(T1, I1, mask_T1, f_I, f_T)
            total_loss += MLM_loss
            
        if "ITM" in task_names:
        
            # ITM loss
            ITM_loss =ITM(T1, I1, false_image_1, f_I, f_T)
            total_loss += ITM_loss
            
    return total_loss
\end{lstlisting}
\end{algorithm}
\vspace{-2.5mm}

\section{Overview of Vision-Language Pre-training Models}

\begin{table}[!h]
\centering

\vspace{2mm}
  \resizebox{\columnwidth}{!}{\begin{tabular}{l|l|c|c|c|c c|c}
    \hline
    \bf \multirow{2}{*}{Model} & \bf \multirow{2}{*}{Venue} & \bf \multirow{2}{*}{Vision Encoder} & \bf \multirow{2}{*}{Text Enc.} & \bf \multirow{2}{*}{Multimodality Fusion} & \multicolumn{2}{c|}{\textbf{Pre-train}} & \bf \multirow{2}{*}{Pre-training Objectives}\\
     & & & & & \bf I-T & \bf I-T-B & \\
    \hline
    ViLBERT & NeurIPS'$19$ &
\multirow{2}{*}{OD+Xformer}  & \multirow{2}{*}{Xformer}    & \multirow{2}{*}{Co-attn} & \ding{51} & & MLM+ITM+MIM\\
LXMERT & EMNLP'$19$ & &   & & \ding{51}  & & MLM+ITM+MIM+VQA \\
\cdashline{3-5}
VisualBERT & ACL'$20$ & \multirow{6}{*}{OD}   & \multirow{6}{*}{Emb.}  & \multirow{6}{*}{Merged attn} &  \ding{51}  & & MLM+ITM\\
VL-BERT & ICLR'$20$ &  &  & &  \ding{51}  & & MLM+MIM\\
UNITER & ECCV'$20$ & &   & & \ding{51} & & MLM+ITM+MIM+WRA \\
OSCAR & ECCV'$20$ &  &   & & \ding{51} & & MLM+ITM \\
VinVL & CVPR'$21$ & &  & & \ding{51} & & MLM+ITM\\
VL-T5  & ICML'$21$ &  &    & & & \ding{51} & MLM+ITM+VQA+Grnd+Cap\\
\hline
SOHO & CVPR'$21$ & \multirow{3}{*}{CNN} & \multirow{2}{*}{Emb.} & \multirow{3}{*}{Merged attn} & \ding{51} & & MLM+ITM+MIM\\
SimVLM & ICLR'$22$ &  &   &    & \ding{51} & & PrefixLM\\
\cdashline{4-4}
MDETR & ICCV'$21$ &  & Xformer &  & & \ding{51} & OD+TP+CA\\
\hline
ViLT & ICML'$21$ & \multirow{1}{*}{Patch Emb.} & \multirow{2}{*}{Emb.}  & \multirow{2}{*}{Merged attn} & \ding{51} & & MLM+ITM\\
\cdashline{3-3}
Visual Parsing & NeurIPS'$21$ & \multirow{3}{*}{Xformer} &  & & \ding{51} & & MLM+ITM+MIM\\
\cdashline{4-5}
ALBEF & NeurIPS'$21$ & &   \multirow{2}{*}{Xformer}  & \multirow{2}{*}{Co-attn}& \ding{51} & & MLM+ITM+ITC\\
METER & CVPR'$22$ & & &  & \ding{51} & & MLM+ITM\\
\hline
CLIP & ICML'$21$ & \multirow{2}{*}{CNN/Xformer} & \multirow{23}{*}{Xformer} & \multirow{3}{*}{None} &  \ding{51} & & ITC\\
DeCLIP  & ICLR'$21$ & & &  & \ding{51} & & ITC+MLM+SL+MVS+NNS\\
\cdashline{3-3}
ALIGN  & ICML'$21$ & CNN & &  & \ding{51} & & ITC\\
\cdashline{3-3}
\cdashline{5-5}
GLIP  & CVPR'$22$ & \multirow{2}{*}{OD+Xformer} & & \multirow{5}{*}{Cross-modality MHA} & \ding{51}  & \ding{51} & OD+CE+WRA\\
GLIPv2  & NeurIPS'$22$ &  & &  & \ding{51} & \ding{51} & OD+CE+WRA+MLM\\
\cdashline{3-3}
BLIP & ICML'$22$ & \multirow{6}{*}{Xformer} & &  & \ding{51} & & ITC+ITM+LM\\
\textcolor{black}{OmniVL} & \textcolor{black}{NeurIPS'$22$} & & &  & \textcolor{black}{\ding{51}} & & \textcolor{black}{UniVLC+VLM+LM}\\
\textcolor{black}{X-VLM} & \textcolor{black}{ICML'$22$} & & &  & \textcolor{black}{\ding{51}} & \textcolor{black}{\ding{51}} & \textcolor{black}{BBP+ITC+MP+MLM}\\
\cdashline{5-5}
\textcolor{black}{CMAL} & \textcolor{black}{ACM MM'$22$} & & & \multirow{4}{*}{None} &  \textcolor{black}{\ding{51}} & & \textcolor{black}{AMC+MLM+MRM+ITM+ITC}\\
\textcolor{black}{LOUPE} & \textcolor{black}{NeurIPS'$22$} & & & &  \textcolor{black}{\ding{51}} & & \textcolor{black}{ITC+FSA+TSA}\\
\textcolor{black}{FILIP} & \textcolor{black}{ICLR'$22$} & & & & \textcolor{black}{\ding{51}} & & \textcolor{black}{ITC}\\
\cdashline{3-3}
UniCL & CVPR'$22$ & CNN/Xformer & &  &  \ding{51} & & ITC\\
\cdashline{5-5}
UniTAB & ECCV'$22$ & CNN &  & \multirow{2}{*}{Merged attn} & & \ding{51} & LM\\ 
\cdashline{3-3}
\textcolor{black}{TCL} & \textcolor{black}{CVPR'$22$} & \multirow{4}{*}{\textcolor{black}{Xformer}} &  & & \textcolor{black}{\ding{51}} &  & \textcolor{black}{CMA+IMC+LMI+ITM+MLM}\\
\cdashline{5-5}
\textcolor{black}{MS-CLIP} & \textcolor{black}{ECCV'$22$} & \textcolor{black}{} & \textcolor{black}{} & \textcolor{black}{Shared Attention} & \textcolor{black}{\ding{51}} & & \textcolor{black}{ITC}\\
\cdashline{5-5}
\textcolor{black}{FLM} & \textcolor{black}{CVPR'$23$} & \textcolor{black}{} & \textcolor{black}{} & \multirow{2}{*}{\textcolor{black}{Cross-modality MHA}} & \textcolor{black}{\ding{51}} & & \textcolor{black}{FLM + ITM}\\
\textcolor{black}{BLIP-2} & \textcolor{black}{ICML'$23$} & \textcolor{black}{} & \textcolor{black}{} & & \textcolor{black}{\ding{51}} & & \textcolor{black}{ITC+ITM+ITG}\\
\cdashline{5-5}
\textcolor{black}{Fame-ViL} & \textcolor{black}{CVPR'$23$} & \textcolor{black}{} & \textcolor{black}{} & \textcolor{black}{Cross-modality Adaptive Attention} & \textcolor{black}{\ding{51}} & & \textcolor{black}{ITC}\\
\cdashline{5-5}
\cdashline{3-3}
\textcolor{black}{PTP} & \textcolor{black}{CVPR'$23$} & \textcolor{black}{OD+Xformer} & \textcolor{black}{} & \textcolor{black}{Cross-modality MHA} & \textcolor{black}{\ding{51}} & & \textcolor{black}{ITC+ITM+LM}\\
\cdashline{5-5}
\cdashline{3-3}
\textcolor{black}{Softmask++} & \textcolor{black}{CVPR'$23$} & \multirow{5}{*}{\textcolor{black}{Xformer}} & \textcolor{black}{} & \textcolor{black}{Merged attn} & \textcolor{black}{\ding{51}} & & \textcolor{black}{ITC+ITM+MLM}\\
\cdashline{5-5}
\textcolor{black}{OneR} & \textcolor{black}{AAAI'$23$} & \textcolor{black}{} & \textcolor{black}{} & \textcolor{black}{Unified attn} & \textcolor{black}{\ding{51}} & & \textcolor{black}{ITC+XMC+CIC+CMC}\\
\cdashline{5-5}
\textcolor{black}{BEiT-3} & \textcolor{black}{CVPR'$23$} & \textcolor{black}{} & \textcolor{black}{} & \textcolor{black}{Shared MHA} & \textcolor{black}{\ding{51}} & & \textcolor{black}{MDM}\\
\cdashline{5-5}
\textcolor{black}{FLIP} & \textcolor{black}{CVPR'$23$} & \textcolor{black}{} & \textcolor{black}{} & \textcolor{black}{None} & \textcolor{black}{\ding{51}} & & \textcolor{black}{ITC}\\
\cdashline{5-5}
\cdashline{4-4}
\textcolor{black}{GIT} & \textcolor{black}{TMLR'$23$} & \textcolor{black}{} & \textcolor{black}{Emb.} & \textcolor{black}{Merged attn} & \textcolor{black}{\ding{51}} & & \textcolor{black}{LM}\\
\hline
FIBER & NeurIPS'$22$ & Xformer & \multirow{1}{*}{Xformer} & \multirow{1}{*}{Merged Co-attn} & \ding{51} & \ding{51} & MLM+ITM+ITC\\
\hline

\rowcolor{Light}
\bf \model & \centering TMLR'$23$ & CNN/Xformer & \multirow{1}{*}{Xformer} & \multirow{1}{*}{Merged Co-attn}  & \ding{51} & & BT+GOT+MLM+ITM\\
\hline
\end{tabular}}
\caption{\textbf{Overview of VLP models}. OD: objective detector. Xformer: transformer. Emb.: embedding. MLM/MIM: masked language/image modeling. ITM: image-text matching. WRA: word-region alignment. ITC: image-text contrastive learning. Grnd: Grounding. Cap: Captioning. TP: Token Prediction. CA: Contrastive Alignment, NNS: Nearest Neighbour Supervision, MVS: Multiview Supervision, SL: Sim-siam Loss, MHA: Multi-head attn., LM: Language Modeling, \textcolor{black}{UniVLC: Unified Vision Language Contrastive, VLM: Vision Language Matching, BBP: Bounding Box Prediction, MP: Matching Prediction, FSA: Fine-grained Semantic Alignment, TSA: Token-level Semantic Alignment, AMC: Associative Mapping Classification, CMA: Cross-Modal Alignment, IMC: Intra-Modal Contrastive, LMI: Local Mutual Information Maximization,} I-T: Image-Text, I-T-B: Image-Text-Box, ITG: Image grounded text generation, XMC: Cross-Modal Mixup Contrastive, CIC: Contextual Invariance Contrastive, CMC: Contextual Mixup Contrast, MDM: Masked Data Modeling.}
\label{tab:vlp_glossary}
\end{table}

Vision-Language Pre-trained (VLP) models have proven extremely beneficial for multi-modal tasks in recent years. Earlier works were predominantly focused on using pre-trained object detectors to extract patch (region) level information from corresponding images \citep{lu2019vilbert,li2020visualbert,tan2019lxmert,chen2020uniter, su2019vl}. In some of these models, such as ViLBERT \citep{lu2019vilbert}, and LXMERT \citep{tan2019lxmert}, multi-modality fusion has been achieved via co-attention using a third transformer which contains fused information independently obtained from respective vision and language encoders. On the contrary, VisualBERT \citep{li2020visualbert}, VL-BERT \citep{su2019vl}, and UNITER \citep{chen2020uniter} employ a merged attention strategy to fuse both image patches and text features together into a unified transformer through corresponding image and text embedders. In addition to these, OSCAR \citep{li2020oscar} uses object tags as inputs. VinVL \citep{zhang2021vinvl} follows a similar strategy to that of OSCAR, the only difference being their novel 3-way contrastive loss which optimizes the training objectives used for VQA and text-image matching. VL-T5 \citep{cho2021unifying} exploits bounding-box coordinate information, image IDs, and region IDs along with ROI features for visual embedding. Here, encoded visual and textual features are fed into a bi-directional multi-modal encoder and an auto-regressive text decoder framework, respectively, for pre-training.

In all the above methods, pre-trained object detectors are kept frozen during the training. Furthermore, extracting region-level features from images can be tedious. To address these shortcomings, end-to-end pre-training methods have been developed. PixelBERT \citep{huang2020pixel} uses a CNN-based vision encoder and sentence encoder to obtain image and text representations, respectively. These representations are subsequently fed into a transformer via a cross-modality alignment. SOHO \citep{huang2021seeing} uses grid features-based discretization via a learned vision dictionary which is then fed into a cross-modal module. SimVLM \citep{wang2021simvlm} uses CNN and text token embedding for image and text feature representation extraction with a unified encoder-decoder transformer trained on a PrefixLM objective. Finally, MDETR \citep{kamath2021mdetr} uses CNN and RoBERTa (along with corresponding projection layers) for image and text feature extraction. These extracted features are concatenated before passing through a unified transformer trained on 1.3M Image-Text-Box (I-T-B) annotated data.

In recent years, the rise of Vision Transformers (ViT) \citep{dosovitskiy2021an} has motivated the research community to have an all-transformer framework by incorporating ViTs (instead of CNN backbones) in VLP models. Image patch features and text token embeddings are fed directly into a ViT model for pre-training in ViLT \citep{kim2021vilt}. Visual Parsing \citep{xue2021probing}, ALBEF \citep{li2021align}, and METER \citep{dou2022empirical} use ViTs as vision encoders for image feature generation. ALBEF and METER use co-attention in their pre-training frameworks for multimodality fusion.

Another class of VLP models in the form of CLIP \citep{radford2021learning}, DeCLIP \citep{li2021supervision}, and ALIGN \citep{jia2021scaling} has been introduced lately. Although known for their impressive zero-shot recognition ability and excellent transferability to downstream tasks, these models typically rely on huge amounts of image-text pairs for pre-training. Contrastive loss forms the core component of the pre-training objectives in these VLP models. \textcolor{black}{In such models (e.g., CLIP \citep{radford2021learning}, DeCLIP \citep{li2021supervision}), separate encoders have been used for each modality. On the contrary, modality-shared contrastive language-image pre-training (MS-CLIP) \citep{you2022msclip} leverages knowledge distribution across multiple modalities (image and text) through parameter sharing. In their unified framework, the parameters which are being shared between two modalities include the attention and feedforward modules and the layerNorm layers.}

GLIP \citep{li2022grounded} and GLIPv2 \citep{zhang2022glipv2} use a localization loss along with a word-region alignment loss for pre-training corresponding encoders using image-text-box annotations. BLIP \citep{li2022blip} employs image and text encoders connected through a cross-modality multi-head attention which are pre-trained on image-text pairs using contrastive and language modeling objectives. OmniVL \citep{wang2022omnivl} utilizes a unified image (and video) encoder and a text encoder pre-trained on image-text, image-label, video-text, and video-label pairs using unified vision-language contrastive, vision-language matching and language modeling losses. Furthermore, a visual-grounded alignment decoder is also present for facilitating better learning and alignment between various modalities. X-VLM \citep{zeng2022xvlm} employs a vision transformer to extract features from the subset of patches representing images/regions/objects. These patch features are then paired with associated text features for contrastive learning, matching, and masked language modeling. Additionally, image and text pairings are also done for bounding-box prediction which is used to locate visual concepts in the image. CMAL \citep{ma2022cmal} proposes interactions between features (obtained from respective image and text encoders) via cross-modal associative mappings which help in fine-grained semantic alignment between the learned representations. LOUPE \citep{li2022loupe} implements token-level and semantics-level Shapley interaction modeling with global image-text contrastive loss (in a dual-encoder setting) for explicit learning of fine-grained semantic alignment between visual regions and textual phrases without using expensive bounding-box annotations. FILIP \citep{yao2022filip} removes the need for cross-modality attention fusion by modeling the fine-grained semantic alignment between visual and textual tokens via a novel cross-modal late interaction mechanism in contrastive loss. TCL \citep{yang2022vision} uses global cross-modal alignment, intra-modal alignment, and local mutual information maximization losses along with masked language modeling and image-text matching to learn robust image-text representations during pre-training. UniCL \citep{yang2022unified} utilizes a unified learning method with a two-way contrastive loss (image-to-text and text-to-image) in the image-text-label space which can learn representations from either of the image-label and image-text data or both. UniTAB \citep{yang2022unitab} employs a transformer-based encoder-decoder framework that can jointly output open-ended text and box, encouraging alignment between words and boxes.

\textcolor{black}{In order to accelerate the convergence of VL pretraining, \cite{flm2023} proposed free language modeling (FLM) which addresses the issues inherent to masked language modeling (MLM) and autoregressive modeling. Using FLM as a pre-training objective, the authors have achieved impressive performance on several downstream tasks. Fame-ViL \citep{famevil2023} introduces a parameter-efficient VLP approach employing a task-versatile architecture with cross-attention and task-specific adapters. Fame-ViL applies a single model for various heterogeneous fashion tasks achieving performance gains over previous SOTA benchmarks. \cite{ptp2023} have introduced a simple yet effective position-guided text prompt (PTP) paradigm to improve the visual grounding capability of existing cross-modal VL architectures and help them better handle various downstream tasks. \cite{softmask2023} have proposed a VL framework based on the explainable soft feature masking and regularization via diversification strategies for improving the performance of VL models in several downstream tasks. \cite{blip22023} have devised BLIP-2, where a lightweight querying transformer is pre-trained using a two-stage strategy to bridge the modality gap. A frozen encoder is used in the first stage to bootstrap VL representation learning, and vision-to-language generative learning is bootstrapped in the second stage employing a frozen LLM, allowing zero-shot generation capabilities. \cite{oneR2023} have developed a simple and unified VL model (as a single tower) in a modality-agnostic manner. BEiT-3 \citep{beit2023} introduces a general-purpose multimodal foundation model to pre-train a multiway transformer by performing masked data modeling on inputs irrespective of modalities (i.e., images, texts, and image-text pairs). FLIP \citep{flip2023} extends CLIP \citep{radford2021learning} by performing contrastive learning on pairs of masked image patches and corresponding texts without reconstructing the masked image content. GIT \citep{git2023} unifies the VL architecture (an image encoder and a text decoder) under a single language modeling task while also scaling up the pre-training data and the model size to gain superior performance on captioning and question-answering downstream tasks.}

FIBER \citep{dou2022coarse} fuses vision and language encoder backbones through merged co-attention which are then pre-trained on 4M data with two-stage pre-training (coarse- and fine-grained). Image-text pairs are used in the coarse-grained pre-training stage which is then followed by a fine-grained pre-training stage with image-text-box annotations. However, these bounding box annotations come with extra overheads. Therefore, in our model, VoLTA, we propose an alternate solution for optimal-transport based local feature-level alignment using global image-caption annotations which performs well not only on coarse-grained tasks (such as VQA and Image Captioning), but also on fine-grained tasks (such as Referring Expression Comprehension and Object Detection). Table \ref{tab:vlp_glossary} encapsulates an overview of the details of all these aforementioned methods.

\section{Downstream Datasets} \label{sec:downstream_datasets}

\begin{table}[!t]
\centering
\vspace{2mm}
\resizebox{0.75\columnwidth}{!}{\begin{tabular}{p{2.2 cm}|l|c|c|c|c|c}
\hline
\bf Modality & \bf Task & \bf Dataset & \bf Image Src & \bf \#Images & \bf \#Text & \bf Metric\\
\hline
\multirow{6}{*}{Uni-modal} & \multirow{3}{*}{Image Classfn.} & IN$-$1K & IN$-$1K & 1.3M & - & \multirow{1}{*}{Accuracy}\\
& & COCO & COCO & 123K & - & F$1$\\
& & VOC07$+$12 & VOC07$+$12 & 16K & - & mAP\\
\cdashline{2-7}
& \multirow{2}{*}{Object Det.} & COCO & COCO & 123K & - & \multirow{2}{*}{AP$^{bb}$}\\
& & VOC07$+$12 & VOC07$+$12 & 16K & - & \\
\cdashline{2-7}
& Instance Seg. & COCO & COCO & 123K & - & AP$^{mk}$\\
\hline
\multirow{4}{2.2 cm}{Multi-modal coarse-grained} & VQA & VQA & COCO & 204K & 1.1M & VQA-Score\\
& NLVR$^2$ & NLVR$^2$ & Web Crawled & 214K & 107K & Accuracy\\
& IR-TR & Flickr30K & Flickr30K & 32K & 160K & Recall@1\\
& Captioning & COCO & COCO & 123K & 615K & B$@4$,M,C,S\\
\cdashline{1-7}
\multirow{5}{2.2 cm}{Multi-modal fine-grained} & \multirow{3}{*}{Ref. Exp. Comp.} & RefCOCO & \multirow{3}{*}{COCO} & 20K & 142K & \multirow{3}{*}{Accuracy}\\
& & RefCOCO$+$ & & 20K & 142K & \\
& & RefCOCOg & & 26K & 95K & \\
\cdashline{2-7}
& \multirow{2}{*}{Mul. Obj. Det.} & COCO & COCO & 123K & 615K & \multirow{2}{*}{AP}\\
& & LVIS Mini & COCO & 123K & 615K & \\
\hline
\end{tabular}}
\vspace{-2mm}
\caption{\textbf{Dataset statistics for uni-modal and multi-modal downstream tasks.}}
\label{tab:downstream_datasets}
\end{table}

Our downstream tasks can be categorized into three groups: uni-modal, multi-modal coarse-grained, and multi-modal fine-grained.

\noindent \textbf{Uni-modal:} For uni-modal tasks, we fine-tune (and validate) our pre-trained model on ImageNet-1k \citep{deng2009imagenet} for image classification, VOC07+12 \citep{everingham2010pascal} for image classification and object detection, and COCO \citep{lin2014microsoft} for image classification, object detection, and instance segmentation.

\noindent \textbf{Multi-modal Coarse-grained:} Here, we fine-tune (and validate) our pre-trained model on VQAv2 \citep{antol2015vqa} for visual question answering, NLVR$^2$ \citep{suhr2019corpus} for visual reasoning, Flickr30k \citep{plummer2015flickr30k} for image and text retrieval, and COCO \citep{lin2014microsoft} for image captioning.

\noindent \textbf{Multi-modal Fine-grained:} For these tasks, we fine-tune (and validate) our pre-trained model on RefCOCO, RefCOCO+, and RefCOCOg \citep{kazemzadeh2014referitgame, yu2016modeling} for referring expression comprehension, and COCO \citep{lin2014microsoft} and LVIS Mini \citep{gupta2019lvis} for language-conditioned object detection.

Several multi-modal downstream tasks are built based on the COCO dataset, where the validation and test splits of these downstream tasks are scattered across the raw COCO splits. Therefore, during pre-training, we carefully selected the portion of the COCO dataset which does not overlap with the validation/test splits of these multi-modal downstream tasks.

\section{Implementation Details $\&$ Hyper-parameter Values} \label{sec:hyperparameter_values}

\subsection{Data Augmentation} \label{sec:augmentation}

 We use ResNet$50$/Swin-T/Swin-B \citep{he2016deep, liu2021swin} as image encoder  and RoBERTa \citep{liu2019roberta} as text encoder. Each encoder is followed by a projector network which is a $3$-layer MLP with the configuration [$d$-$2048$-$2048$-$1024$]. Here, $d$ represents the embedding dimension of the encoder's output.

\begin{table}[!h]
\centering
\vspace{2mm}
\resizebox{0.5\columnwidth}{!}{\begin{tabular}{l|c|c|c}
\hline
\bf Data Type & \bf View \# & \bf Augmentation & \bf Probability\\
\hline
\multirow{12}{*}{Image} & 1 & \texttt{RandomResizedCrop} & 1.0\\
 & 1 & \texttt{RandomHorizontalFlip} & 0.5\\
 & 1 & \texttt{ColorJitter} & 0.8\\
 & 1 & \texttt{RandomGrayscale} & 0.2\\
 & 1 & \texttt{GaussianBlur} & 1.0\\
 & 1 & \texttt{Solarization} & 0.0\\
 \cdashline{2-4}
 & 2 & \texttt{RandomResizedCrop} & 1.0\\
 & 2 & \texttt{RandomHorizontalFlip} & 0.5\\
 & 2 & \texttt{ColorJitter} & 0.8\\
 & 2 & \texttt{RandomGrayscale} & 0.2\\
 & 2 & \texttt{GaussianBlur} & 0.1\\
 & 2 & \texttt{Solarization} & 0.2\\
\hline
\multirow{8}{*}{Text} & 1 & Synonym Replacement & 0.1\\
 & 1 & Random Insertion & 0.1\\
 & 1 & Random Swap & 0.1\\
 & 1 & Random Deletion & 0.1\\
 \cdashline{2-4}
 & 2 & Synonym Replacement & 0.1\\
 & 2 & Random Insertion & 0.2\\
 & 2 & Random Swap & 0.1\\
 & 2 & Random Deletion & 0.2\\
\hline
\end{tabular}}
\caption{\textbf{Image and text augmentation details.}}
\label{tab:img_txt_aug}
\vspace{-2mm}
\end{table}

\noindent \textbf{Image Augmentations:} Two sets of random transformations sampled from an augmentation pool are applied on each input image to generate two disparate distorted views. The augmentation policy is composed of \texttt{RandomResizedCrop}, \texttt{RandomHorizontalFlip}, \texttt{ColorJitter}, \texttt{RandomGrayscale}, \texttt{GaussianBlur}, and \texttt{Solarization} augmentations. \texttt{RandomResizedCrop} is applied with a probability of 1.0, whereas the remaining ones are applied randomly with varying probabilities following \citet{zbontar2021barlow} (see Table \ref{tab:img_txt_aug}).

\noindent \textbf{Text Augmentations:} Two sets of random transformations are applied on input text using EDA \citep{wei2019eda} including synonym replacement, random insertion, random swap, and random deletion with different probabilities as outlined in Table \ref{tab:img_txt_aug}.

\subsection{Pre-training Setup} \label{sec:pre-training_setup}

Table \ref{tab:hyperparams} shows the details of hyper-parameters used during training.

\model\ comprises a vision encoder and a language encoder with a merged co-attention for cross-modality fusion. In our experiments, we have considered two types of vision encoder backbones - ResNet-$50$ \citep{he2016deep} and Swin Transformer \citep{liu2021swin}. For fair comparisons with related works \citep{dou2022empirical,dou2022coarse}, the input image resolution for ResNet-50 encoder backbone is kept as 224 $\times$ 224, whereas for Swin-B, it is 384 $\times$ 384. The output embedding dimension of the image encoder in both cases is 1024. Similarly, to be consistent with \citet{dou2022empirical,dou2022coarse}, we have selected RoBERTa as the language encoder with a vocabulary size of 50265, a tokenizer as `roberta-base', a maximum input text length of 30, and an output embedding dimension of 768 (please refer to Table \ref{tab:hyperparams} for more details).

Separate projector heads follow vision and language encoders. A projector head consists of 3 linear layers, each with 2048 output units (except for the last one, which has 1024 output units), followed by a Batch Normalization layer and ReLU activation (for exact configuration, please refer to Table \ref{tab:hyperparams}). The final projected output denotes the input (image/text) feature representation used in downstream tasks. The embeddings (i.e., output from respective encoders) are fed into the loss function of VoLTA to learn these representations.

The loss function of VoLTA includes four different loss components, namely, multi-modal Barlow Twins for intra- and inter-modality redundancy reduction, GOT for alignment of local features, and MLM and ITM together for encouraging cross-modal attention fusion. For MLM, we randomly mask $15\%$\footnote{\textcolor{black}{Following BERT, we decompose this $15\%$ into $10\%$ random words, $10\%$ unchanged, and $80\%$ with a special token [MASK].}} (MLM probability in Table \ref{tab:hyperparams}) of the input tokens, and the model is trained to reconstruct the original tokens. For ITM, the model predicts whether a given image-text pair is matched.

For optimization, we follow the same protocol as described in \citet{zbontar2021barlow}, where we use the LARS \citep{you2017large} optimizer to train our model for 20 epochs with a batch size of 256. A base LR of 0.1 is used for the weights and 0.0048 for the biases and batch normalization parameters which are then multiplied by a factor of 2. We employ a learning rate warm-up (linear) up to a period of 2 epochs followed by a cosine decay schedule to reduce the LR by a factor of 1000. A weight decay parameter 1e-6 is used, excluding the biases and batch normalization parameters. We conduct a grid search for the GOT loss hyperparameter ($w_\mathrm{GOT}$), and we empirically found the best value to be 100.

\begin{table}[!t]
\centering
\vspace{2mm}
{\resizebox{0.7\columnwidth}{!}{\begin{tabular}{l|c|c}
\hline
\bf Hyper-parameters & \bf Notation & \bf Value\\
\hline
\multicolumn{3}{c}{Model}\\
\hline
Img. proj. layer config. & - & $[d_i-2048-2048-1024]$\\
Img. embed. dim & $d_i$ & 1024\\
Img. reso. (RN-50 \& Swin-T) & - & 224 $\times$ 224\\
Img. reso. (Swin-B) & - & 384 $\times$ 384\\
Txt. proj. layer config. & - & $[d_t-2048-2048-1024]$\\
Txt. embed. dim & $d_t$ & 768\\
Tokenizer & - & 'roberta-base'\\
Vocab size & - & 50265\\
MLM prob. & - & 0.15\\
Max. length of text & - & 30\\
\# Heads of Xformer & - & 12\\
\# Layers of Xformer & - & 12\\
\# Fusion block & - & 6\\
Dropout rate & - & 0.1\\
Task names & - & 'BTGOT, MLM, ITM'\\
\hline
\multicolumn{3}{c}{Training}\\
\hline
Batch size & - & 256\\
Epochs & - & 20\\
Lambda\_BT & $\lambda$ & 0.005\\
WD and GWD loss weight & $\gamma$ & 0.1 \\
GOT loss weight & $w_{GOT}$ & 100.0 \\
Optimizer & - & LARS\\
Base LR for weights & - & 0.1\\
Base LR for biases & - & 0.0048\\
Momentum & - & 0.9\\
LR scheduler & - & Cosine LR decay (with linear warm-up)\\
Warm-up steps & - & 2 $\times$ Epochs\\
Weight decay & - & 1e-6\\
End LR factor & - & 0.001\\
Cosine LR amplitude factor & - & 0.5\\
\hline
\end{tabular}}

\caption{\textbf{Pre-training hyper-parameter details for \model.}}
\label{tab:hyperparams}}
\vspace{-2mm}
\end{table}

\noindent \textbf{Pre-training Cost:} Our Swin-B backbone takes 6 hours per epoch to train on 64 V$100$ GPUs, with per GPU batch-size of $4$.

\subsection{Downstream Setup} \label{sec:downstream_setup}

\subsubsection{Uni-modal downstream tasks}

\paragraph{Linear Evaluation:} For ImageNet, the linear classifier has been trained for 100 epochs with a batch size of 256, an LR of 0.3, and a cosine LR schedule. Cross-entropy loss is minimized with SGDM optimizer (momentum of 0.9), and a weight decay of 1e-6. For both COCO and VOC, the linear classifier has been trained for 100 epochs with AdamW optimizer with batch size of 256, an LR of 5e-2, and a weight decay of 1e-6.



\noindent \textbf{Object Detection:} For training the detection model, the detectron2 library \citep{wu2019detectron2} has been used. The backbone networks for Faster R-CNN \citep{ren2015faster} and Mask R-CNN \citep{he2017mask} has been initialized using our pre-trained model.

For VOC07+12, we used the \texttt{trainval} set comprising 16K images for training a Faster R-CNN \citep{ren2015faster} C-4 backbone for 24K iterations using a batch size of 16 across 8 GPUs (using SyncBatchNorm). The initial learning rate for the model is 0.15, which is reduced by a factor of 10 after 18K and 22K iterations. Linear warmup \citep{goyal2017accurate} is used with a slope of 0.333 for 1000 iterations.

For COCO, Mask R-CNN \citep{he2017mask} with a C-4 backbone on the COCO 2017 train split is used for training, and the results are reported on the val split. A learning rate of 0.03 is used, and other parameters are kept the same as in the 1$\times$ schedule in detectron2 \citep{wu2019detectron2}.

\subsubsection{Coarse-grained multi-modal downstream tasks}

\noindent \textbf{Vision-Language Classification (VQAv2 and NLVR$^2$):} Vision-Language Classification task encompasses VQAv2 and NLVR$^2$, whose hyper-parameter setup has been taken from METER \citep{dou2022empirical} and FIBER \citep{dou2022coarse}. Model finetuning is done with peak learning rates of 2e-5 for the backbones, 1e-4 for the cross-modal parameters, and 1e-3 for the head layer for 10 epochs with a batch size of 512. The image resolutions are set to 576 for VQAv2 and 384 for NLVR$^2$ and the models are evaluated with the VQA-Scores for VQAv2 and accuracy for NLVR$^2$ (Table \ref{tab:downstream_datasets}).

\noindent \textbf{Image-Text Retrieval (IRTR):} We follow \citet{dou2022coarse} for IR-TR setup for the Flickr30k dataset, where the cross-attention layers in the backbones are removed during IR-TR fine-tuning and evaluation. The peak learning rates are set to 2e-5 for the backbones, and 1e-4 for the head layer. Furthermore, a batch size of 1024 is considered, and each image resolution is set to 576. We evaluate the Recall@1 metric for both the text and image retrieval tasks as outlined in Table \ref{tab:downstream_datasets}.

\noindent \textbf{Image Captioning:} For image captioning, only the image-to-text attentions are kept for the cross-modality attention fusion, and the model is converted into a standard seq2seq model \citep{dou2022coarse}. We used a causal mask on the decoding side, and the outputs are predicted auto-regressively \citep{dou2022coarse}. Models are trained with the cross-entropy loss for 5 epochs with the peak learning rates of 5e-5 for the backbones, and 2.5e-4 for the rest of the parameters, followed by a two-stage finetuning. In the first stage, finetuning with GOLD \citep{pang2021text} is done for 5 epochs with a peak learning rate of 1e-5 for the backbones, since it is efficient and has been proven to be effective when the model input can correspond to different outputs. The second stage of fine-tuning involves CIDEr optimization where the learning rate is further reduced to 1e-6, and the model is trained for 3 epochs. A batch size of 512 is considered in both these cases, and a beam size of 5 is used during inference. Evaluation metrics include BLEU \citep{papineni2002bleu}, METEOR \citep{banerjee2005meteor}, CIDEr \citep{vedantam2015cider}, and SPICE \citep{anderson2016spice} scores (shown in Table \ref{tab:downstream_datasets}).

\subsubsection{Fine-grained multi-modal downstream tasks}

\noindent \textbf{Referring Expression Comprehension (REC):} We follow \citet{dou2022coarse} for training and evaluation on 3 different datasets (RefCOCO, RefCOCO+, and RefCOCO) where the models are finetuned with a batch size of 16 for 20 epochs. A warmup of 2000 steps with a peak LR of 1e-5 is used for the OD head as well as the rest of the model’s parameters. LR drops twice, once at 67\% and the other at 89\% of the total number of steps. Horizontal flip augmentation has been turned off during REC training because it was observed by \citet{dou2022coarse} that horizontal flip adversely affected the performance, particularly on the RefCOCO dataset. Accuracy is used as the evaluation metric in this case (Table \ref{tab:downstream_datasets}).

\noindent \textbf{Object Detection:} We follow the training and evaluation setup of \citet{dou2022coarse} for text-conditioned (multi-modal) object detection. For both COCO and LVIS datasets, the model has been finetuned for 24 epochs with a batch size of 32, an LR of 1e-5, and two learning rate drops, once at 67\% and the other at 89\% of the total number of steps. AP scores are used in this case for model evaluation (Table \ref{tab:downstream_datasets}).

\section{Error Analysis} \label{sec:error_analysis}

\begin{figure}[!t]
\centering
\includegraphics[scale=0.575]{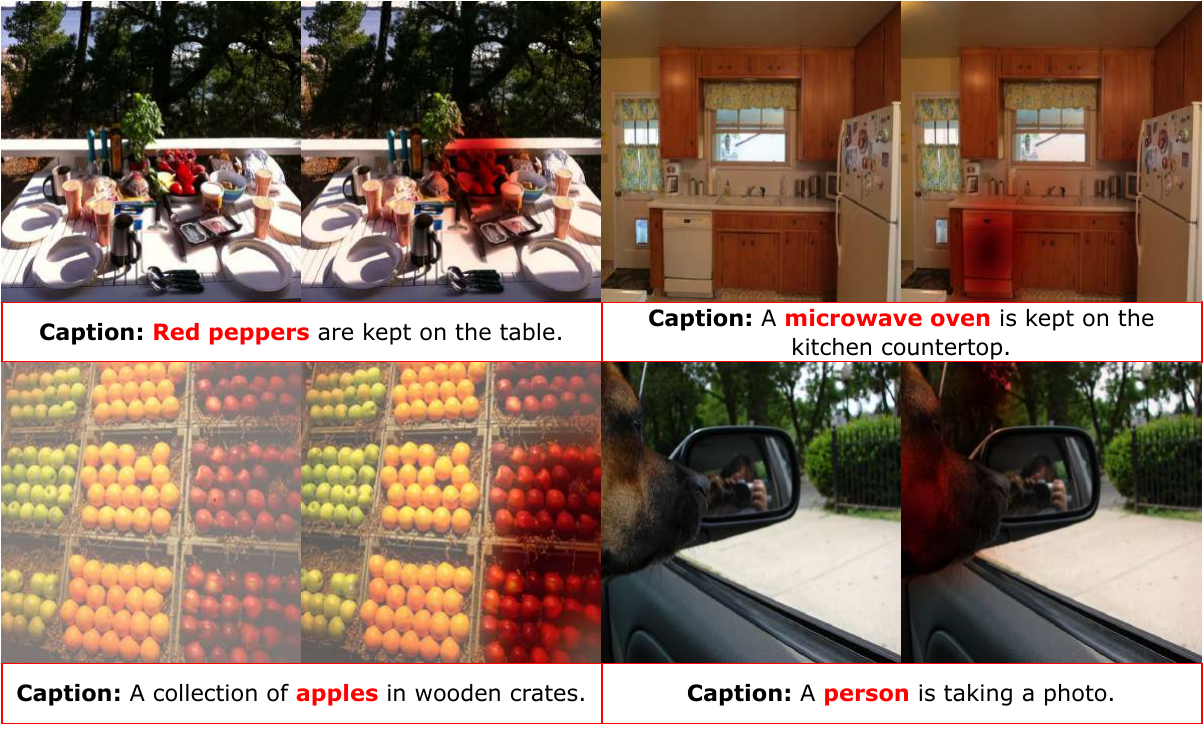} 
\caption{\textcolor{black}{\textbf{Limitations of our method:} tiny and hindered objects in cluttered environments are not distinctly attended by \model. We look at the optimal transport plan from GOT to represent the alignment between the words in \textcolor{red}{red} with the corresponding input image. The visualizations are generated with $224$p images, resulting in sequences of $196$ tokens for $16 \times 16$ patches. All image-caption pairs are taken from the COCO$2017$ train split.}}
\label{fig:error_analysis}
\end{figure}

\textcolor{black}{Although \model\ learns impressive fine-grained region-level understanding during pre-training, there are still some cases where the model fails to identify tiny and hindered objects, especially in cluttered environments. We show four such examples in Figure \ref{fig:error_analysis}. In the first image, the object `red peppers' is barely visible even in human eyes, and thus, \model\ can not precisely identify these objects. However, it can identify the coarse region (the fruit basket) where `red peppers' can be present. In the second image, \model\ confuses a `dishwasher' with a `microwave oven,' probably because the `microwave oven' is present in a cluttered environment, and both objects have similar appearances in low-resolution frames. In the third image, \model\ can correctly identify `red apples', but fails to spot `green apples', probably because \model\ has not seen enough such samples. In the last image, the face of the `person' is hindered by the camera, and \model\ fails to locate it. Since we pre-train \model\ with $224 \times 224$ images, such tiny objects are often hard to be distinguished. However, higher-resolution images will be more helpful in addressing such intricate scenarios, which we plan to explore in future works.}

\section{Additional Quantitative Results on Coarse-grained Vision-Language Tasks: Comparison with Methods using More Pre-training Data}

Table \ref{tab:vqa_nlvr_irtr_captioning_appendix} presents a comparison of \model\ on the multi-modal coarse-grained tasks with state-of-the-art methods pre-trained using magnitude more data. On VQA, \model\ beats ViLBERT, UNITER-B, \textcolor{black}{VILLA-B, UNIMO-B}, and ViLT-B, each pre-trained on $3-4$M datasets. Please note that \model\ is trained only on COCO \textcolor{black}{and VG}, whereas the other methods use a combination of COCO, VG, CC, and SBU datasets. Such strong performance proves the generalizability of \model. On captioning, \model\ beats Unified VLP, OSCAR, UFO-B, ViTCAP, \textcolor{black}{VinVL-B, METER-CLIP-B}, and XGPT. However, for IRTR and NLVR \model\ can not yield better performance over these baselines. We assume that the large domain difference between pre-training and downstream datasets is the reason behind the limited performance on IRTR and NLVR.

\begin{table}[!t]
\centering
\vspace{2mm}
\small
\setlength{\tabcolsep}{4pt}
\resizebox{0.99\textwidth}{!}{\begin{tabular}{@{}l | c | c c | c c | c c | l | c | c c c c@{}}
\hline

\multirow{2}{*}{\bf Method} & \multirow{2}{1.9 cm}{\bf \centering \#Pre-train Data} & \multicolumn{2}{c |}{\bf VQAv$2$} & \multicolumn{2}{c |}{\bf NLVR$^{2}$} & \multicolumn{2}{c |}{\bf F$30$k IRTR} & \multirow{2}{*}{\bf Method} & \multirow{2}{1.9 cm}{\bf \centering \#Pre-train Data} & \multicolumn{4}{c}{\bf COCO Captioning}\\ 

\cline{3-4}\cline{5-6}\cline{7-8}\cline{11-14}

& & \bf dev & \bf std & \bf dev & \bf test-P & \bf IR@$1$ & \bf TR@$1$ & & & \bf B$@4$ & \bf M & \bf C & \bf S \\

\hline
\multicolumn{8}{l |}{{\it{Models pre-trained on COCO ($123$k) and/or VG ($108$k)}}} & \multicolumn{6}{l}{{\it{Models fine-tuned without CIDEr optimization}}}\\
\cline{1-8}\cline{9-14}
SCAN & $108$k & $-$ & $-$ & $-$ & $-$ & 48.6 & 67.4 & VL-T5 & $180$k & 34.5 & 28.7 & 116.5 & 21.9\\
SCG & $108$k & $-$ & $-$ & $-$ & $-$ & 49.3 & 71.8 & VL-BART & $180$k & 35.1 & 28.7 & 116.6 & 21.5\\
PFAN & $108$k & $-$ & $-$ & $-$ & $-$ & 50.4 & 70.0 & \demph{Unified VLP} & \demph{$3$M} & \demph{36.5} & \demph{28.4} & \demph{117.7} & \demph{21.3}\\
MaxEnt & $123$k & 54.1 & 54.8 & $-$ & $-$ & $-$ & $-$ & \demph{OSCAR} & \demph{$4$M} & \demph{36.5} & \demph{30.3} & \demph{123.7} & \demph{23.1}\\
VisualBERT & $123$k & 70.8 & 71.0 & 67.4 & 67.0 & $-$ & $-$ & \demph{UFO-B} & \demph{$4$M} & \demph{36.0} & \demph{28.9} & \demph{122.8} & \demph{22.2}\\
LXMERT & $231$k & 72.4 & 72.5 & 74.9 & 74.5 & $-$ & $-$ & \demph{ViTCAP} & \demph{$4$M} & \demph{36.3} & \demph{29.3} & \demph{125.2} & \demph{22.6}\\
\textcolor{black}{SOHO} & \textcolor{black}{$231$k} & \textcolor{black}{73.2} & \textcolor{black}{73.4} & \textcolor{black}{76.3} & \textcolor{black}{77.3} & \textcolor{black}{72.5} & \textcolor{black}{\bf 86.5} & \demph{METER-CLIP-B} & \demph{$4$M} & \demph{38.8} & \demph{30.0} & \demph{128.2} & \demph{23.0}\\
\cline{1-8}
\multicolumn{8}{l|}{{\demph{\it{Models pre-trained on COCO, VG, SBU ($1$M) and/or CC ($3$M)}}}} & \demph{VinVL-B} & \demph{$4$M} & \demph{38.2} & \demph{30.3} & \demph{129.3} & \demph{23.6} \\
\cline{1-8}
\demph{ViLBERT} & \demph{$3$M} & \demph{70.5} & \demph{70.9} & \demph{$-$} & \demph{$-$} & \demph{58.2} & \demph{$-$} & \demph{XGPT} & \demph{$3.1$M} & \demph{37.2} & \demph{28.6} & \demph{120.1} & \demph{21.8}\\
\demph{UNITER-B} & \demph{$4$M} & \demph{72.7} & \demph{72.9} & \demph{77.2} & \demph{77.9} & \demph{72.5} & \demph{85.9} & \demph{FIBER-B} & \demph{$4$M} & \demph{39.1} & \demph{30.4} & \demph{128.4} & \demph{23.1}\\
\demph{VILLA-B} & \demph{$4$M} & \demph{73.6} & \demph{73.7} & \demph{78.4} & \demph{79.3} & \demph{74.7} & \demph{86.6} & \demph{FIBER-GOLD-B} & \demph{$4$M} & \demph{40.3} & \demph{30.7} & \demph{133.6} & \demph{23.6}\\
\cline{9-14}
\demph{UNIMO-B} & \demph{$4$M} & \demph{73.8} & \demph{74.0} & \demph{$-$} & \demph{$-$} & \demph{$-$} & \demph{$-$} & \textcolor{black}{\CC{}\model-GOLD-B} & \textcolor{black}{\CC{}$231$k} & \textcolor{black}{\CC{}38.9} & \textcolor{black}{\CC{}30.5} & \textcolor{black}{\CC{}128.5} & \textcolor{black}{\CC{}23.4}\\
\cline{9-14}
\demph{ViLT-B} & \demph{$4$M} & \demph{71.3} & - & \demph{75.7} & \demph{76.1} & \demph{64.4} & \demph{83.5} & \multicolumn{6}{l}{{\it{Models fine-tuned with CIDEr optimization}}}\\
\cline{9-14}
\demph{ALBEF-B} & \demph{$4$M} & \demph{74.5} & \demph{74.7} & \demph{80.2} & \demph{80.5} & \demph{82.8} & \demph{94.3} & \demph{ViTCAP} & \demph{$4$M} & \demph{41.2} & \demph{30.1} & \demph{138.1} & \demph{24.1}\\
\demph{VLMo-B} & \demph{$4$M} & \demph{76.6} & \demph{76.9} & \demph{82.8} & \demph{83.3} & \demph{79.3} & \demph{92.3} & \demph{VinVL-B} & \demph{$4$M} & \demph{40.9} & \demph{30.9} & \demph{140.4} & \demph{25.1}\\
\demph{METER-Swin-B} & \demph{$4$M} & \demph{76.4} & \demph{76.4} & \demph{82.2} & \demph{83.5} & \demph{79.0} & \demph{92.4} & \demph{FIBER-B} & \demph{$4$M} & \demph{42.8} & \demph{31.0} & \demph{142.8} & \demph{24.3}\\
\demph{FIBER-B} & \demph{$4$M} & \demph{78.6} & \demph{78.4} & \demph{84.6} & \demph{85.5} & \demph{81.4} & \demph{92.9} & \demph{FIBER-GOLD-B} & \demph{$4$M} & \demph{43.4} & \demph{31.3} & \demph{144.4} & \demph{24.6}\\
\hline
\rowcolor{Light}
\textcolor{black}{\model-B} & \textcolor{black}{$231$k} & \textcolor{black}{\bf 74.6} & \textcolor{black}{\bf 74.6} & \textcolor{black}{\bf 76.7} & \textcolor{black}{\bf 78.1} & \textcolor{black}{\bf 72.7} & \textcolor{black}{83.6} & \textcolor{black}{\model-GOLD-B} & \textcolor{black}{$231$k} & \textcolor{black}{\bf 40.2} & \textcolor{black}{\bf 30.9} & \textcolor{black}{\bf 137.5} & \textcolor{black}{\bf 23.7} \\
\hline
\end{tabular}}
\caption{\textbf{Multi-modal coarse-grained downstream: visual question answering, visual reasoning, retrieval, and captioning.} Methods pre-training with a significantly larger dataset are colored gray. For captioning, $4$ metrics are reported - B$@4$: BLEU$@4$, M: METEOR, C: CIDEr, S: SPICE. The best comparable results are in \textbf{bold}. \model-B denotes Swin-B backbone. }\label{tab:vqa_nlvr_irtr_captioning_appendix}
\end{table}

\section{Additional Qualitative Results}

\begin{figure}[!t]
\centering
\includegraphics[width=\textwidth]{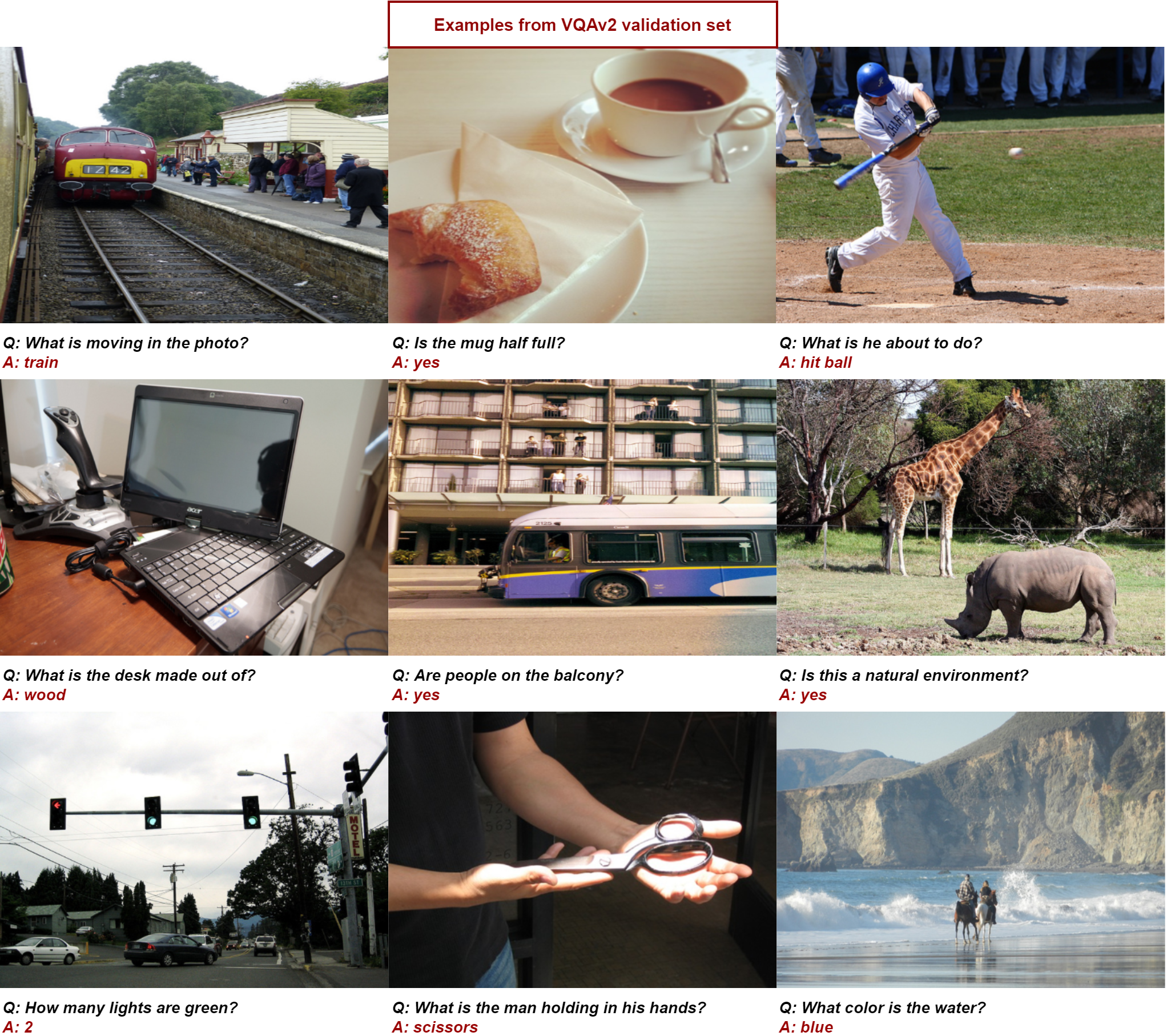} 
\caption{\textbf{Examples on Visual Question Answering from VQAv2 validation dataset.} We display a variety of examples (e.g., number of items, the color of objects, type of objects, events, and actions) with respective answers predicted by VoLTA.}
\label{fig:vqav2_examples}
\vspace{-3mm}
\end{figure}

\begin{figure}[!h]
\centering
\includegraphics[width=\textwidth]{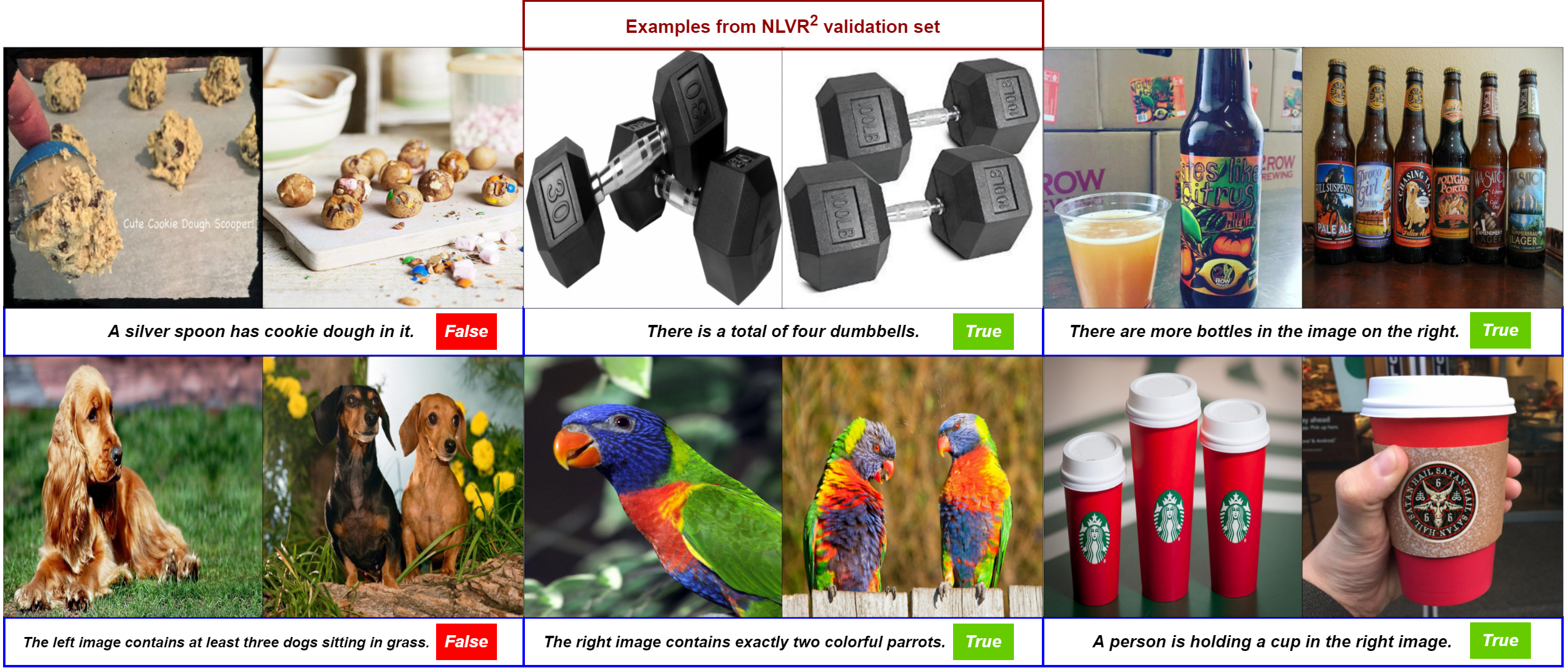} 
\caption{\textbf{Examples on Visual Reasoning from NLVR$^2$ validation dataset.} For each statement (text prompt), 2 images are shown alongside each other and VoLTA predicts whether the given statement is True (green box) or False (red box).}
\label{fig:nlvr2_examples}
\end{figure}

\noindent \textbf{Visual Question Answering and Visual Reasoning:} Visual question answering (VQA) is a widely recognized multi-modal task that infers an answer in response to a text-based question about an image. In Figure \ref{fig:vqav2_examples}, we demonstrated several examples image-question pairs and corresponding answers predicted by \model\ on the VQAv2 validation set. The primary aim of the visual reasoning task is to ascertain the veracity of a natural language statement against an associated image pair. Figure \ref{fig:nlvr2_examples} displays examples of responses (True/False) predicted by \model\ on the NLVR$^2$ validation set.

\begin{figure}[!h]
\centering
\includegraphics[width=0.8\textwidth]{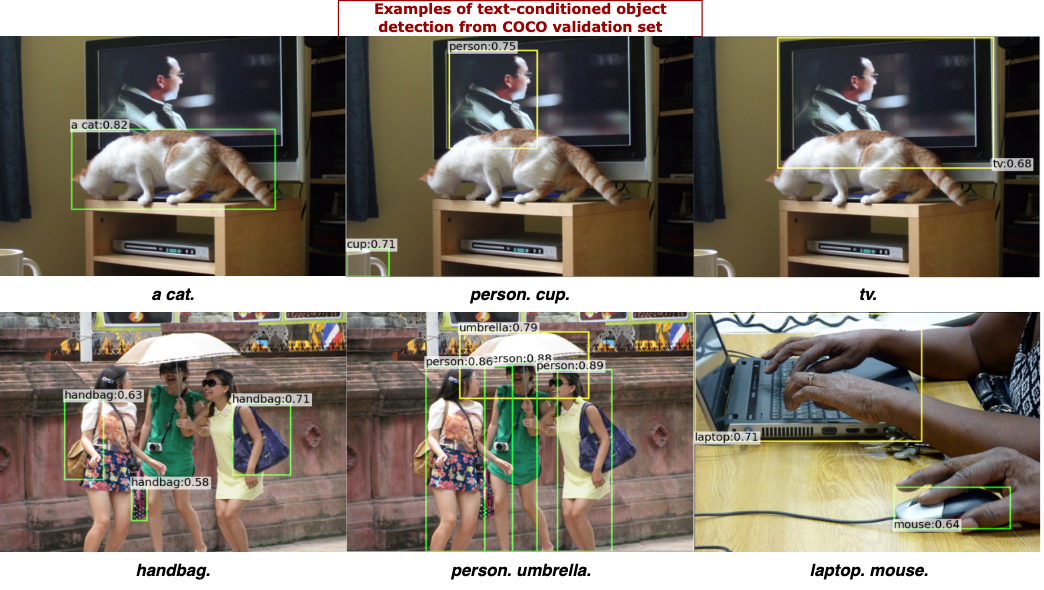} 
\caption{\textbf{Examples of Object Detection from COCO validation dataset with various text prompts.} Our model predicts boxes relevant to the text (caption) and labels them with the corresponding spans.}
\label{fig:coco_detection_examples}
\end{figure}

\noindent \textbf{Language-conditioned Object Detection:} Object detection forms an indispensable constituent of several multi-modal understanding systems. However, the conventional object detection pipeline is employed as a black-box tool and predicts all possible objects in the image. On the other hand, for better apprehension of combinations of these objects in free-form texts, a language-conditioned object detection task is considered \citep{kamath2021mdetr, dou2022coarse}. We use pre-trained \model\ and fine-tuned and evaluated COCO and LVIS datasets for the text-conditioned object detection task. As illustrated in Figure \ref{fig:coco_detection_examples}, \model\ predicts bounding boxes relevant to the text prompts (captions) and labels them with the corresponding spans from the text. For example, the top-middle image has 4 objects. However, based on the text prompt, our model predicts boxes only for \textit{person} and \textit{cup}.

\begin{figure}[!h]
\centering
\includegraphics[width=0.85\textwidth]{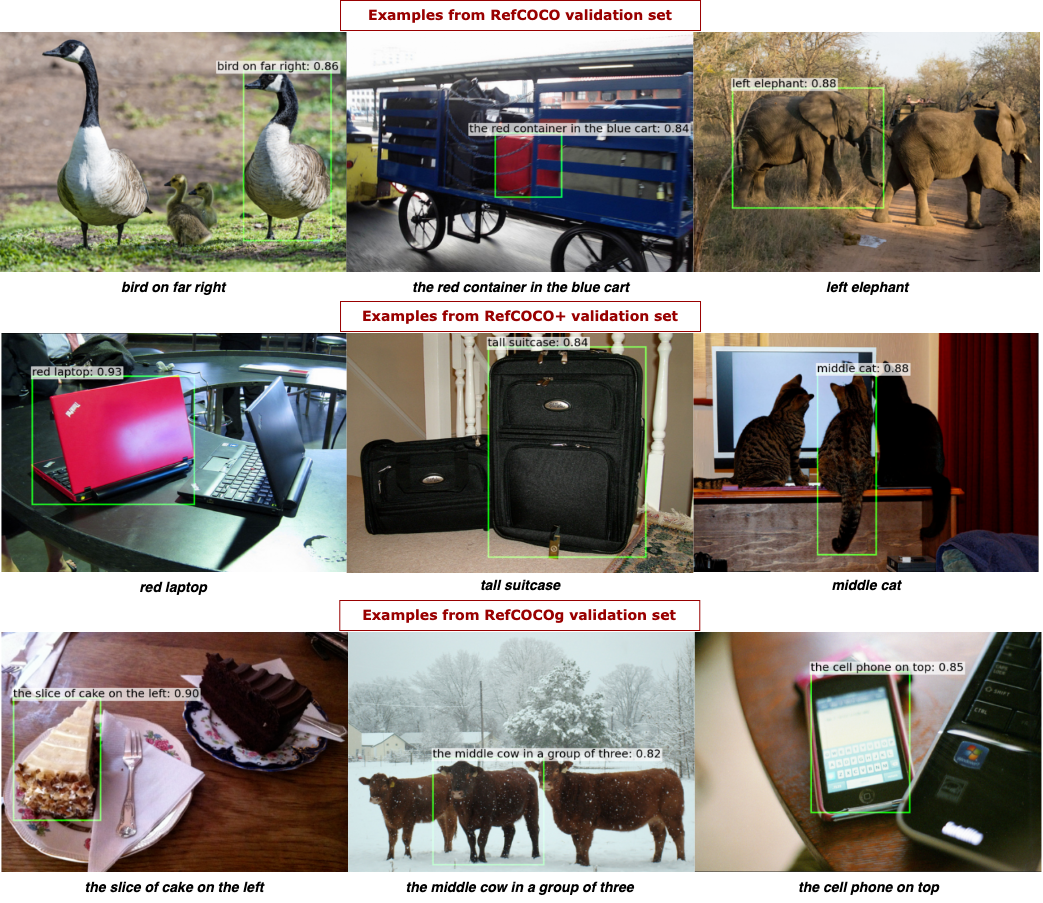} 
\caption{\textbf{Examples of Referring Expression Comprehension from RefCOCO (top), RefCOCO+ (middle) and RefCOCOg (bottom) validation datasets.} The expressions in RefCOCOg typically have florid and longer constructions as compared to RefCOCO and RefCOCO+. The model has access to the entire text and uses it to disambiguate amongst different objects in the image.}
\label{fig:refcoco_examples}
\end{figure}

\noindent \textbf{Referring Expression Comprehension (REC):} The objective of REC is to align the entire referring expression (text) with the corresponding box by disambiguating among the several occurrences of an object belonging to the same category and therefore, one box per expression is to be predicted. For example, the bottom-left image in Figure \ref{fig:refcoco_examples} depicts \model's box prediction for the corresponding referring expression: \emph{the slice of cake on the left}.

\begin{figure}[!ht]
     \centering
     \begin{subfigure}[b]{0.49\textwidth}
         \centering
         \includegraphics[width=\textwidth]{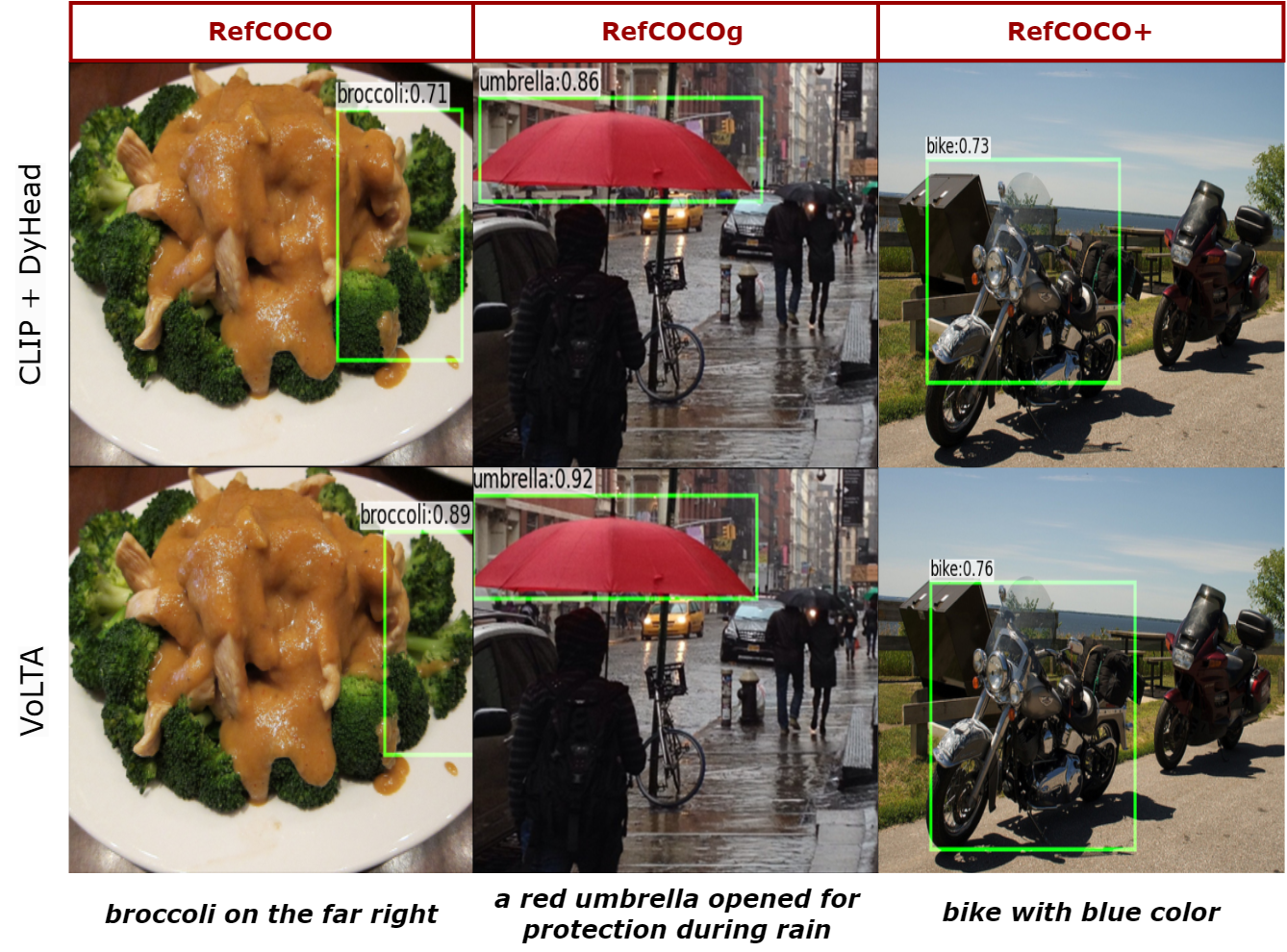}
         \caption{\textcolor{black}{Simpler samples, both CLIP and \model\ succeed.}}
         \label{fig:}
     \end{subfigure}
     \begin{subfigure}[b]{0.49\textwidth}
         \centering
         \includegraphics[width=\textwidth]{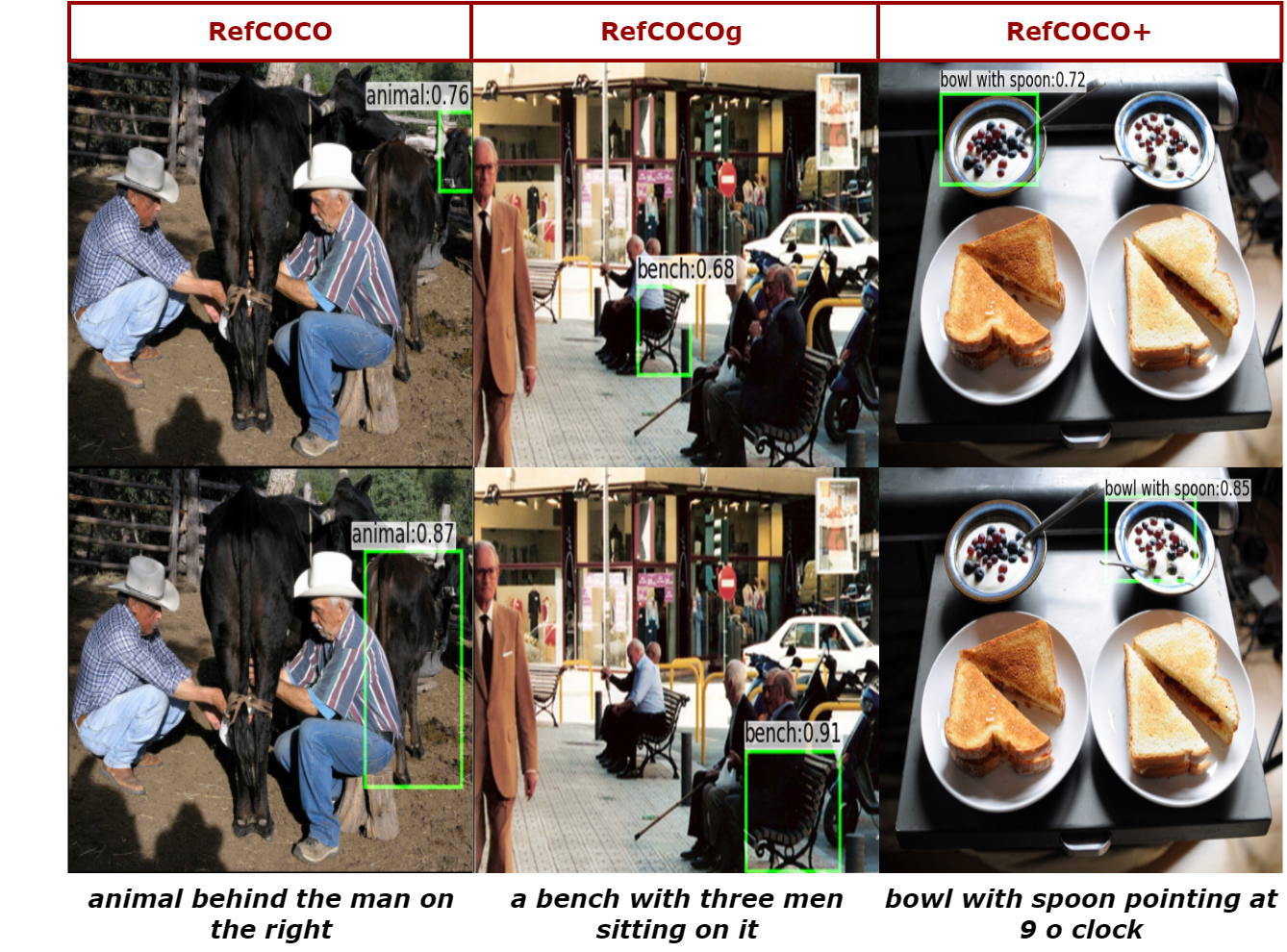}
         \caption{\textcolor{black}{Harder samples, CLIP fails, but \model\ succeeds.}}
     \end{subfigure}
         \caption{\textcolor{black}{Comparative qualitative evaluation between frozen CLIP + Dynamic Head \citep{dai2021dynamic}, and VoLTA on different RefCOCO, RefCOCOg, RefCOCO+ validation samples.}}
    \label{fig:clip_volta_rec}
\end{figure}

\noindent \textbf{\textcolor{black}{Comparison of CLIP vs. \model\ on Referring Expression Comprehension (REC):}} \textcolor{black}{Figure \ref{fig:clip_volta_rec} shows a comparative qualitative evaluation between frozen CLIP + dynamic head \citep{dai2021dynamic}, and VoLTA. We concatenate the vision and text features from the CLIP encoders and train a dynamic head on top of frozen features for the REC task. Since CLIP is a dual-encoder system pre-trained with image-level features, it can not learn superior fine-grained features. Hence, CLIP fails on harder REC samples. For example, if multiple similar-looking objects (benches or bowls) exist in an image, CLIP fails to distinguish between them. However, \model\ succeeds on such complex samples, which can be attributed to the fine-grained alignment achieved by the GOT objective.}

\begin{figure}[!h]
\centering
\includegraphics[scale=0.32]{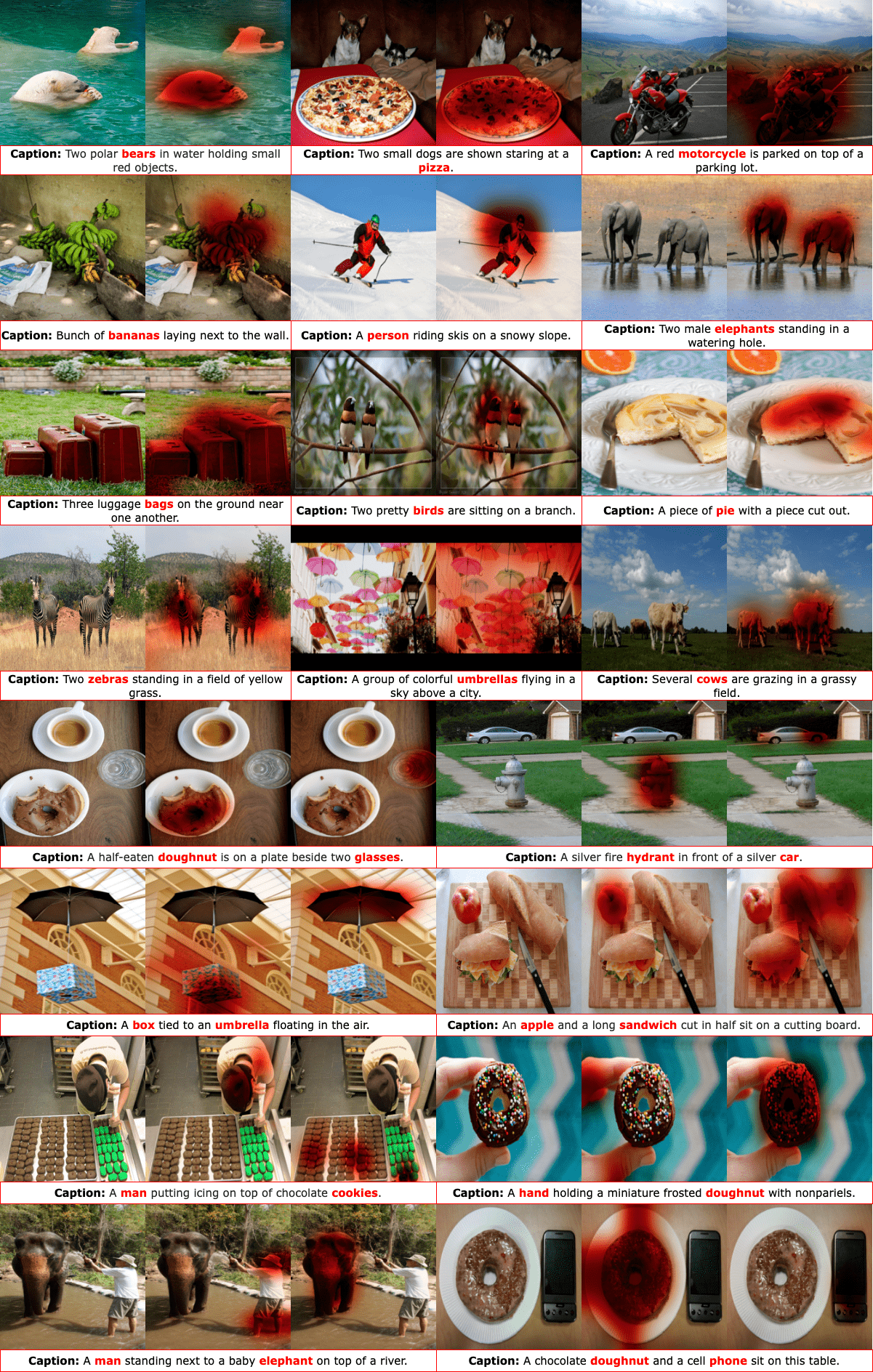} 
\caption{\textbf{This figure shows how different words in captions attend relevant image regions, produced by the GOT module of \model\ pre-trained on COCO.} Extension of Figure \ref{fig:main_visualization}. All image-caption pairs are taken from the COCO$2017$ train split.  The visualizations  are generated with $224$p images, resulting in sequences of $196$ tokens for $16 \times 16$ patches.}
\label{fig:appendix_visualization}
\end{figure}

\end{document}